\documentclass[11pt]{article}
\usepackage[margin=1in]{geometry}
\usepackage[T1]{fontenc}
\usepackage[utf8]{inputenc}
\usepackage{mathptmx}
\usepackage[scaled=0.92]{helvet}
\usepackage{amsmath,amssymb,amsthm,mathtools}
\usepackage{bm}
\usepackage{graphicx}
\usepackage{booktabs,multirow,array,longtable}
\usepackage{caption,subcaption}
\usepackage{xcolor}
\usepackage{setspace}
\usepackage[expansion=false]{microtype}
\usepackage{enumitem}
\usepackage{authblk}
\usepackage[numbers,sort&compress,super]{natbib}
\usepackage[colorlinks=true,linkcolor=blue!60!black,citecolor=blue!60!black,urlcolor=blue!60!black]{hyperref}
\usepackage{cleveref}

\graphicspath{{figs/}}
\usepackage{tabularx}

\crefname{figure}{Fig.}{Figs.}
\Crefname{figure}{Figure}{Figures}
\crefname{table}{Table}{Tables}
\crefname{equation}{equation}{equations}
\creflabelformat{equation}{(#2#1#3)}

\newcommand{\R}{\mathbb{R}}
\newcommand{\M}{\mathcal{M}}

\newcommand{\K}{\mathcal{K}}
\newcommand{\Pc}{\mathcal{P}}
\newcommand{\Sm}{\mathcal{S}}
\newcommand{\G}{\mathcal{G}}
\newcommand{\J}{\mathcal{J}}
\newcommand{\F}{\mathcal{F}}
\newcommand{\xv}{\mathbf{x}}
\newcommand{\uv}{\mathbf{u}}
\newcommand{\bv}{\mathbf{b}}
\newcommand{\nv}{\mathbf{n}}
\newcommand{\Kb}{\mathbf{K}}
\newcommand{\Id}{\mathbf{I}}
\DeclareMathOperator{\tr}{tr}
\newcommand{\norm}[1]{\left\lVert #1 \right\rVert}
\newcommand{\cand}[1]{\texttt{#1}}

\title{\bfseries Multi-agent System for Automatic Discovery of Pareto-optimal Meta-solvers for PDEs}

\author[1,2,$\dagger$]{Youngkyu Lee}
\author[1,$\dagger$]{Shanqing Liu}
\author[1]{J\'er\^ome Darbon}
\author[1,*]{George Em Karniadakis}
\affil[1]{Division of Applied Mathematics, Brown University, Providence, RI 02912, USA}
\affil[2]{Department of Mathematics, Kyung Hee University, Seoul 02447, Republic of Korea}
\affil[$\dagger$]{These authors contributed equally to this work.}
\affil[*]{Corresponding author: george\_karniadakis@brown.edu}
\date{}

\begin{document}
\maketitle

\begin{abstract}
\noindent

Designing linear solvers for discretized partial differential equations requires problem-specific expertise. We present a multi-agent framework, in which language-model agents propose, critique, implement, and evaluate solver configurations assembled from Krylov methods, preconditioners, block factorizations, and neural operators.
Inside the proposing agent, a non-dominated sorting genetic evolutionary algorithm evolves the population of meta-solvers toward the Pareto front of six measured quantities, so that one run returns a front of trade-offs rather than a single tuned solver.
The system runs on small open-weight language models served on a single laptop, and we compare it with a single agent on four problems. 
For three-dimensional acoustic scattering and elasticity, it delivers 47 and 48 convergent hybrid solvers out of 60 candidates. 
Relative to the Jacobi-preconditioned Krylov baselines, the best of the discovered solvers 
reduce the iteration counts by factors of 4.7 and 60 and the online solve time by factors of 2.3 and 3.0, whereas the single agent produces no executable plan. 
For coupled dynamic thermoelasticity in two and three dimensions, where the baselines fail, it completes 98 of 100 transient candidates and, 
compared with the single agent, 
reaches a given accuracy 15 times faster in two dimensions 
and a displacement error 20 times lower in three dimensions. 
\end{abstract}

\section*{Introduction}

The numerical solution of partial differential equations~(PDEs) requires the solution of large linear systems, and in many simulation pipelines this step dominates the computational cost. 
Krylov subspace methods such as conjugate gradients, GMRES, and BiCGStab~\cite{hestenes1952methods,saad1986gmres,vandervorst1992bicgstab,saad2003iterative} are standard tools, and their performance depends on the choice of preconditioner. Multigrid~\cite{trottenberg2001multigrid,ruge1987algebraic,vanek1996algebraic}, domain decomposition~\cite{toselli2005domain,xu1992iterative}, and incomplete factorizations~\cite{benzi2002preconditioning} provide classical building blocks, but assembling them into an efficient solver for a given operator remains a task that requires expert knowledge. 
For example, the indefinite Helmholtz operator at high wave numbers defeats many standard preconditioners unless a complex shift is  introduced~\cite{erlangga2004class,ernst2012why,gander2019class}.
Algebraic multigrid can coarsen elasticity operators poorly unless their rigid-body modes are explicitly supplied to the hierarchy~\cite{vanek1996algebraic}. Coupled multiphysics operators, in turn, often require block factorizations together with suitable approximations of the Schur complement~\cite{benzi2005numerical,elman2014finite,brown2012composable}.
Although each of these design choices is well documented, constructing an effective solver for a new operator is still largely a process of trial and error, repeated from one problem to another.
This gap between the availability of solver components and their effective composition was also studied~\cite{lee2025automatic}.

Neural operators~\cite{lu2021learning,li2021fourier,kovachki2023neural} provide an additional class of components for iterative solvers.
Due to spectral bias~\cite{rahaman2019spectral}, a trained operator typically captures the low frequency components of the error more accurately than the high frequency ones, making it complementary to relaxation methods and incomplete factorizations, which tend to have the opposite behavior.
Hybrid solvers based on this observation have used neural operators as smoothers, coarse-space approximations, or initial guesses~\cite{zhang2022hybrid,kopanivcakova2024deeponet,kahana2023geometry,lee2024fast,azulay2022multigrid,gotz2018machine,greenfeld2019learning,luz2020learning,oh2026spectrally}. In previous work~\cite{lee2024automatic,lee2025automatic,cao2026automatic}, the authors referred to such combinations as meta-solvers, parameterized their construction, evaluated the resulting solver configurations, and selected among the Pareto-optimal candidates using a preference function. This approach, however, still requires the design space to be specified manually for each equation and relies on an exhaustive evaluation in which many candidate solvers are ultimately dominated. Earlier work on automatic solver selection instead trained classifiers on matrix features to select a method from a fixed set~\cite{bhowmick2006application,eller2012dynamic}, but did not address the composition of new solver configurations. The automatic composition of solvers by evolution was introduced by EvoStencils~\cite{schmitt2021evostencils}, which builds geometric multigrid methods from a formal grammar, generates their implementations, measures their performance, and selects them by non-dominated sorting within grammar-guided genetic programming. 

Large language models~(LLMs) have recently been incorporated into systems for writing and evaluating scientific code, including program search~\cite{romera2024mathematical,novikov2025alphaevolve}, agent-based systems for chemistry~\cite{boiko2023autonomous,bran2024augmenting} and research workflows~\cite{lu2024ai}, and numerical problem solving~\cite{toscano2025athena,toscano2026graft,li2025codepde}. Multi-agent debate has also been reported to improve the factual accuracy of model reasoning~\cite{du2023improving}. Among the works most closely related to ours, AgenticSciML~\cite{jiang2026agenticsciml} organizes proposer, critic, engineer, and analyst agents around structured debate, a knowledge base, a memory of previous attempts, and evolutionary parent selection. 
It applies this framework to the design of physics-informed neural networks and neural operators, and evaluates the contributions of the multi-agent structure and knowledge base through single-agent and knowledge-base ablations. AutoNumerics~\cite{du2026autonumerics} uses a team of agents to construct and verify complete numerical solvers for PDEs from plain-language descriptions. 
Neither, however, focuses on the design of preconditioned linear solvers under measured multi-objective criteria, where each proposed solver is implemented in a production library, evaluated on the target discretization, and compared using measured computational cost and accuracy.

\begin{figure}[!tbp]
	\centering
	\includegraphics[width=\textwidth]{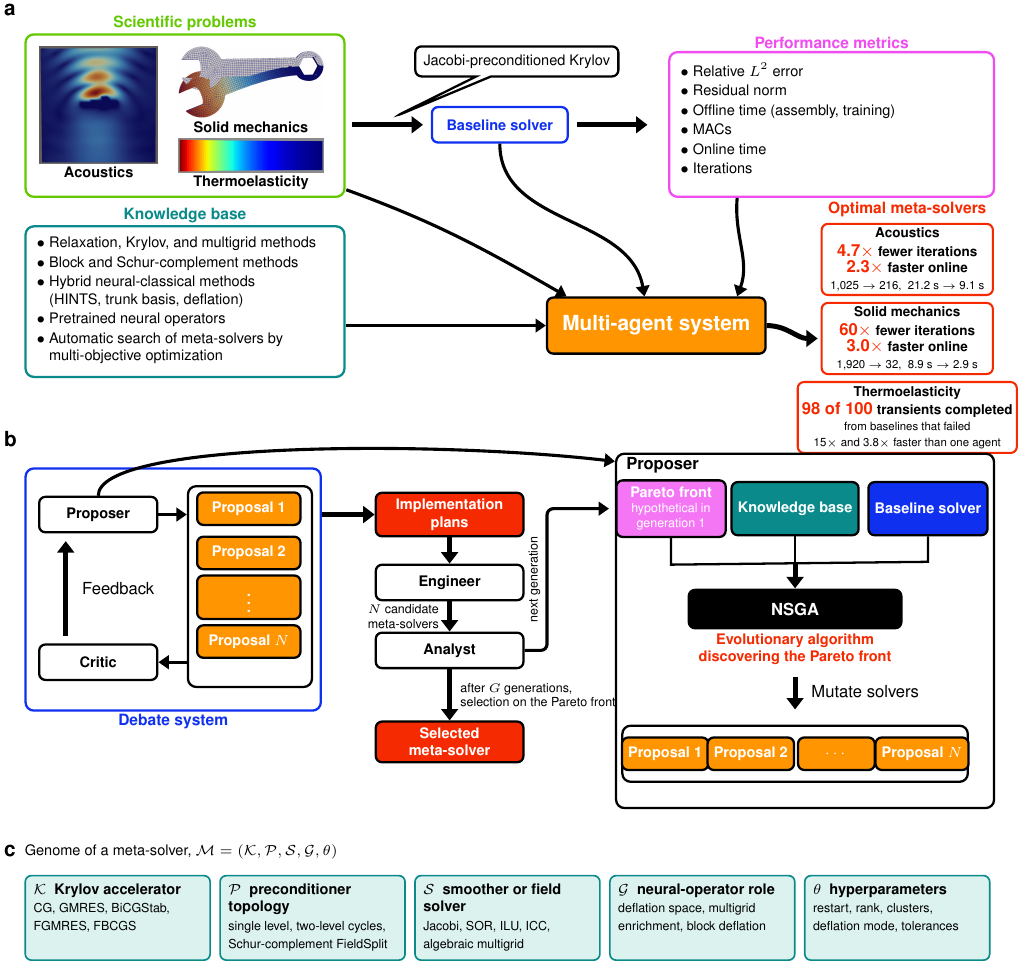}
	\caption{
		\textbf{Multi-agent discovery of meta-solvers.}
		\textbf{a}, Overview. The inputs are a scientific problem specified by a plain-language description and a mesh, a baseline solver given by a Krylov method with a Jacobi preconditioner, a knowledge base of numerical methods including pretrained neural operators, and a fixed suite of performance metrics. The multi-agent system returns a meta-solver selected from the Pareto front of the evaluated candidates. The thumbnails illustrate the three problem classes considered in this paper, and the cards give the gains measured with open-weight language models.
		\textbf{b}, Workflow of one generation. In the debate system, the Proposer generates $N$ proposals and the Critic provides feedback over $m$ rounds. The Engineer implements the resulting plans and runs every candidate on the target problem, while the Analyst measures the metrics for all $N$ candidates. The measured Pareto front is then passed to the Proposer in the next generation, and after $G$ generations, one solver is selected from the final front. Within the Proposer, a non-dominated sorting genetic algorithm (NSGA) evolves the population using the baseline solver, the knowledge base, and the current Pareto front, which is hypothetical in the first generation. The resulting mutated solvers form the proposals for the next generation.
		\textbf{c}, Genome of a meta-solver, defined in \cref{eq:genome}, together with the alleles available at each locus.
		}
	\label{fig:framework}
\end{figure}

In this paper, we introduce a multi-agent system that carries out this solver-design process autonomously (see, \cref{fig:framework}). Four agents, a Proposer, a Critic, an Engineer, and an Analyst, cooperate over several generations. The Proposer maintains a population of solver genomes and evolves it using a non-dominated sorting genetic algorithm (NSGA)~\cite{deb2002fast}, with variation operators generated through reasoning rather than random perturbation. The search is multi-objective from the outset. The population is organized into non-domination fronts according to six measured quantities, with crowding distance used to maintain diversity along each front. As a result, a single run produces a family of trade-off solutions rather than a solver optimized for a single scalar objective, which would recover only one point on the Pareto front and may fail to capture its nonconvex regions~\cite{miettinen1999nonlinear,manoj2026singularities}.

This formulation replaces the exhaustive enumeration used in our earlier meta-solver work~\cite{lee2024automatic,lee2024fast,lee2025automatic} with an open-ended search and distinguishes the system from evolutionary agent frameworks driven by a single objective.
Each candidate solver is implemented using PETSc~\cite{dalcin2011parallel}, FEniCSx~\cite{BarattaEtal2023}, and PyTorch~\cite{paszke2019pytorch}, executed on the target discretization, and evaluated using a common set of measured metrics.
A second question addressed in this paper is what the team of agents contributes beyond a single agent, and at which cost.
To answer it under the least favorable conditions for the agents, we run the whole system on small open-weight language models~\cite{yang2025qwen3,openai2025gptoss} served locally on one laptop, so that a discovery has no inference cost and can be repeated by anyone, and we repeat every discovery with a single agent in place of the team.
We applied both systems to time-harmonic acoustic scattering at high wave number, three-dimensional linear elasticity on an unstructured mesh, and two-way coupled dynamic thermoelasticity in two and three dimensions. The agent roles, the debate protocol, the genome grammar, the six metrics, the NSGA selection, and the structure of the knowledge base were identical in all four runs.
For the two single-physics problems, the multi-agent system produced families of convergent hybrid solvers that improved substantially over the baselines, with different neural roles selected for the two operators, while the single agent did not produce a plan that could be executed.
For the coupled problems, whose baselines fail, the multi-agent system produced solvers that complete the transient, with multigrid block preconditioners that the single agent did not find. Its candidates reach the same accuracy level 15 times faster in two dimensions, meet the temperature target 3.8 times faster in three dimensions, and are more accurate in the displacement, which remains the field that neither system resolves to the prescribed target.
The price of the debate is 2.4 to 2.7 times the planning tokens of the single agent.

\section*{Results}

\subsection*{A multi-agent system that evolves a population of meta-solvers}
In this section, we concern meta-solvers for solving the following linear system induced by discretized scientific problems:
\begin{equation*}
    A u = f,
\end{equation*}
where $A$ is a discretized differential operator, $u$ is a discretized solution, and $f$ belongs to the range of $A$.
A preconditioner $M^{-1}$, an inexpensive approximation of the action of $A^{-1}$, enters in the right-preconditioned form used throughout this paper,
\begin{equation*}
    AM^{-1} y = f, \qquad u = M^{-1}y,
\end{equation*}
so the Krylov method builds its search directions from residuals of the preconditioned operator $AM^{-1}$, whose spectrum governs the iteration count (see Methods for the flexible variants in which $M^{-1}$ changes between iterations).
The system receives four inputs (\cref{fig:framework}a).
The problem is specified by a plain-language description of the governing equations, boundary conditions, and discretization, together with a mesh file. A baseline solver provides a functional but simple starting point. In our experiments, this is a Krylov method with a Jacobi or block-Jacobi preconditioner. A knowledge base of text documents describes relaxation, Krylov, multigrid, block and Schur-complement, and hybrid neural–classical methods, and makes pretrained neural operators available through interface specifications and saved weights.
The output is a set of mutually non-dominated candidates, which we call $\alpha$-solvers, forming a Pareto front from which a solver can be selected according to the user's preferences.

A candidate meta-solver is encoded as a genome (\cref{fig:framework}c),
\begin{equation}
	\label{eq:genome}
	\M=\big(\K,\ \Pc,\ \Sm,\ \G,\ \theta\big),
\end{equation}
whose loci specify the Krylov accelerator $\K$, the preconditioner topology $\Pc$, the smoother or field solver $\Sm$, the role of the neural operator $\G$, and the numerical hyperparameters $\theta$.
The preconditioner topology may range from a single-level method to a two-level cycle and, for coupled problems, to Schur-complement FieldSplit factorizations. The neural operator may supply a deflation space, enrich the hierarchy of an algebraic multigrid method, or deflate the field blocks of a coupled system. In the additive setting, a two-level cycle takes the form
\begin{equation}
	\label{eq:twolevel}
	M^{-1}=\Sm+Q\,A_c^{-1}Q^{\top},\qquad A_c=Q^{\top}AQ,
\end{equation}
where the columns of $Q$ span a coarse space that may, for example, be supplied by the trunk basis of a DeepONet~\cite{kopanivcakova2024deeponet}.
Because the loci are symbolic, the genome space consists of compositions of these primitives rather than a fixed set of configurations that can be enumerated in advance.
In the runs reported here every candidate is required to contain a neural component, so the search is over hybrid solvers only.

Each candidate is evaluated by the objective vector
\begin{equation}
	\label{eq:moo}
	\J(\M)=\big(\mathrm{Iterations}(\M),\ \mathrm{MAC}(\M),\ \mathrm{Online\ time}(\M),\ \mathrm{Error}(\M),\ \mathrm{Residual}(\M),\ \mathrm{Offline\ time}(\M)\big),
\end{equation}
whose components are the iteration count, the measured multiply-accumulate~(MAC), defined in Methods, the wall-clock solve time, the relative $L^2$ error with respect to a reference solution, the final residual, and the setup time. The search minimizes $\J$ in the Pareto sense~\cite{censor1977pareto,miettinen1999nonlinear} over all six components. We report the first four together with the setup time, which contains the construction of the neural basis, while the final residual is used by the agents primarily as a consistency check during the debate.

The discovery process proceeds in generations (\cref{fig:framework}b). In each generation, the Proposer produces $N=20$ genomes. The Critic examines every proposal over $m=2$ rounds and requests corrections when a combination is inconsistent, for example when a symmetric solver is paired with a nonsymmetric operator or a restarted accelerator is given no restart length. The Proposer then produces an implementation plan. A deterministic validator checks the plan against the interface of the solver library and returns it to the Proposer for repair, at most twice, when a configuration is not admissible. A generation whose plan is still rejected after the repairs has no candidates. The Engineer turns each admissible plan into self-contained PETSc and FEniCSx scripts, executes all candidates, and repairs runtime failures using execution traces. The Analyst extracts the resulting metrics and appends every outcome, including failed runs, to a methodological memory.

Before the next generation, the archive is sorted into non-domination fronts over all six objectives, and each candidate is assigned a crowding distance. The Proposer reads the resulting ranked table and selects parents front by front, breaking ties in favor of larger crowding distance, so that the best candidate for any individual objective is retained. It then generates offspring by crossing and mutating the loci of the selected parents. Each modification is justified using properties of the operator and information from previously recorded failures rather than by random perturbation. Each offspring is also assigned to a niche in the objective space, fewest iterations, lowest arithmetic cost, lowest error, or lowest setup cost, and the Critic checks whether the proposed design is consistent with that niche. We run three generations for each problem, so that the budget of every run is sixty candidates.
The Proposer is Qwen3-30B~\cite{yang2025qwen3}, the Critic is gpt-oss-20b~\cite{openai2025gptoss}, and the Engineer is Qwen3-Coder-30B~\cite{qwen3coder}, all served locally.
In the single-agent comparison, the Critic and its debate rounds are removed and one Qwen3-Coder model plans and engineers every generation, with the same problem statement, knowledge base, neural operators, validator, repair attempts, baseline, and metrics. The agents, their prompts, and the knowledge base are described in Methods.

\subsection*{Acoustic scattering at high wave number}

\begin{figure}[!tbp]
	\centering
	\includegraphics[width=\textwidth]{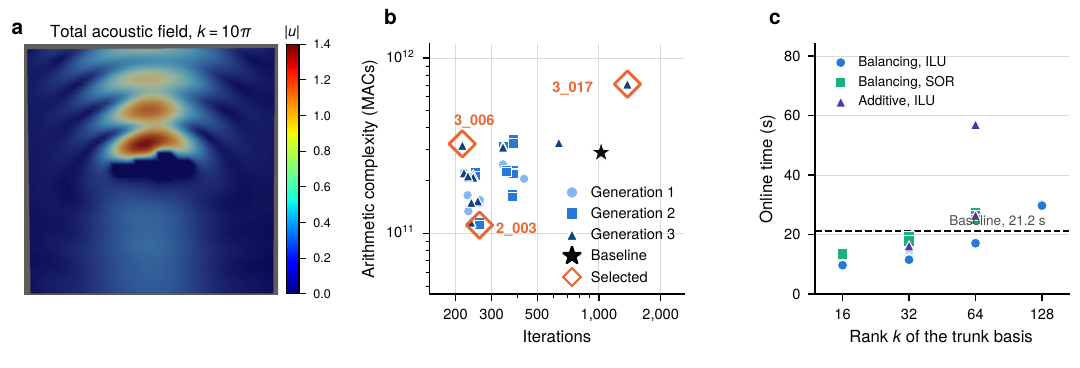}
	\caption{
		\textbf{Acoustic scattering for the Helmholtz equation at $k=10\pi$.}
		\textbf{a}, Magnitude of the total acoustic field for a Gaussian beam scattered by a sound-soft obstacle inside the unit cube, computed using the MUMPS direct reference solution.
		\textbf{b}, Converged candidates of the multi-agent system from the three generations shown in the plane of iteration count versus measured arithmetic complexity. The black star denotes the GMRES(100), GMRES restarting at every $100$ iterations, baseline with Jacobi preconditioning.
        The orange diamonds denote the candidate with the lowest arithmetic complexity and online time, \cand{2\_003}, the candidate with the fewest iterations, \cand{3\_006}, and the candidate with the lowest error, \cand{3\_017}.
		\textbf{c}, Online time of the converged candidates against the rank $k$ of the trunk basis used for deflation, for the three families of the population. The dashed line is the online time of the baseline. The single agent executed no candidate for this problem.
	}
	\label{fig:helmholtz}
\end{figure}
The first problem is the time-harmonic scattering of a Gaussian beam by a sound-soft obstacle inside the unit cube, governed by the Helmholtz equation with wave number $k=10\pi$ and absorbing boundary conditions on the outer faces~\cite{ihlenburg1998finite,colton2013inverse} (\cref{fig:helmholtz}a). First-order hexahedral elements on a mesh with $h=1/64$ yield a complex indefinite system with 274,492 unknowns. The baseline, GMRES(100), GMRES restarting at each $100$ iterations, with a Jacobi preconditioner, converges in 1,025 iterations and 21.2~s, with a relative $L^2$ error of $2.3\times10^{-8}$ against a MUMPS direct solution~\cite{MUMPS:1,MUMPS:2}. The main difficulty comes from the indefiniteness of the operator. Positive-definite preconditioners and standard multigrid methods can fail or diverge, while the iteration count of diagonally preconditioned Krylov methods increases with $k$~\cite{ernst2012why}.

\Cref{tab:single} lists representative candidates, and \cref{fig:helmholtz}b shows the converged candidates in the plane of iteration count versus arithmetic complexity. Two neural roles are available for this problem, both built from the trunk basis of a DeepONet~\cite{kopanivcakova2024deeponet}, the enrichment of an algebraic multigrid hierarchy and the deflation of the Krylov method (Methods). In the first generation, the Proposer assigned nine candidates to the first role and eleven to the second. A one-application acceptance test rejected the nine multigrid candidates with residual ratios between $10^{2}$ and $10^{36}$, while all eleven deflation candidates converged. The Proposer of the second generation recorded this outcome and placed every candidate in the deflation family (Supplementary Note~2), and the Critic asked for restart lengths and cluster counts that had not been tried. Sixteen candidates of the second generation and all twenty of the third converged.

The arithmetic and time champion, \cand{2\_003}, uses FGMRES(20) with a balancing deflation of rank 16 around an incomplete LU factorization. It requires 262 iterations, $1.11\times10^{11}$ MAC, and 9.1~s, factors of 3.9, 2.6, and 2.3 below the baseline. The construction of the neural basis adds 6.9~s of offline time, and with it the solver still finishes in 16.0~s against 22.9~s for the baseline. The iteration champion, \cand{3\_006}, raises the rank to 128 and requires 216 iterations, a factor of 4.7 below the baseline. The accuracy champion, \cand{3\_017}, uses the additive variant with two clusters and reaches an error of $8.5\times10^{-9}$, 2.7 times below that of the baseline. \Cref{fig:helmholtz}c explains the shape of the front. A larger basis removes few iterations, 238 at rank 16 against 216 at rank 128 with restart 50, but every application becomes more expensive, so the online time grows from about 9~s to 30~s. The agents found this trade-off within three generations and kept both of its ends. In total 47 of the 60 candidates converged, each with an error at most seven percent above that of the baseline, 45 of them need fewer iterations than the baseline, and 35 are faster in online time.

The single agent did not reach the execution stage on this problem. In the first two generations its planner declared that no executable hybrid configuration existed, and in the third its plan omitted a required field that two repair requests did not restore, so no candidate was run (\cref{tab:ablation}).

\subsection*{Linear elasticity on an unstructured mesh}

\begin{figure}[!tbp]
	\centering
	\includegraphics[width=\textwidth]{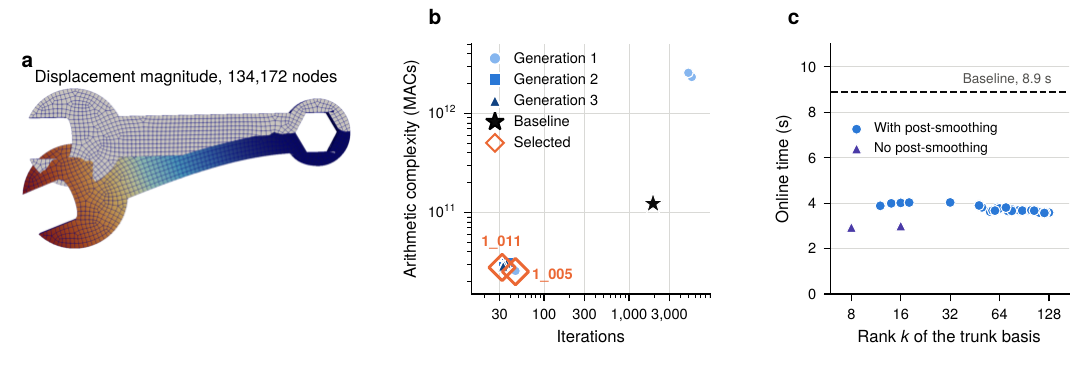}
	\caption{
		\textbf{Linear elasticity on an unstructured hexahedral mesh.}
		\textbf{a}, Displacement magnitude of the loaded component on the deformed configuration, with the undeformed mesh shown in white, computed using the MUMPS direct reference solution.
		\textbf{b}, Converged candidates of the multi-agent system from the three generations shown in the plane of iteration count versus measured arithmetic complexity. The black star denotes the conjugate gradient~(CG) baseline with Jacobi preconditioning. The orange diamonds denote \cand{1\_005}, which has the lowest arithmetic complexity and online time, and \cand{1\_011}, which has the fewest iterations. The two points at the upper right are deflation candidates of the first generation.
		\textbf{c}, Online time of the converged multigrid candidates against the rank $k$ of the trunk basis that enriches the hierarchy. The dashed line is the online time of the baseline. The single agent executed no candidate for this problem.
		}
	\label{fig:elasticity}
\end{figure}

The second problem is the three-dimensional linear elasticity for a mechanical component discretized using first-order vector elements on an unstructured hexahedral mesh with 134,172 nodes (\cref{fig:elasticity}a).
The material parameters are $E=210$ and $\nu=0.3$.
The resulting symmetric positive-definite system has 402,516 unknowns and 29.5 million nonzero entries.
The baseline solver, conjugate gradients with a Jacobi preconditioner, requires 1,920 iterations and 8.9~s.

The agents followed a different route for this problem (\cref{tab:single}, \cref{fig:elasticity}b). The deflation that was effective for the Helmholtz problem gave no improvement here. Three of the four deflation candidates of the first generation converged, but they needed more than 5,000 iterations and more than 200~s, and the fourth diverged. By contrast, smoothed-aggregation algebraic multigrid~\cite{vanek1996algebraic,balay2019petsc} enriched with the basis induced by Transolver~\cite{wu2024transolver} converged in 32 to 46 iterations for five of the six candidates whose levels were smoothed by SOR, while the ten candidates that smoothed with Jacobi were rejected by the acceptance test. Candidate \cand{1\_005}, with eight basis vectors and no post-smoothing, converges in 46 iterations, 2.9~s, and $2.52\times10^{10}$ MAC, and candidate \cand{1\_011}, with 128 basis vectors, converges in 32 iterations. Relative to the baseline, these results correspond to reductions by factors of 60 in iteration count, 3.0 in online time, and 4.9 in arithmetic complexity.

In the second generation, the Proposer read this evidence from the archive, discarded both the deflation role and the Jacobi smoother, and varied the rank of the basis between 12 and 128 (Supplementary Note~2). All forty candidates of the second and third generations converged, in 32 to 39 iterations and 3.6 to 4.0~s (\cref{fig:elasticity}c). The lowest error of the run, $2.3\times10^{-10}$, about half that of the baseline, belongs to \cand{3\_003}. In total 48 of the 60 candidates converged and 45 of them are faster than the baseline in online time.

The gains refer to the online time. The construction of the neural basis and of the multigrid hierarchy takes 12 to 15~s of offline time, so the hybrid solver pays off when the operator is solved more than once, as in a sequence of load cases. Multigrid is the standard approach for elasticity~\cite{vanek1996algebraic}, so this run also serves as a control showing that the system discards an unsuitable neural role and recovers the standard expert design from a Jacobi baseline within one generation. The single agent proposed plans in all three generations, but each was rejected for a configuration that the solver library does not admit, and none could be repaired within the allowed attempts (\cref{tab:ablation}).

\begin{table}[!tbp]
\centering
\caption{
	\textbf{Representative meta-solvers discovered by the multi-agent system for the two single-physics classes.} Candidates are labeled by generation and index. Iteration count, online time, offline time, and arithmetic complexity refer to one solve. Error is the relative $L^2$ error with respect to the MUMPS direct reference, and the selected candidates are shown in bold.
    TB denotes the trunk basis of the neural operator, of rank $k$, and GAMG denotes smoothed-aggregation algebraic multigrid.
    GMRES($m$) denotes the GMRES restarting at each $m$ iterations.
    Furthermore, FGMRES denotes the flexible variant of GMRES.
    The construction of the neural basis is charged to the offline time. All candidates use the baseline residual tolerance of $10^{-8}$. Complete results for all sixty candidates in each class are provided in Supplementary Tables~\ref{tab:si_helm} and~\ref{tab:si_elas}.
	}
\label{tab:single}
\footnotesize
\setlength{\tabcolsep}{3pt}
\begin{tabularx}{\textwidth}{@{}l>{\raggedright\arraybackslash}Xrrrcc@{}}
\toprule
ID & Meta-solver & Iterations & Online (s) & Offline (s) & MAC & Rel.\ $L^2$ error \\
\midrule
\multicolumn{7}{@{}l}{\emph{Acoustic scattering, Helmholtz equation at $k=10\pi$ (274,492 complex unknowns)}} \\ \midrule
0\_000 & GMRES(100) with Jacobi, baseline & 1,025 & 21.2 & 1.7 & $2.88\times10^{11}$ & $2.25\times10^{-8}$ \\
1\_001 & FGMRES(50), GAMG with trunk basis $k=16$, SOR, post-smoothing & \multicolumn{5}{c}{rejected by the acceptance test} \\
1\_009 & FGMRES(100), balancing TB deflation $k=16$, ILU & 231 & 10.6 & 6.8 & $1.33\times10^{11}$ & $2.32\times10^{-8}$ \\
1\_012 & FGMRES(100), balancing TB deflation $k=32$, SOR & 341 & 19.4 & 7.5 & $2.47\times10^{11}$ & $2.30\times10^{-8}$ \\
1\_015 & FGMRES(100), additive TB deflation $k=32$, ILU & 432 & 16.2 & 7.7 & $2.05\times10^{11}$ & $1.37\times10^{-8}$ \\
3\_005 & GMRES(50), balancing TB deflation $k=16$, ILU & 238 & 9.5 & 6.7 & $1.17\times10^{11}$ & $2.33\times10^{-8}$ \\
\textbf{2\_003} & \textbf{FGMRES(20), balancing TB deflation $k=16$, ILU} & \textbf{262} & \textbf{9.1} & \textbf{6.9} & \textbf{$1.11\times10^{11}$} & \textbf{$2.34\times10^{-8}$} \\
\textbf{3\_006} & \textbf{GMRES(50), balancing TB deflation $k=128$, ILU} & \textbf{216} & \textbf{30.1} & \textbf{18.0} & \textbf{$3.23\times10^{11}$} & \textbf{$2.34\times10^{-8}$} \\
\textbf{3\_017} & \textbf{FGMRES(50), additive TB deflation $k=64$, 2 clusters, ILU} & \textbf{1,377} & \textbf{57.0} & \textbf{10.3} & \textbf{$7.09\times10^{11}$} & \textbf{$8.46\times10^{-9}$} \\
\addlinespace[3pt]
\multicolumn{7}{@{}l}{\emph{Linear elasticity on an unstructured hexahedral mesh (402,516 unknowns)}} \\ \midrule
0\_000 & CG with Jacobi, baseline & 1,920 & 8.9 & 2.8 & $1.23\times10^{11}$ & $4.41\times10^{-10}$ \\
1\_002 & FGMRES(100), balancing TB deflation $k=32$, ILU & 5,498 & 207.6 & 10.3 & $2.33\times10^{12}$ & $7.29\times10^{-8}$ \\
1\_013 & FGMRES(50), GAMG with trunk basis $k=16$, Jacobi, post-smoothing & \multicolumn{5}{c}{rejected by the acceptance test} \\
1\_007 & FGMRES(50), GAMG with trunk basis $k=32$, SOR, post-smoothing & 38 & 4.0 & 12.6 & $3.18\times10^{10}$ & $5.47\times10^{-9}$ \\
2\_017 & FGMRES(50), GAMG with trunk basis $k=12$, SOR, post-smoothing & 38 & 3.9 & 12.0 & $3.10\times10^{10}$ & $1.83\times10^{-9}$ \\
2\_014 & FGMRES(50), GAMG with trunk basis $k=120$, SOR, post-smoothing & 32 & 3.6 & 14.6 & $2.78\times10^{10}$ & $6.28\times10^{-9}$ \\
\textbf{1\_005} & \textbf{FGMRES(50), GAMG with trunk basis $k=8$, SOR, no post-smoothing} & \textbf{46} & \textbf{2.9} & \textbf{11.8} & \textbf{$2.52\times10^{10}$} & \textbf{$1.00\times10^{-9}$} \\
\textbf{1\_011} & \textbf{FGMRES(50), GAMG with trunk basis $k=128$, SOR, post-smoothing} & \textbf{32} & \textbf{3.6} & \textbf{14.5} & \textbf{$2.78\times10^{10}$} & \textbf{$1.65\times10^{-9}$} \\
\textbf{3\_003} & \textbf{FGMRES(50), GAMG with trunk basis $k=66$, SOR, post-smoothing} & \textbf{34} & \textbf{3.8} & \textbf{13.9} & \textbf{$2.94\times10^{10}$} & \textbf{$2.26\times10^{-10}$} \\
\bottomrule
\end{tabularx}
\end{table}

\subsection*{Coupled dynamic thermoelasticity in two dimensions}

\begin{figure}[!tbp]
	\centering
	\includegraphics[width=\textwidth]{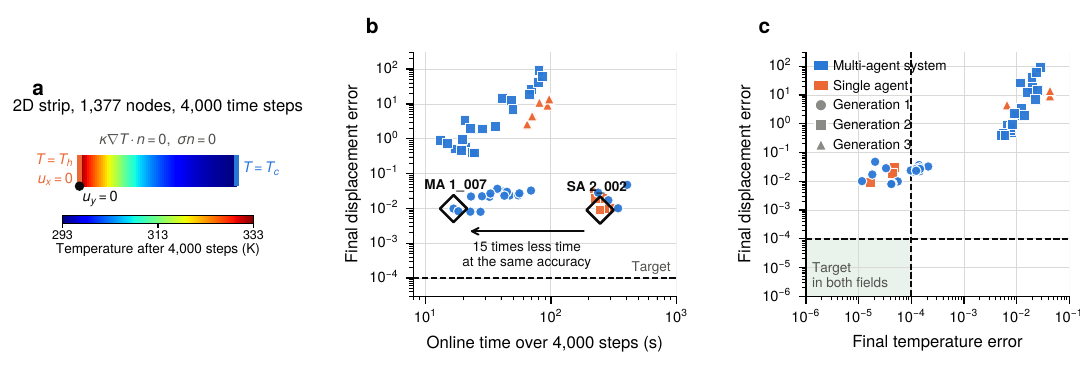}
	\caption{
		\textbf{Two-way coupled dynamic thermoelasticity in two dimensions.}
		\textbf{a}, The strip with its boundary conditions and the temperature after 4,000 time steps. The left edge is heated and clamped axially, the right edge is cooled, the origin is pinned, and the remaining boundary is insulated and traction-free.
		\textbf{b}, Completed candidates of the multi-agent system and of the single agent in the plane of online time over the 4,000 steps versus final relative error of the displacement. The baseline diverged and is not shown. The black diamonds mark the fastest candidate of each system at the common accuracy level, \cand{1\_007} of the multi-agent system and \cand{2\_002} of the single agent. Several candidates of the single agent coincide.
		\textbf{c}, The same candidates in the plane of final temperature error versus final displacement error. The dashed lines are the target of $10^{-4}$ in each field.
		}
	\label{fig:thermo2d}
\end{figure}

For given computational spatial domain $\Omega$ and the final time $t_{f}$, the third and fourth problems consider transient, two-way coupled thermoelasticity, in which the temperature $T$ and displacement $\uv$ of a linear thermoelastic body satisfy~\cite{nowacki1986thermoelasticity,boley1960theory}
\begin{equation}
	\label{eq:thermo}
\left\{
\begin{aligned}
\rho c_{\varepsilon} \frac{\partial T}{\partial t} &= \nabla \cdot (\kappa \nabla T) + Q - \beta T_{0} \tr\left(\frac{\partial 
\varepsilon(\uv)}{\partial t}\right),\\
\nabla \cdot \sigma\left(\uv\right) & = \rho \frac{\partial^{2}\uv}{\partial t^{2}},
\end{aligned}
\right.
\mathrm{in}\ \Omega \times (0, t_{f}],
\end{equation}
where
\begin{equation*}
\left\{
\begin{aligned}
\varepsilon (\uv)  & = \frac{1}{2} \left( \nabla \uv + {\left(\nabla \uv\right)}^{T}\right),  \\
\varepsilon^{\mathrm{th}} &= \alpha \left(T-T_{0}\right) \mathbf{I}, \\
\sigma(\uv) &= \mathbf{C} : \left( \varepsilon - \varepsilon^{\mathrm{th}} \right).
\end{aligned}
\right.
\end{equation*}
The coupling modulus is $\beta=3K\alpha$, with $K$ the bulk modulus and $\alpha$ the thermal expansion coefficient.
The last term in the energy balance represents the thermoelastic coupling through which the rate of dilatation feeds back into the temperature.
The material is an aluminium-like benchmark whose parameters are given in Supplementary Note~1.
One end of the body is held at $T_h=333.15$~K and clamped in the axial direction, while the opposite end is held at $T_c=293.15$~K. The remaining boundary is insulated and traction-free, and the body starts from rest at $T_0=293.15$~K.

The Crank-Nicolson step~\cite{crank1947practical} for the temperature and the trapezoidal Newmark step~\cite{newmark1959method} for the momentum balance, with $\Delta t=1/400$ over 4,000 steps up to $t_f=10$, yield at each step the block system
\begin{equation}
	\label{eq:block}
	\begin{bmatrix} K_{TT} & K_{Tu}\\ K_{uT} & K_{uu}\end{bmatrix}
	\begin{bmatrix} \xv_T \\ \xv_u\end{bmatrix}=
	\begin{bmatrix} \bv_T \\ \bv_u\end{bmatrix}.
\end{equation}
Here, $K_{TT}$ is the parabolic temperature block, $K_{uu}$ combines the mass and stiffness contributions of the displacement equation, and the off-diagonal blocks represent the two-way coupling. This coupling makes a block-diagonal preconditioner inadequate. The solver is called 4,000 times in each run, and all metrics are accumulated over the full transient. 

The two-dimensional problem is a strip of length $0.1$ and height $0.02$, with 1,377 nodes and 4,131 unknowns at each time step (\cref{fig:thermo2d}a). The baseline uses BiCGStab with a Schur-complement FieldSplit and Jacobi preconditioning on both fields, and it fails to converge.
A reference trajectory, computed with the same block structure and MUMPS factorizations of both fields formed once and reused at every step, is used to define the accuracy target. We measure the final relative $L^2$ error of the temperature and of the displacement separately, and we regard a candidate as accurate when both are at most $10^{-4}$. This is stricter than a criterion on the combined field, which is dominated by the temperature because the temperature is of order 300~K while the displacement is of order $10^{-5}$~m (see Methods). One neural role is available for the coupled problems, the deflation of one or both field blocks with a basis built from a multi-field Transolver.

Starting from the failed baseline, all twenty candidates of the first generation of the multi-agent system completed the 4,000 steps (\cref{tab:multi}, \cref{fig:thermo2d}b). During the debate the Critic had pointed out that the first proposal applied conjugate gradients to the nonsymmetric block system, and the final plan used FGMRES around a full Schur factorization in every candidate (Supplementary Note~2). The Proposer spread the candidates over the block preconditioner, with incomplete LU, Jacobi, or algebraic multigrid on the two blocks, over the deflated field, and over the rank and the clustering of the basis. Nine candidates meet the temperature target, with errors down to $1.2\times10^{-5}$. The four candidates with multigrid on both blocks need the fewest iterations and are the most accurate in the displacement. Candidate \cand{1\_007} needs 40,453 cumulative iterations, about ten per step, and 16.8~s, against 1.4 million iterations and 238~s for the same deflation around Jacobi blocks in \cand{1\_005}, and its displacement error is $9.9\times10^{-3}$. The lowest displacement error, $8.0\times10^{-3}$, belongs to \cand{1\_020}. The second generation replaced FGMRES by GMRES. Its candidates are faster still, down to 13.2~s, but none of them meets the temperature target, and the archive keeps this as negative evidence. The plan of the third generation was rejected by the validator for incomplete block options.

The single agent had its first plan rejected and executed two generations. It kept Jacobi on both blocks in all forty candidates, and its two generations contain only six and five distinct results. The twenty candidates of its second generation meet the temperature target, and the best of them, \cand{2\_002}, reaches a displacement error of $9.0\times10^{-3}$ after 1.8 million iterations and 247~s. Its third generation made the same change to GMRES and lost the temperature accuracy in the same way.

The two systems therefore reach the same accuracy level, and they differ in the cost of reaching it. At a temperature error below $10^{-4}$ and a displacement error near $10^{-2}$, the solver found by the multi-agent system needs 15 times less online time, 46 times fewer iterations, and 16 times less arithmetic than the solver found by the single agent (\cref{fig:thermo2d}b). No candidate of either system meets the target in the displacement (\cref{fig:thermo2d}c), which we discuss after the three-dimensional results.

\subsection*{Dynamic thermoelasticity in three dimensions}

\begin{figure}[!tbp]
	\centering
	\includegraphics[width=\textwidth]{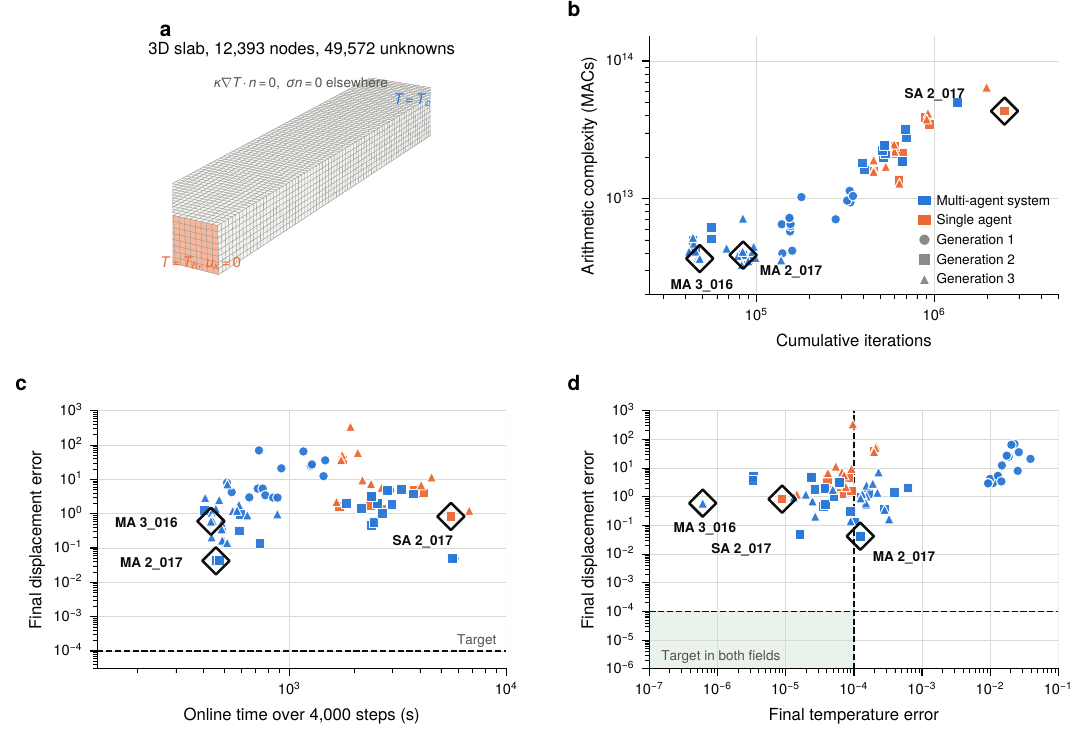}
	\caption{
		\textbf{Two-way coupled dynamic thermoelasticity in three dimensions.}
		\textbf{a}, The slab with its hexahedral mesh. The face at $x=0$ is heated and clamped axially, the face at $x=0.1$ is cooled, and the remaining boundary is insulated and traction-free.
		\textbf{b}, Completed candidates of the multi-agent system and of the single agent in the plane of cumulative iterations versus measured arithmetic complexity over 4,000 steps. The baseline did not finish and is not shown. The black diamonds mark \cand{2\_017} of the multi-agent system, with the lowest displacement error, \cand{3\_016} of the multi-agent system, with the lowest temperature error, and \cand{2\_017} of the single agent, its most accurate candidate.
		\textbf{c}, The same candidates in the plane of online time versus final relative error of the displacement.
		\textbf{d}, The same candidates in the plane of final temperature error versus final displacement error. The dashed lines are the target of $10^{-4}$ in each field.
		}
	\label{fig:thermo3d}
\end{figure}

The three-dimensional problem is a slab of size $0.1$ by $0.02$ by $0.01$ with the same material, boundary conditions, and time stepping, discretized on a hexahedral mesh with 12,393 nodes. Each time step therefore requires solving a coupled system with 49,572 unknowns (\cref{fig:thermo3d}a). The baseline, FGMRES with a block-Jacobi Schur FieldSplit, does not finish within its time limit. A reference trajectory, computed with a full Schur FieldSplit and MUMPS factorizations of both blocks formed once and reused, defines the accuracy target as in two dimensions.

The multi-agent system executed all three generations, and 58 of its 60 candidates completed the transient (\cref{tab:multi}, \cref{fig:thermo3d}b). In the first generation, the most accurate candidate, \cand{1\_016}, FGMRES with a rank-10 deflation of the displacement block around incomplete LU factorizations, reaches a temperature error of $1.6\times10^{-5}$ and a displacement error of $5.0\times10^{-2}$, at the price of 1.35 million iterations and 5,777~s. The fifteen candidates of this generation that used GMRES returned temperature errors near $10^{-2}$, while the three completed candidates under FGMRES met the temperature target, and the Proposer of the second generation kept FGMRES in all twenty candidates. The second generation kept this parent and mutated the clustering of the basis. With 32 clusters and both blocks deflated, \cand{2\_017} reaches the lowest displacement error of the run, $4.2\times10^{-2}$, in 84,159 iterations and 459~s, that is, 16 times fewer iterations and 13 times less time than \cand{1\_016}, with a temperature error of $1.2\times10^{-4}$. The number of candidates that meet the temperature target rose from three to twelve. The third generation crossed the clustered basis with multigrid on the displacement block. All twenty of its candidates completed in 409 to 881~s, and \cand{3\_016} reaches the lowest temperature error, $6.1\times10^{-7}$, in 48,000 iterations, twelve per step, and 435~s (\cref{fig:thermo3d}c).

The single agent had its first plan rejected, and 19 of the 40 candidates of its two executed generations completed. Eleven did not converge in the first time step and ten exceeded the time limits. As in two dimensions, every candidate uses Jacobi on both blocks. Fifteen candidates meet the temperature target, the fastest in 1,651~s, and the lowest displacement error is 0.83.

Among the candidates that meet the temperature target, the fastest solver of the multi-agent system is therefore 3.8 times faster than that of the single agent, and the lowest displacement error is 17 times smaller, or 20 times smaller among all completed candidates (\cref{fig:thermo3d}c,d). The displacement remains the open part of the coupled problems. No candidate of either system reaches $10^{-4}$ in this field in two or three dimensions, and tightening the residual tolerance to $10^{-9}$ lowers the temperature error to $3\times10^{-6}$ and leaves the displacement error above one (\cand{2\_011}, \cand{2\_013}). We attribute this to the stopping test, which acts on the residual of the combined system, where the temperature rows dominate, so that a test on each field is the first change to make.

\begin{table}[!tbp]
\centering
\caption{
	\textbf{Representative meta-solvers for the two coupled thermoelastic instances.} MA and SA denote the multi-agent system and the single agent. Iteration count, online time, and arithmetic complexity are accumulated over the 4,000 time steps. The errors are the final relative $L^2$ errors of the temperature $T$ and of the displacement $\uv$ with respect to the MUMPS Schur-FieldSplit reference, and the selected candidates are shown in bold. In each candidate description, the Schur factorization type is followed by the preconditioner of the temperature and displacement blocks, by the deflated field, and by the rank $k$ of the neural basis. The residual tolerance is $10^{-8}$ unless stated. Complete results are given in Supplementary Tables~\ref{tab:si_t2} to~\ref{tab:si_t3_sa}.
	}
\label{tab:multi}
\footnotesize
\setlength{\tabcolsep}{3pt}
\begin{tabularx}{\textwidth}{@{}l>{\raggedright\arraybackslash}Xrrccc@{}}
\toprule
ID & Meta-solver & Iterations & Online (s) & MAC & Error in $T$ & Error in $\uv$ \\
\midrule
\multicolumn{7}{@{}l}{\emph{Two-dimensional dynamic thermoelasticity (4,131 unknowns per step, 4,000 steps)}} \\ \midrule
0\_000 & BiCGStab, Schur FieldSplit with Jacobi on both fields, baseline & \multicolumn{5}{c}{diverged} \\
MA 1\_005 & FGMRES(50), full Schur, Jacobi, balancing deflation of $\uv$, $k=20$ & 1,387,224 & 237.7 & $1.92\times10^{12}$ & $3.3\times10^{-5}$ & $2.8\times10^{-2}$ \\
MA 1\_003 & FGMRES(50), full Schur, ILU, balancing deflation of $T$, $k=20$ & 1,638,060 & 343.2 & $2.34\times10^{12}$ & $1.2\times10^{-5}$ & $1.0\times10^{-2}$ \\
MA 1\_011 & FGMRES(50), full Schur, ILU, balancing deflation of $\uv$, $k=20$, 4 clusters & 117,370 & 23.1 & $1.61\times10^{11}$ & $1.1\times10^{-4}$ & $2.2\times10^{-2}$ \\
\textbf{MA 1\_007} & \textbf{FGMRES(50), full Schur, GAMG, balancing deflation of $\uv$, $k=20$} & \textbf{40,453} & \textbf{16.8} & \textbf{$1.19\times10^{11}$} & \textbf{$5.5\times10^{-5}$} & \textbf{$9.9\times10^{-3}$} \\
\textbf{MA 1\_020} & \textbf{FGMRES(50), full Schur, GAMG, balancing deflation of $T$, $k=40$} & \textbf{40,228} & \textbf{27.5} & \textbf{$2.12\times10^{11}$} & \textbf{$4.2\times10^{-5}$} & \textbf{$8.0\times10^{-3}$} \\
MA 2\_011 & GMRES, full Schur, GAMG, balancing deflation of $\uv$, $k=20$, 4 clusters & 32,000 & 13.2 & $1.04\times10^{11}$ & $6.8\times10^{-3}$ & 0.91 \\
SA 2\_006 & FGMRES, full Schur, Jacobi, additive deflation of $\uv$, $k=32$ & 1,801,198 & 223.2 & $1.73\times10^{12}$ & $4.9\times10^{-5}$ & $3.0\times10^{-2}$ \\
\textbf{SA 2\_002} & \textbf{FGMRES, full Schur, Jacobi, additive deflation of $\uv$, $k=64$} & \textbf{1,848,704} & \textbf{246.7} & \textbf{$1.94\times10^{12}$} & \textbf{$1.7\times10^{-5}$} & \textbf{$9.0\times10^{-3}$} \\
SA 3\_012 & GMRES, full Schur, Jacobi, additive deflation of $\uv$, $k=32$, tolerance $10^{-9}$ & 514,370 & 64.5 & $4.97\times10^{11}$ & $1.0\times10^{-2}$ & 2.63 \\
\addlinespace[3pt]
\multicolumn{7}{@{}l}{\emph{Three-dimensional dynamic thermoelasticity (49,572 unknowns per step, 4,000 steps)}} \\ \midrule
0\_000 & FGMRES, Schur FieldSplit with Jacobi on both fields, baseline & \multicolumn{5}{c}{not finished within the time limit} \\
MA 1\_014 & GMRES, full Schur, ILU, additive deflation of $T$ and $\uv$, $k=20$ & 139,970 & 515.8 & $4.00\times10^{12}$ & $2.5\times10^{-2}$ & 8.07 \\
MA 1\_016 & FGMRES, full Schur, ILU, balancing deflation of $\uv$, $k=10$ & 1,354,075 & 5,777.1 & $4.97\times10^{13}$ & $1.6\times10^{-5}$ & $5.0\times10^{-2}$ \\
MA 2\_005 & FGMRES, full Schur, GAMG, balancing deflation of $\uv$, $k=20$ & 55,850 & 588.1 & $5.07\times10^{12}$ & $9.0\times10^{-5}$ & 0.32 \\
MA 2\_011 & FGMRES, full Schur, ILU, balancing deflation of $\uv$, $k=20$, tolerance $10^{-9}$ & 702,214 & 3,292.6 & $2.78\times10^{13}$ & $3.4\times10^{-6}$ & 5.19 \\
MA 2\_016 & FGMRES, full Schur, ILU, balancing deflation of $\uv$, $k=20$, 32 clusters & 89,872 & 404.9 & $3.51\times10^{12}$ & $1.6\times10^{-4}$ & 1.27 \\
\textbf{MA 2\_017} & \textbf{FGMRES, full Schur, ILU, balancing deflation of $T$ and $\uv$, $k=20$, 32 clusters} & \textbf{84,159} & \textbf{458.7} & \textbf{$3.88\times10^{12}$} & \textbf{$1.2\times10^{-4}$} & \textbf{$4.2\times10^{-2}$} \\
\textbf{MA 3\_016} & \textbf{FGMRES, full Schur, ILU and GAMG, balancing deflation of $\uv$, $k=15$, 64 clusters} & \textbf{48,000} & \textbf{434.9} & \textbf{$3.66\times10^{12}$} & \textbf{$6.1\times10^{-7}$} & \textbf{0.60} \\
MA 3\_018 & FGMRES, full Schur, ICC, additive deflation of $\uv$, $k=20$, 32 clusters & 137,756 & 516.7 & $3.58\times10^{12}$ & $9.0\times10^{-5}$ & 0.14 \\
SA 3\_008 & FGMRES, full Schur, Jacobi, additive deflation of $\uv$, $k=24$, 4 clusters & 637,471 & 1,650.6 & $1.29\times10^{13}$ & $7.6\times10^{-5}$ & 2.25 \\
\textbf{SA 2\_017} & \textbf{FGMRES, upper Schur, Jacobi, additive deflation of $\uv$, $k=16$} & \textbf{2,487,948} & \textbf{5,559.0} & \textbf{$4.32\times10^{13}$} & \textbf{$8.9\times10^{-6}$} & \textbf{0.83} \\
\bottomrule
\end{tabularx}
\end{table}

\begin{table}[!tbp]
\centering
\caption{
	\textbf{Multi-agent system and single agent on the four problems.} Both systems use the same problem statement, knowledge base, neural operators, solver library, validator, baseline, metrics, and budget of three generations of twenty candidates. A generation is marked as rejected when its plan did not pass the validator after two repairs, in which case no candidate was executed. For the thermoelastic problems, a candidate is completed when it ran the 4,000 steps and returned all metrics, and the temperature target is a final relative error at or below $10^{-4}$. Planning covers the Proposer and, for the multi-agent system, the Critic, and engineering covers the implementation, the repairs, and the analysis of the results.
	}
\label{tab:ablation}
\footnotesize
\setlength{\tabcolsep}{4pt}
\begin{tabularx}{\textwidth}{@{}Xcc@{}}
\toprule
 & Multi-agent system & Single agent \\
\midrule
\multicolumn{3}{@{}l}{\emph{Acoustic scattering, Helmholtz equation}} \\ \midrule
Executed candidates, generations 1, 2, 3 & 20, 20, 20 & rejected, rejected, rejected \\
Converged candidates, generations 1, 2, 3 & 11, 16, 20 & 0, 0, 0 \\
Shortest online time (s) & 9.15 (\cand{2\_003}) & --- \\
Candidates faster than the baseline online & 35 of 47 & --- \\
Plan repairs requested by the validator & 3 & 6 \\
Planning tokens, input and output (million) & 0.78 and 0.14 & 0.33 and 0.03 \\
Engineering tokens (million) & 11.72 & 0 \\
Model time, planning and engineering (min) & 61 and 30 & 15 and 0 \\
\addlinespace[3pt]
\multicolumn{3}{@{}l}{\emph{Linear elasticity}} \\ \midrule
Executed candidates, generations 1, 2, 3 & 20, 20, 20 & rejected, rejected, rejected \\
Converged candidates, generations 1, 2, 3 & 8, 20, 20 & 0, 0, 0 \\
Shortest online time (s) & 2.93 (\cand{1\_005}) & --- \\
Candidates faster than the baseline online & 45 of 48 & --- \\
Plan repairs requested by the validator & 4 & 6 \\
Planning tokens, input and output (million) & 0.81 and 0.14 & 0.34 and 0.05 \\
Engineering tokens (million) & 7.41 & 0 \\
Model time, planning and engineering (min) & 60 and 16 & 27 and 0 \\
\addlinespace[3pt]
\multicolumn{3}{@{}l}{\emph{Dynamic thermoelasticity in two dimensions}} \\ \midrule
Executed candidates, generations 1, 2, 3 & 20, 20, rejected & rejected, 20, 20 \\
Completed candidates & 40 of 40 & 40 of 40 \\
Candidates meeting the target in both fields & 0 & 0 \\
Candidates meeting the temperature target & 9 & 20 \\
Distinct results among completed candidates & 40 & 11 \\
Lowest displacement error & $8.0\times10^{-3}$ (\cand{1\_020}) & $9.0\times10^{-3}$ (\cand{2\_002}) \\
Shortest online time at the temperature target (s) & 16.8 (\cand{1\_007}) & 223.2 (\cand{2\_006}) \\
Wall-clock time of the benchmark jobs (h) & 0.8 & 2.2 \\
Plan repairs requested by the validator & 6 & 6 \\
Planning tokens, input and output (million) & 1.02 and 0.17 & 0.40 and 0.05 \\
Engineering tokens (million) & 7.34 & 6.47 \\
Model time, planning and engineering (min) & 89 and 21 & 29 and 17 \\
\addlinespace[3pt]
\multicolumn{3}{@{}l}{\emph{Dynamic thermoelasticity in three dimensions}} \\ \midrule
Executed candidates, generations 1, 2, 3 & 20, 20, 20 & rejected, 20, 20 \\
Completed candidates & 58 of 60 & 19 of 40 \\
Candidates meeting the target in both fields & 0 & 0 \\
Candidates meeting the temperature target & 23 & 15 \\
Distinct results among completed candidates & 52 & 17 \\
Lowest displacement error & $4.2\times10^{-2}$ (\cand{2\_018}) & 0.83 (\cand{2\_017}) \\
Shortest online time at the temperature target (s) & 434.9 (\cand{3\_016}) & 1,650.6 (\cand{3\_008}) \\
Wall-clock time of the benchmark jobs (h) & 24.2 & 28.6 \\
Plan repairs requested by the validator & 5 & 5 \\
Planning tokens, input and output (million) & 0.97 and 0.16 & 0.36 and 0.07 \\
Engineering tokens (million) & 11.08 & 4.61 \\
Model time, planning and engineering (min) & 87 and 31 & 40 and 12 \\
\bottomrule
\end{tabularx}
\end{table}

\subsection*{Multi-agent system versus single agent}

\begin{figure}[!tbp]
\centering
\includegraphics[width=\textwidth]{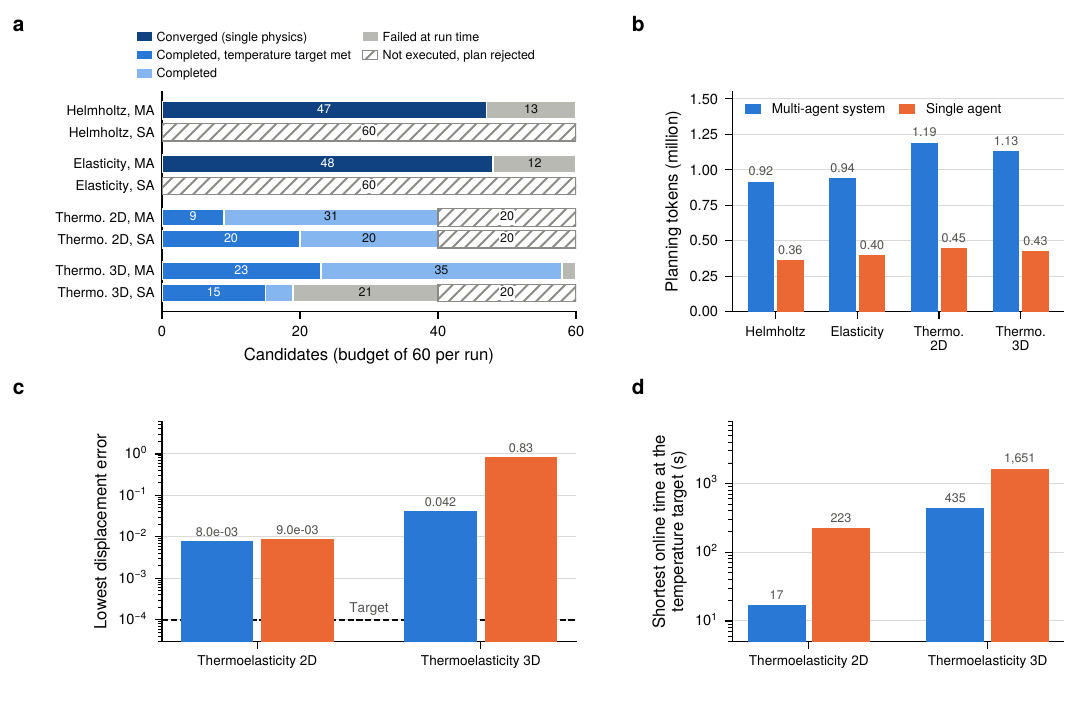}
\caption{
	\textbf{Multi-agent system (MA) and single agent (SA) on the four problems.}
	\textbf{a}, Outcome of the budget of sixty candidates of every run. A generation whose plan was rejected by the validator contributes twenty candidates that were not executed.
	\textbf{b}, Planning tokens, input and output, summed over the three generations.
	\textbf{c}, Lowest final displacement error among the completed candidates of the thermoelastic problems. The dashed line is the target.
	\textbf{d}, Shortest online time among the thermoelastic candidates that meet the temperature target.
    }
\label{fig:mavssa}
\end{figure}

\Cref{tab:ablation} and \cref{fig:mavssa} collect the comparison over the four problems. The first difference is whether a generation can be executed at all. The multi-agent system passed the validator in eleven of its twelve generations and the single agent in four of twelve, although both received the same feedback and the same number of repair attempts. Of the 240 candidates of its budget, the multi-agent system executed 220 and completed 193, while the single agent executed 80 and completed 59 (\cref{fig:mavssa}a). The plans of the single agent failed on the interface of the solver library, with a missing field, a Krylov method that a module does not admit, or options in the wrong form, or its planner stopped with the statement that no executable configuration existed. In the multi-agent system such defects are met twice before the validator sees the plan, since the Critic reviews two drafts, and the same Critic removed the error of applying conjugate gradients to a nonsymmetric system.

The second difference is the quality of what is executed. For the two single-physics problems, only the multi-agent system returns solvers, 95 in total, with the gains over the baselines given above. For the two coupled problems, where both systems return candidates, those of the multi-agent system are more diverse, 40 and 52 distinct results against 11 and 17, they reach the common accuracy level 15 times faster in two dimensions, and they are 3.8 times faster and 20 times more accurate in the displacement in three dimensions (\cref{fig:mavssa}c,d). The origin of this difference is visible in the archive. The single agent never left the Jacobi block preconditioner of the baseline, while the Proposer and the Critic explored incomplete factorizations and multigrid on the blocks.

The third difference is the cost. The debate consumes 0.92 to 1.19 million planning tokens per problem, 2.4 to 2.7 times those of the single agent, and 60 to 89 minutes of model time against 15 to 40 minutes (\cref{fig:mavssa}b). For transient problems, however, the cost of a discovery is dominated by the execution of the candidates, not by the language models. In three dimensions the benchmark jobs of the multi-agent system ran for 24.2 hours and returned 58 completed candidates, and those of the single agent ran for 28.6 hours and returned 19, that is, 25 minutes against 90 minutes of computation per completed candidate. In two dimensions the corresponding figures are 1.2 and 3.3 minutes. The additional planning tokens are therefore repaid by a better use of the computing time, and with locally served models they have no monetary cost.

\subsection*{Reliability, transparency, and cost of the discovery}

\begin{figure}[!tbp]
\centering
\includegraphics[width=\textwidth]{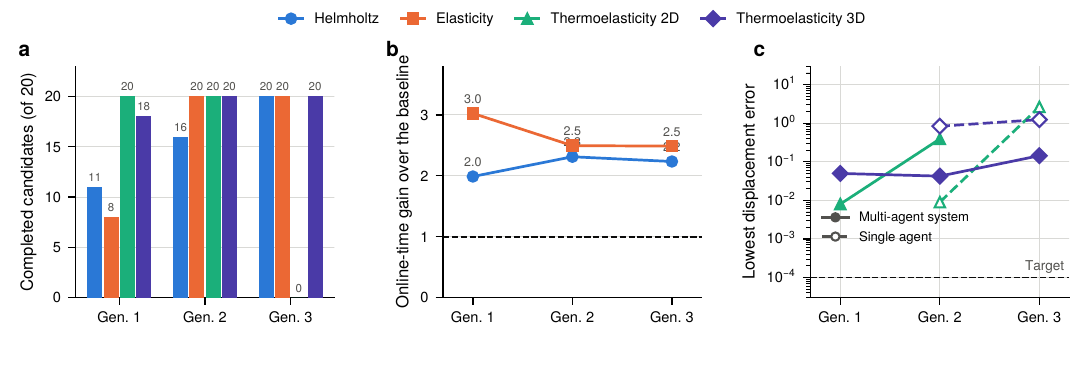}
\caption{
	\textbf{Progress of the discovery across generations for the four problem classes.}
	\textbf{a}, Number of candidates of the multi-agent system that ran to completion among the 20 proposed in each generation. For the Helmholtz and elasticity problems completion means convergence to the prescribed tolerance. The plan of the third generation of the two-dimensional thermoelastic problem was rejected by the validator.
	\textbf{b}, Gain in online time over the Jacobi-preconditioned Krylov baseline of the fastest converged candidate in each generation of the multi-agent system, for the single-physics problems. The dashed line indicates parity with the baseline.
	\textbf{c}, Lowest final displacement error among the completed candidates in each generation for the thermoelastic problems, for the multi-agent system (solid lines) and the single agent (dashed lines). The dashed horizontal line is the target.
    }
\label{fig:reliability}
\end{figure}

Two properties of the discovery process matter beyond the best solver. The first is reliability, defined as the fraction of proposed candidates that run to completion (\cref{fig:reliability}a). For the multi-agent system this fraction increases across generations, from 11, 8, and 18 completed candidates out of 20 in the first generation to 20, 20, and 20 in the third for the Helmholtz, elasticity, and three-dimensional thermoelastic problems. We attribute the increase in reliability to the archive of measured results read by the Proposer before each generation. The gain over the baseline is established early, in the first or second generation (\cref{fig:reliability}b), and the later generations extend the front, since the iteration and accuracy champions of the Helmholtz problem and the accuracy champion of the elasticity problem belong to the third generation. For the three-dimensional thermoelastic problem, the lowest displacement error and the tenfold reduction in time appear in the second generation (\cref{fig:reliability}c).

The second property is transparency. Every candidate is a script that can be re-executed, every metric is measured through library instrumentation, and the debate transcripts record the reasoning behind each design decision. Supplementary Note~2 gives examples, including the Proposer's decision to abandon the multigrid role for the Helmholtz operator after the acceptance test, and the Critic's objection to conjugate gradients for the coupled system.

The cost of discovery contains no fee for language-model inference, because all models are served on the laptop that also runs the candidates. The token counts and model times are listed in Supplementary Note~3. The remaining cost comes from executing the candidates and depends on the problem size, from half an hour for the single-physics problems to one day for the three-dimensional transient.

\section*{Discussion}

The central result is that one agent protocol, with its roles, genome grammar, metrics, and selection rule held fixed and driven by small open-weight language models on a laptop, produced solver families that a numerical analyst would recognize as appropriate for their respective operators. These are Krylov methods with an incomplete factorization and a deflation by the learned basis for the indefinite Helmholtz problem, where multigrid was tried and rejected, smoothed-aggregation multigrid enriched with the learned basis for elasticity, where deflation was tried and rejected, and Schur-complement block preconditioners with multigrid on the blocks and a deflation by the learned basis for the coupled problems. None of these choices was specified in the problem statements. The agents made them from measurements, by assigning the first generation to competing roles and reading the outcome from the archive.

The multi-objective evolution shaped this outcome and is the part of the design we consider new. By maintaining a non-dominated population spread by crowding distance, a single run produced an arithmetic and time champion (\cand{2\_003}) alongside an iteration champion (\cand{3\_006}) and an accuracy champion (\cand{3\_017}) for Helmholtz, a time champion alongside an iteration champion and an accuracy champion for elasticity, and a displacement-accuracy champion (\cand{2\_017}) alongside a temperature-accuracy champion (\cand{3\_016}) for the three-dimensional thermoelastic problem. A single-objective evolution would retain only one member of each group. The niches assigned by the Proposer play a similar role by directing its reasoning toward less explored parts of the front. The final choice remains with the user, through a preference function~\cite{lee2024automatic}, the differential-game exploration developed in related work~\cite{liu2026algorithms}, or continuation along the front, which can reach disconnected parts from a single member~\cite{manoj2026singularities}. The selected solvers are non-dominated among the evaluated candidates, but the search does not certify that better compositions do not exist outside the candidates visited in each run.

The comparison with the single agent shows what the team contributes when the language models are small. A single small model that must plan and satisfy the interface of the solver library at once fails at the interface in two thirds of its generations, and when it succeeds it stays close to the baseline. Separating the roles gives each proposal two reviews before it is validated, and the review is where both the formal defects and the numerical ones, such as a symmetric method for a nonsymmetric system, are removed. The team therefore turns models that cannot complete the task alone into a system that completes it, at 2.4 to 2.7 times the planning tokens and with a more economical use of the computing time spent on the candidates. In our earlier experiments with large proprietary models, a single agent was able to reach solvers comparable to those of the team, so we expect the benefit of the team to be largest in the regime studied here, where the models are small, local, and free of charge.

The closest relatives of this work are automatic meta-solver discovery~\cite{lee2024automatic,lee2025automatic} and AgenticSciML~\cite{jiang2026agenticsciml}.
The former frameworks use the same Pareto view of solver design but enumerate a manually specified design space. AgenticSciML~\cite{jiang2026agenticsciml} uses a similar agent team for physics-informed networks and neural operators, with evolutionary selection driven by a single objective and with single-agent and knowledge-base controls.
The present work applies agents to preconditioned Krylov solvers, where each candidate is an executable composition of library primitives tested on the target discretization.
It replaces both enumeration and a single objective with non-dominated sorting over six measured quantities, includes the block structure of coupled problems, and reports the single-agent control for every problem with models that can be run without a provider.

However, several limitations remain.
For the coupled problems, no candidate meets the accuracy target in the displacement, and the comparison between the two systems rests there on the completed candidates, their errors, and their cost. A stopping test on each field and a scaling of the displacement block are the next steps for these problems.
The gains of the single-physics solvers are gains in online time, and with the offline construction of the neural basis included the best elasticity solver is slower than the baseline for a single solve.
Every candidate was required to contain a neural component, so the runs do not isolate what the neural operator contributes beyond the classical part of each solver.
The single agent plans with the model of the Engineer, so the comparison removes the Critic and changes the planning model together.
The runs are stochastic because the LLMs sample their proposals, and we report one run per system and problem, so the variability between runs is not quantified.
The discretizations are fixed and linear. Extending the approach to nonlinear or adaptively refined problems, where a Newton iteration surrounds the linear solve~\cite{lee2025neural,oh2026spectrally}, is left for future work.
Finally, the knowledge base and the solver library are written by people and limit what the agents can compose. The system automates the search over these compositions, where much of solver design effort is spent, and records each step so that the resulting solver can be checked and ported by a person.

\section*{Methods}

\subsection*{Problem classes and discretizations}

All four problems were discretized with first-order Lagrange finite elements in FEniCSx~\cite{BarattaEtal2023} and assembled into PETSc matrices~\cite{balay2019petsc,dalcin2011parallel}.
The Helmholtz problem is posed on $\Omega=(0,1)^3$ minus a sound-soft scatterer $\Omega_s \subset \Omega$, with 
\begin{equation*}
\left\{
\begin{aligned}
-\Delta u - k^{2} u &= 0, &\text{in } \Omega \setminus \bar{\Omega}_{s},\\
u &=0, &\text{on } \partial\Omega_{s},\\
\frac{\partial u}{\partial n} - iku&=g, &\text{on } \Gamma_{inc},\\
\frac{\partial u}{\partial n} - iku&=0, &\text{on } \Gamma_{abs},
\end{aligned}
\right.
\end{equation*}
where 
$\partial\Omega = \Gamma_{inc} \cup \Gamma_{abs}, \quad \Gamma_{inc} = \{(x,y,z) \colon z=1\}$.
The Gaussian beam $g$ is given by
\begin{equation*}
\left\{
\begin{aligned}
R &= 0.5kw_{0}^{2}, \\
r^{2} &= (x-0.5)^{2}+(y-0.5)^{2}, \\
q &= (z - 0.5) + iR, \\
u_{inc}(x,y,z) &= \frac{iR}{q} e^{-ikz} e^{-ik\frac{r^{2}}{2q}},\\
g(x,y,z)&= \frac{\partial u_{inc}}{\partial n} - iku_{inc}.
\end{aligned}
\right.
\end{equation*}
Here, we set $w_0=0.15$ and $k=10\pi$.
Hexahedral Gauss-Lobatto-Legendre elements with $h=1/64$ and collocated quadrature yield a complex system with 274,492 unknowns.

For given wrench shaped domain $\Omega \subset (0,1)^{3}$, $\partial\Omega = \Gamma_{D} \cup \Gamma_{T} \cup \Gamma_{N}$, $\Gamma_{D}$ denotes the wrench hole and $\Gamma_{T}$ denotes the top of the left jaw of the wrench, the elasticity problem is given by
\begin{equation*}
\left\{
\begin{aligned}
-\nabla \cdot \sigma\left(\uv\right) & = \mathbf{f}, &\text{in } \Omega,\\
\sigma\left(\uv\right) & = \lambda \mathrm{tr}\left(\varepsilon\left(\uv\right)\right)\mathbf{I} + 2 \mu \varepsilon\left(\uv\right),\\
\varepsilon (\uv)  & = \frac{1}{2} ( \nabla \uv + {\left(\nabla \uv\right)}^{T}),  \\
\uv &=\mathbf{0}, &\text{on } \Gamma_{D},\\
\sigma(\uv)\cdot \mathbf{n} &=\mathbf{g}, &\text{on } \Gamma_{T},\\
\sigma(\uv)\cdot \mathbf{n} &=\mathbf{0}, &\text{on } \Gamma_{N},
\end{aligned}
\right.
\end{equation*}
where $\mathbf{f}=\mathbf{0}$ and $\mathbf{g}=\left((0,0,-0.2)\cdot \mathbf{n}\right)\mathbf{n}$.
With $E=210$, $\nu=0.3$, $\lambda = \frac{\nu E}{(1+\nu)(1-2\nu)},$ and $ 
\mu = \frac{E}{2(1+\nu)}$, the system has 402,516 unknowns.

The thermoelastic problems are given by \cref{eq:thermo} on the strip $(0,0.1)\times(0,0.02)$ and the slab $(0,0.1)\times(0,0.02)\times(0,0.01)$, with the material and boundary data stated in Results, mesh size $h=1/800$, time step $\Delta t=2h=1/400$, and 4,000 steps. The temperature equation is advanced by Crank-Nicolson and the momentum balance by the trapezoidal rule. The thermal stress uses the same trapezoidal weights, yielding the block system in \cref{eq:block}, with entries given in Supplementary Note~1. Both problems retain the inertial term and the two-way coupling.

\subsection*{Baseline solvers and reference solutions}

For each problem, the baseline is a functional but simple Krylov solver, prepared before the discovery together with the problem definition, shared by the multi-agent system and the single agent, and executed with the same instrumentation as all later candidates. The baselines are right-preconditioned GMRES with restart 100 and Jacobi preconditioning for Helmholtz, conjugate gradients with Jacobi preconditioning for elasticity, and a global BiCGStab or FGMRES iteration around a PETSc Schur-complement FieldSplit with Jacobi-preconditioned field solves for thermoelasticity. The two thermoelastic baselines fail, the first by divergence and the second by not finishing within its time limit of 1,800~s, so both systems start from a failed baseline for these problems and no gain over the baseline is defined. The stopping criterion is a relative residual tolerance of $10^{-8}$ for the single-physics problems and for most transient candidates, with a tolerance of $10^{-7}$ or $10^{-9}$ in a few transient candidates. The reference solutions for the single-physics problems are MUMPS direct solves~\cite{MUMPS:1,MUMPS:2} of the same assembled systems.
For thermoelasticity, the reference trajectory is computed using the same Schur-FieldSplit topology with MUMPS factorizations of both field blocks, formed once and reused for all 4,000 steps, under FGMRES with a relative tolerance of $10^{-11}$. This trajectory serves only as the accuracy target of the transient classes. Candidates whose outer preconditioner is a complete LU or Cholesky factorization are direct solves and are excluded from the search.

\subsection*{Performance metrics}

Six quantities are recorded for every candidate using definitions fixed in a shared document read by the agents. The error and arithmetic count are
\begin{equation}
	\label{eq:metrics}
	e(\M)=\frac{\norm{u_{\M}-u_{\mathrm{ref}}}_{L^2(\Omega)}}{\norm{u_{\mathrm{ref}}}_{L^2(\Omega)}},\qquad
	\mathrm{MAC}(\M)=\sum_{k=1}^{N_t}\Big(F^{\mathrm{PETSc}}_k + n^{\mathrm{NO}}_k\, C_{\mathrm{NO}}\Big),
\end{equation}
where $u_{\M}$ and $u_{\mathrm{ref}}$ are the finite-element solutions of the candidate and reference, $N_t$ is the number of solves, one for the single-physics problems and 4,000 for the transient problems, $F^{\mathrm{PETSc}}_k$ is the number of floating-point operations recorded by the PETSc counters during solve $k$ and summed over all ranks, $n^{\mathrm{NO}}_k$ is the number of neural forward passes during that solve, and $C_{\mathrm{NO}}$ is the multiply-accumulate count of one forward pass measured by THOP~\cite{thop}.
The error is the $L^2(\Omega)$ norm of the fields, evaluated at the final time for the transient problems.
For thermoelasticity, it is computed for the temperature and for the displacement separately, and the error of a candidate is the larger of the two. The norm of the combined field, $\norm{(T,\uv)}^2=\norm{T}^2+\norm{\uv}^2$, is also recorded. Since the temperature is near 300~K while the displacement is near $10^{-5}$~m, the temperature dominates the combined norm by about seven orders of magnitude, so that a criterion on the combined field does not resolve the displacement accuracy, and we use the criterion on each field. Only the final state is compared, so no trajectory error is available.

The arithmetic count is measured rather than estimated. PETSc counts each floating-point operation once, while the neural part counts one unit per multiply-accumulate, so a neural contribution is undercounted by a factor of two relative to the classical part. We retain the name MAC used in the metric document, although the quantity is the sum defined in \cref{eq:metrics}. In the runs reported here the neural operators are evaluated to build the bases, and this work is charged to the offline time. The residual is $\norm{\bv-A\xv}_2$, recomputed from the converged solution after the final iteration. Offline time includes mesh generation, assembly, solver setup, loading of the neural operator, and construction of the neural basis. Online time covers the solve and excludes warm-up. The iteration count refers to the outer Krylov method. For the transient problems, iteration count, online time, and arithmetic are accumulated over all 4,000 steps.

A candidate of the single-physics problems is successful when it converges to the tolerance and returns the six metrics, and we report in addition whether its error is within ten times that of the baseline. A candidate of the transient problems is completed when it runs the 4,000 steps and returns the metrics, and it is successful when the final error is at most $10^{-4}$ in each field. All timings were taken with one job at a time, on one MPI rank and one thread. The time limits are 1,800~s per job for the single-physics problems and two hours for the transient problems. In three dimensions a job is also stopped when a time step exceeds 120~s or when, after 100 steps, the projected time exceeds 1.5 times the limit.

\subsection*{Genome and discovery space}

The Krylov locus $\K$ ranges over CG, restarted GMRES, BiCGStab, and the flexible variants FGMRES and FBCGS, which admit variable or nonlinear preconditioners~\cite{saad1993flexible}. The topology locus $\Pc$ includes single-level preconditioners, two-level cycles of the form \cref{eq:twolevel} and their deflation variants, multigrid hierarchies, and, for block systems, Schur-complement FieldSplit preconditioners with diagonal, upper, lower, or full factorization.
The smoother locus $\Sm$ includes Jacobi, SOR, ILU and ICC, and smoothed-aggregation algebraic multigrid with PETSc GAMG~\cite{vanek1996algebraic,balay2019petsc}. The neural locus $\G$ includes a deflation space of rank $k$ obtained from the trunk basis of a DeepONet or the internal representation of a Transolver, the enrichment of a multigrid hierarchy with the same basis, and the deflation of one or both field blocks of a coupled system. The hyperparameter locus $\theta$ contains the restart length, the rank $k$, the number of clusters of the basis, the deflation mode, the use of post-smoothing, and the stopping tolerances. For the coupled problems, the smoother locus splits into a temperature-block solver $\Sm_T$ and a displacement-block solver $\Sm_u$, each drawn from the same menu, giving $\M=(\K,\Pc,\Sm_T,\Sm_u,\G,\theta)$. The block preconditioners and neural roles are defined in the next two subsections.
In the runs reported here, the genome is realized through a solver library with one module for each neural role. A candidate is a configuration of one module, written by the Proposer in a structured form, and the validator checks it against the fields and the values that the module admits before it is executed.

\subsection*{Two-level cycles and block preconditioners}

With right preconditioning, the Krylov method is applied to $AM^{-1}y=f$, and the solution is recovered as $u=M^{-1}y$.
Flexible accelerators allow $M^{-1}$ to vary between iterations. For the single-physics problems, the two-level cycles take the additive form in \cref{eq:twolevel} or the multiplicative form
\begin{equation}
	\label{eq:multiplicative}
	M^{-1}r=\Sm r+ Q A_c^{-1} Q^{\top} \left(r - A\Sm r\right),
\end{equation}
where the coarse correction acts on the residual left after one application of the smoother. For the block systems in \cref{eq:block}, write
\begin{equation}
	\label{eq:abcd}
	\Kb=\begin{bmatrix} A & B \\ C & D\end{bmatrix},\qquad A=K_{TT},\quad B=K_{Tu},\quad C=K_{uT},\quad D=K_{uu},\qquad S=D-CA^{-1}B,
\end{equation}
where $S$ is the Schur complement of the temperature block, giving
\begin{equation}
	\label{eq:factor}
	\Kb=\begin{bmatrix} I & 0 \\ CA^{-1} & I\end{bmatrix}\begin{bmatrix} A & 0 \\ 0 & S\end{bmatrix}\begin{bmatrix} I & A^{-1}B \\ 0 & I\end{bmatrix}.
\end{equation}
The inverse of this factorization would be an ideal preconditioner, but $A^{-1}$ and $S^{-1}$ are dense. A Schur-complement FieldSplit preconditioner replaces the action of $A^{-1}$ by the temperature-block solver $\Sm_T$ and that of $S^{-1}$ by the displacement-block solver $\Sm_u$ applied to an approximation $\widehat S$. The topology locus determines which factors are retained. The diagonal type keeps only the middle factor, the lower and upper types retain one triangular factor, and the full type retains all three. The upper type, for example, is
\begin{equation}
	\label{eq:upper}
	M^{-1}=\begin{bmatrix}\widehat A^{-1} & -\widehat A^{-1}B\,\widehat S^{-1} \\ 0 & \widehat S^{-1}\end{bmatrix}.
\end{equation}
In all the transient candidates reported here, the Schur complement is approximated by the displacement block, $\widehat S=D$. The multi-agent system used the full type throughout, and the single agent used the full, upper, and lower types. A block-Jacobi choice of $\widehat A^{-1}$ and $\widehat S^{-1}$ takes no account of the spectrum of the stiff displacement block, which is consistent with the failure of the thermoelastic baselines and with the cost of the candidates that keep Jacobi on both blocks.

\subsection*{Role of the neural operator}

A DeepONet represents its output as
\begin{equation}
	\label{eq:trunk}
	\G(r)(\xv)=\sum_{j=1}^{k}\beta_j(r)\tau_j(\xv),
\end{equation}
where the branch coefficients $\beta_j$ depend on the input and the trunk functions $\tau_j$ form a fixed learned basis.
Sampling the trunk functions at the mesh nodes, zeroing the Dirichlet rows, and orthonormalizing the columns gives the matrix $Q\in\R^{n\times k}$ in \cref{eq:twolevel}, which we call the trunk basis (TB).
The coarse operator $A_c=Q^{\top}AQ$ is formed once and factored.
For Transolver, the internal node representation before the output head is used in the same way, and for the transient problems the basis is built from the multi-field Transolver for the temperature block, the displacement block, or both.
The basis may also be split over $c$ clusters of the mesh nodes, which localizes its columns.
The basis is built before the solve, and its construction is charged to the offline time.

The solver library exposes three roles for this basis. In the deflation role, used for the Helmholtz problem, the coarse correction $QA_c^{-1}Q^{\top}$ is combined with a single-level preconditioner $\Sm$, either additively as in \cref{eq:twolevel} or in the balancing form~\cite{tang2009comparison}
\begin{equation}
	\label{eq:balancing}
	M^{-1}=\big(I-QA_c^{-1}Q^{\top}A\big)\,\Sm\,\big(I-AQA_c^{-1}Q^{\top}\big)+QA_c^{-1}Q^{\top},
\end{equation}
which removes the span of $Q$ from the part of the residual seen by $\Sm$.
In the multigrid role, used for the elasticity problem, the columns of $Q$ enrich a smoothed-aggregation hierarchy, whose levels are smoothed by Jacobi or SOR, with or without post-smoothing, under a flexible outer method. In the block-deflation role, used for the coupled problems, the deflation is applied inside the Schur-complement FieldSplit to the solver of one or both field blocks. Before a candidate of the first two roles is run, an acceptance test applies its preconditioner once and rejects the candidate when the residual grows by more than a factor of ten. Such a rejection is recorded in the archive as an experimental outcome and is not repaired.

\subsection*{Agents and protocol}

All language models are open-weight models served locally by Ollama~\cite{ollama} with a context window of 131,072 tokens, one request at a time. In the multi-agent system, the Proposer uses Qwen3-30B~\cite{yang2025qwen3}, the Critic uses gpt-oss-20b~\cite{openai2025gptoss}, and the Engineer and the Analyst use Qwen3-Coder-30B~\cite{qwen3coder}. Each generation begins with two rounds in which the Proposer presents $N=20$ candidates with their parents, crossovers, mutations, and expected niches.
The Critic returns a verdict on each candidate, after which the Proposer issues a final plan.
The Proposer receives the problem statement, metric definitions, neural-operator interfaces, the current Pareto state of the archive with front indices and crowding distances, and the full results history.
A deterministic validator then checks the final plan against the interface of the solver library. When it finds an inadmissible configuration, it returns the list of defects to the Proposer, which may submit a corrected plan up to two times. If the plan is still rejected, or if the Proposer reports that it cannot produce an executable plan, the generation has no candidates.
The Engineer instantiates each candidate of an admissible plan as a self-contained script, runs the twenty candidates one at a time, records their outputs, and repairs candidates that fail at runtime from their traces.
Candidates that still fail are recorded together with the evidence.
The Analyst combines deterministic parsing of the execution logs with a model-written observation for each candidate, while the orchestrator computes the non-dominated sorting and crowding distances of the archive before each generation. Every prompt, response, plan, script, and metric file is retained.
In the single-agent comparison, one Qwen3-Coder-30B model writes the final plan of each generation directly, with no Critic and no debate rounds, and the same model engineers and analyses the candidates. It reads the same problem statement, knowledge base, neural-operator interfaces, and archive of measured results, and it is subject to the same validator, the same two repair attempts, the same baseline, and the same limits. The single agent therefore differs from the multi-agent system in two respects, the absence of the Critic and the model that plans. Token counts and model times of both systems are reported in Supplementary Note~3, with planning and engineering counted separately.

\subsection*{NSGA-driven proposal}

The selection follows the non-dominated sorting genetic algorithm~\cite{deb2002fast}.
A candidate $\M_a$ dominates $\M_b$, written $\M_a\prec\M_b$, when
\begin{equation}
	\label{eq:dominance}
	\J_i(\M_a)\le\J_i(\M_b)\ \text{ for all } i=1,\dots,6\qquad\text{and}\qquad \J_j(\M_a)<\J_j(\M_b)\ \text{ for some } j,
\end{equation}
and the archive $\mathcal{A}$ of all evaluated candidates is partitioned into fronts
\begin{equation}
	\label{eq:fronts}
	\begin{aligned}
		\F_1&=\{\M\in\mathcal{A}:\ \text{no }\M'\in\mathcal{A}\text{ satisfies }\M'\prec\M\},\\
		\F_{k+1}&=\{\M\in\mathcal{A}_k:\ \text{no }\M'\in\mathcal{A}_k\text{ satisfies }\M'\prec\M\},\qquad
		\mathcal{A}_k=\mathcal{A}\setminus(\F_1\cup\dots\cup\F_k).
	\end{aligned}
\end{equation}
The sorting uses all six components of \cref{eq:moo}, so the first front can be broad, and candidates within it are ordered by crowding distance. For each objective, the candidates are sorted and the crowding distance of $\M$ is
\begin{equation}
	\label{eq:crowding}
	d(\M)=\sum_{i=1}^{6}
	\frac{\J_i(\M_i^{+})-\J_i(\M_i^{-})}
	{\max_{\mathcal{A}}\J_i-\min_{\mathcal{A}}\J_i},
\end{equation}
where $\M^{+}_i$ and $\M^{-}_i$ are the neighbours of $\M$ in the ordering by objective $i$. The two extreme candidates for each objective receive infinite distance. Parents are selected front by front, with ties broken in favor of larger crowding distance, so that the best candidate for each individual objective is retained and the population remains spread along the front rather than collapsing to one region. Failed candidates remain in the archive with the values they reached, or with penalty values when none are available, so the sorting also records unsuccessful designs. In the first generation, the archive contains only the baseline, and the Proposer combines it with archetypes from the knowledge base, such as a deflation by the trunk basis or a multigrid hierarchy enriched with it. In later generations, crossover recombines loci from two parents, while mutations replace an allele at one locus, such as the smoother, the accelerator, the rank or the clustering of the basis, or the block preconditioner. The Proposer is asked to distribute offspring across niches of the objective space, including fewest iterations, lowest complexity, lowest error, and cheapest setup, and to state the expected niche of each candidate for the Critic to check.

\subsection*{Knowledge base and neural operators} 

The knowledge base consists of Markdown documents on classical relaxation and Krylov methods, domain decomposition, and multigrid~\cite{saad2003iterative,brenner2008mathematical}, hybrid neural-classical solvers including HINTS~\cite{zhang2022hybrid}, trunk-basis coarse spaces~\cite{kopanivcakova2024deeponet}, and deflation~\cite{frank2001construction,tang2009comparison}, meta-solver construction and multi-objective selection~\cite{lee2024automatic,lee2025automatic}, the complex shifted-Laplacian preconditioner~\cite{erlangga2004class}, block and Schur-complement preconditioning~\cite{benzi2005numerical,brown2012composable}, and the NSPOD construction~\cite{levrero2026nspod}. In the runs reported here, the knowledge base also contains a capability document of the solver library, which states for each module the admissible fields and values.
The pretrained operators are provided through interface specifications, implementation files, and saved weights. For Helmholtz, the operator is a DeepONet~\cite{lu2021learning} with 128 basis functions whose branch network takes the boundary data and wave number and whose trunk network takes the coordinates, together with three variants using Kolmogorov-Arnold trunk networks with Legendre, Chebyshev, or Jacobi bases. For elasticity, the operator is a Transolver~\cite{wu2024transolver,luo2025transolver} with hidden dimension 128, 32 physics slices, 8 attention heads, and depth 5, mapping the load and coordinates to the displacement. For thermoelasticity, the operator is a multi-field Transolver with two subnetworks of the same size that map the current temperature and displacement, time, and coordinates to the fields at the next step. All operators were trained offline on trajectories from the corresponding problem families and then frozen, so discovery incurs inference cost only. In the runs reported here, each operator enters through the roles of the solver library described above, the deflation and the multigrid enrichment for the single-physics problems and the block deflation for the coupled problems.

\subsection*{Selection on the Pareto front} 

The discovery returns the front $\F_1$ of the final archive. A single solver is selected from this front by applying a preference function $p$ to the rescaled objectives, following our previous work~\cite{lee2024automatic,cao2026automatic}. Any increasing $p$ selects a Pareto-optimal candidate. In this paper, we report candidates selected by simple preferences including the fewest iterations, lowest arithmetic complexity, lowest online time, and lowest error. A finer exploration of the front, including its nonconvex parts, is possible through the differential-game representation developed in our related work~\cite{liu2026algorithms} or by continuation in the hyperparameters $\theta$, which traces the front as a curve and crosses scalarization singularities rather than restarting beyond them~\cite{manoj2026singularities}. The user may also impose constraints such as a memory ceiling or an accuracy floor through the preference function.

\subsection*{Software and hardware}

The candidates were implemented in Python using petsc4py~\cite{dalcin2011parallel}, FEniCSx~\cite{BarattaEtal2023}, MUMPS~\cite{MUMPS:1}, PyTorch~\cite{paszke2019pytorch}, and THOP~\cite{thop}. The language models and all candidates were run on one Apple M5 Max laptop with 18 cores and 64~GB of memory. The timings of the thermoelastic candidates include a load of about two cores from the orchestration process, which is the same for both systems, and seventeen timings affected by system sleep, battery operation, or external load were measured again in isolation, with identical iteration counts, residuals, and errors (Supplementary Note~4).

\subsection*{Use of large language models}

Large language models~(LLM) are components of the method reported here and were used as described above in the Proposer, Critic, Engineer, and Analyst roles.
All numerical results were produced by the generated scripts and were measured by library instrumentation, and all model outputs are archived.  
The text of this manuscript was drafted by the authors with the assistance of a LLM (Claude, Anthropic). 
The authors assume full responsibility for all mathematical content,  statements, computational results, and the final manuscript.

\subsection*{Statistics and reproducibility}

One discovery run of three generations with a budget of twenty candidates per generation is reported for each of the four problems and each of the two systems, eight runs in total. The numbers of completed and successful candidates, iteration counts, arithmetic complexities, and errors are exact measurements from these runs and have no sampling variability, and the timings are single measurements. Language-model sampling of the proposals is the only source of randomness between runs, and its effect was not measured by repeating the runs. Every candidate can be re-executed from its archived script, and complete results for all 300 executed candidates are provided in Supplementary Tables~\ref{tab:si_helm} to~\ref{tab:si_t3_sa}. Two runs on a heterogeneous Poisson problem, made with an earlier configuration of the system, are compared in Supplementary Note~5.

\section*{Data availability}
The problem statements, meshes, reference solutions, generated solver scripts, execution logs, metric files, and debate transcripts of the eight discovery runs, four with the multi-agent system and four with the single agent, and of the two Poisson runs are available from the corresponding author upon reasonable request and will be deposited in a public repository upon publication.

\section*{Code availability}
The orchestration code of the multi-agent system, the knowledge base, and the pretrained neural operators will be made available in a public repository upon publication.

\section*{Acknowledgements}
This work was supported by the DARPA DIAL grant HR00112490484. The computations were carried out on the Oscar cluster of the Center for Computation and Visualization at Brown University.

\section*{Author contributions}
Y.L. and S.L. designed the multi-agent system, carried out the discovery runs, and analyzed the results. J.D. and G.E.K. supervised the research. GEK proposed the original idea. All authors wrote the manuscript.

\section*{Competing interests}
The authors declare no competing interests.

\bibliographystyle{unsrtnat}
\bibliography{references}

@book{saad2003iterative,
  title={Iterative Methods for Sparse Linear Systems},
  author={Saad, Yousef},
  year={2003},
  edition={2},
  publisher={SIAM},
  address={Philadelphia}
}

@article{hestenes1952methods,
  title={Methods of conjugate gradients for solving linear systems},
  author={Hestenes, Magnus R. and Stiefel, Eduard},
  journal={Journal of Research of the National Bureau of Standards},
  volume={49},
  number={6},
  pages={409--436},
  year={1952}
}

@article{saad1986gmres,
  title={{GMRES}: A generalized minimal residual algorithm for solving nonsymmetric linear systems},
  author={Saad, Youcef and Schultz, Martin H.},
  journal={SIAM Journal on Scientific and Statistical Computing},
  volume={7},
  number={3},
  pages={856--869},
  year={1986}
}

@article{vandervorst1992bicgstab,
  title={{Bi-CGSTAB}: A fast and smoothly converging variant of {Bi-CG} for the solution of nonsymmetric linear systems},
  author={van der Vorst, Henk A.},
  journal={SIAM Journal on Scientific and Statistical Computing},
  volume={13},
  number={2},
  pages={631--644},
  year={1992}
}

@article{saad1993flexible,
  title={A flexible inner-outer preconditioned {GMRES} algorithm},
  author={Saad, Youcef},
  journal={SIAM Journal on Scientific Computing},
  volume={14},
  number={2},
  pages={461--469},
  year={1993}
}

@article{benzi2002preconditioning,
  title={Preconditioning techniques for large linear systems: A survey},
  author={Benzi, Michele},
  journal={Journal of Computational Physics},
  volume={182},
  number={2},
  pages={418--477},
  year={2002}
}

@book{trottenberg2001multigrid,
  title={Multigrid},
  author={Trottenberg, Ulrich and Oosterlee, Cornelis W. and Sch{\"u}ller, Anton},
  year={2001},
  publisher={Academic Press},
  address={San Diego}
}

@incollection{ruge1987algebraic,
  title={Algebraic multigrid},
  author={Ruge, John W. and St{\"u}ben, Klaus},
  booktitle={Multigrid Methods},
  editor={McCormick, Stephen F.},
  pages={73--130},
  year={1987},
  publisher={SIAM},
  address={Philadelphia}
}

@article{vanek1996algebraic,
  title={Algebraic multigrid by smoothed aggregation for second and fourth order elliptic problems},
  author={Van{\v{e}}k, Petr and Mandel, Jan and Brezina, Marian},
  journal={Computing},
  volume={56},
  number={3},
  pages={179--196},
  year={1996}
}

@book{toselli2005domain,
  title={Domain Decomposition Methods: Algorithms and Theory},
  author={Toselli, Andrea and Widlund, Olof},
  year={2005},
  publisher={Springer},
  address={Berlin}
}

@article{xu1992iterative,
  title={Iterative methods by space decomposition and subspace correction},
  author={Xu, Jinchao},
  journal={SIAM Review},
  volume={34},
  number={4},
  pages={581--613},
  year={1992}
}

@article{frank2001construction,
  title={On the construction of deflation-based preconditioners},
  author={Frank, Jason and Vuik, Cornelis},
  journal={SIAM Journal on Scientific Computing},
  volume={23},
  number={2},
  pages={442--462},
  year={2001}
}

@article{tang2009comparison,
  title={Comparison of two-level preconditioners derived from deflation, domain decomposition and multigrid methods},
  author={Tang, Jok M. and Nabben, Reinhard and Vuik, Cornelis and Erlangga, Yogi A.},
  journal={Journal of Scientific Computing},
  volume={39},
  number={3},
  pages={340--370},
  year={2009}
}

@article{benzi2005numerical,
  title={Numerical solution of saddle point problems},
  author={Benzi, Michele and Golub, Gene H. and Liesen, J{\"o}rg},
  journal={Acta Numerica},
  volume={14},
  pages={1--137},
  year={2005}
}

@book{elman2014finite,
  title={Finite Elements and Fast Iterative Solvers: With Applications in Incompressible Fluid Dynamics},
  author={Elman, Howard C. and Silvester, David J. and Wathen, Andrew J.},
  year={2014},
  edition={2},
  publisher={Oxford University Press},
  address={Oxford}
}

@inproceedings{brown2012composable,
  title={Composable linear solvers for multiphysics},
  author={Brown, Jed and Knepley, Matthew G. and May, David A. and McInnes, Lois Curfman and Smith, Barry},
  booktitle={2012 11th International Symposium on Parallel and Distributed Computing},
  pages={55--62},
  year={2012},
  organization={IEEE}
}

@article{erlangga2004class,
  title={On a class of preconditioners for solving the {Helmholtz} equation},
  author={Erlangga, Yogi A. and Vuik, Cornelis and Oosterlee, Cornelis W.},
  journal={Applied Numerical Mathematics},
  volume={50},
  number={3--4},
  pages={409--425},
  year={2004}
}

@incollection{ernst2012why,
  title={Why it is difficult to solve {Helmholtz} problems with classical iterative methods},
  author={Ernst, Oliver G. and Gander, Martin J.},
  booktitle={Numerical Analysis of Multiscale Problems},
  editor={Graham, Ivan G. and Hou, Thomas Y. and Lakkis, Omar and Scheichl, Robert},
  pages={325--363},
  year={2012},
  publisher={Springer},
  address={Berlin}
}

@article{gander2019class,
  title={A class of iterative solvers for the {Helmholtz} equation: Factorizations, sweeping preconditioners, source transfer, single layer potentials, polarized traces, and optimized {Schwarz} methods},
  author={Gander, Martin J. and Zhang, Hui},
  journal={SIAM Review},
  volume={61},
  number={1},
  pages={3--76},
  year={2019}
}

@article{MUMPS:1,
  title={A fully asynchronous multifrontal solver using distributed dynamic scheduling},
  author={Amestoy, Patrick R. and Duff, Iain S. and L'Excellent, Jean-Yves and Koster, Jacko},
  journal={SIAM Journal on Matrix Analysis and Applications},
  volume={23},
  number={1},
  pages={15--41},
  year={2001}
}

@article{MUMPS:2,
  title={Performance and scalability of the block low-rank multifrontal factorization on multicore architectures},
  author={Amestoy, Patrick R. and Buttari, Alfredo and L'Excellent, Jean-Yves and Mary, Theo},
  journal={ACM Transactions on Mathematical Software},
  volume={45},
  number={1},
  pages={2:1--2:26},
  year={2019}
}

@techreport{balay2019petsc,
  title={{PETSc} users manual},
  author={Balay, Satish and Abhyankar, Shrirang and Adams, Mark F. and Brown, Jed and Brune, Peter and Buschelman, Kris and Dalcin, Lisandro and Dener, Alp and Eijkhout, Victor and Gropp, William and others},
  institution={Argonne National Laboratory},
  number={ANL-95/11 - Revision 3.12},
  year={2019}
}

@article{dalcin2011parallel,
  title={Parallel distributed computing using {Python}},
  author={Dalcin, Lisandro D. and Paz, Rodrigo R. and Kler, Pablo A. and Cosimo, Alejandro},
  journal={Advances in Water Resources},
  volume={34},
  number={9},
  pages={1124--1139},
  year={2011}
}

@misc{BarattaEtal2023,
  title={{DOLFINx}: the next generation {FEniCS} problem solving environment},
  author={Baratta, Igor A. and Dean, Joseph P. and Dokken, J{\o}rgen S. and Habera, Michal and Hale, Jack S. and Richardson, Chris N. and Rognes, Marie E. and Scroggs, Matthew W. and Sime, Nathan and Wells, Garth N.},
  year={2023},
  howpublished={preprint},
  doi={10.5281/zenodo.10447666}
}

@article{paszke2019pytorch,
  title={{PyTorch}: An imperative style, high-performance deep learning library},
  author={Paszke, Adam and Gross, Sam and Massa, Francisco and Lerer, Adam and Bradbury, James and Chanan, Gregory and Killeen, Trevor and Lin, Zeming and Gimelshein, Natalia and Antiga, Luca and others},
  journal={Advances in Neural Information Processing Systems},
  volume={32},
  pages={8024--8035},
  year={2019}
}

@book{brenner2008mathematical,
  title={The Mathematical Theory of Finite Element Methods},
  author={Brenner, Susanne C. and Scott, L. Ridgway},
  year={2008},
  edition={3},
  publisher={Springer},
  address={New York}
}

@article{newmark1959method,
  title={A method of computation for structural dynamics},
  author={Newmark, Nathan M.},
  journal={Journal of the Engineering Mechanics Division},
  volume={85},
  number={3},
  pages={67--94},
  year={1959}
}

@article{crank1947practical,
  title={A practical method for numerical evaluation of solutions of partial differential equations of the heat-conduction type},
  author={Crank, John and Nicolson, Phyllis},
  journal={Mathematical Proceedings of the Cambridge Philosophical Society},
  volume={43},
  number={1},
  pages={50--67},
  year={1947}
}

@book{ihlenburg1998finite,
  title={Finite Element Analysis of Acoustic Scattering},
  author={Ihlenburg, Frank},
  year={1998},
  publisher={Springer},
  address={New York}
}

@book{colton2013inverse,
  title={Inverse Acoustic and Electromagnetic Scattering Theory},
  author={Colton, David and Kress, Rainer},
  year={2013},
  edition={3},
  publisher={Springer},
  address={New York}
}

@book{nowacki1986thermoelasticity,
  title={Thermoelasticity},
  author={Nowacki, Witold},
  year={1986},
  edition={2},
  publisher={Pergamon Press},
  address={Oxford}
}

@book{boley1960theory,
  title={Theory of Thermal Stresses},
  author={Boley, Bruno A. and Weiner, Jerome H.},
  year={1960},
  publisher={Wiley},
  address={New York}
}

@techreport{bhowmick2006application,
  title={Application of machine learning to the selection of sparse linear solvers},
  author={Bhowmick, Sanjukta and Eijkhout, Victor and Freund, Yoav and Fuentes, Erika and Keyes, David},
  institution={Columbia University},
  year={2006},
  note={Technical report}
}

@inproceedings{eller2012dynamic,
  title={Dynamic linear solver selection for transient simulations using machine learning on distributed systems},
  author={Eller, Paul R. and Cheng, Jing-Ru C. and Maier, Robert S.},
  booktitle={2012 IEEE 26th International Parallel and Distributed Processing Symposium Workshops \& PhD Forum},
  pages={1523--1532},
  year={2012},
  organization={IEEE}
}

@article{lu2021learning,
  title={Learning nonlinear operators via {DeepONet} based on the universal approximation theorem of operators},
  author={Lu, Lu and Jin, Pengzhan and Pang, Guofei and Zhang, Zhongqiang and Karniadakis, George Em},
  journal={Nature Machine Intelligence},
  volume={3},
  number={3},
  pages={218--229},
  year={2021}
}

@inproceedings{li2021fourier,
  title={Fourier neural operator for parametric partial differential equations},
  author={Li, Zongyi and Kovachki, Nikola B. and Azizzadenesheli, Kamyar and Liu, Burigede and Bhattacharya, Kaushik and Stuart, Andrew and Anandkumar, Anima},
  booktitle={International Conference on Learning Representations},
  year={2021}
}

@article{kovachki2023neural,
  title={Neural operator: Learning maps between function spaces with applications to {PDEs}},
  author={Kovachki, Nikola and Li, Zongyi and Liu, Burigede and Azizzadenesheli, Kamyar and Bhattacharya, Kaushik and Stuart, Andrew and Anandkumar, Anima},
  journal={Journal of Machine Learning Research},
  volume={24},
  number={89},
  pages={1--97},
  year={2023}
}

@inproceedings{wu2024transolver,
  title={Transolver: A fast transformer solver for {PDEs} on general geometries},
  author={Wu, Haixu and Luo, Huakun and Wang, Haowen and Wang, Jianmin and Long, Mingsheng},
  booktitle={Proceedings of the 41st International Conference on Machine Learning},
  year={2024}
}

@inproceedings{luo2025transolver,
  title={Transolver++: An accurate neural solver for {PDEs} on million-scale geometries},
  author={Luo, Huakun and Wu, Haixu and Zhou, Hang and Xing, Lanxiang and Di, Yichen and Wang, Jianmin and Long, Mingsheng},
  booktitle={Proceedings of the 42nd International Conference on Machine Learning},
  year={2025}
}

@inproceedings{rahaman2019spectral,
  title={On the spectral bias of neural networks},
  author={Rahaman, Nasim and Baratin, Aristide and Arpit, Devansh and Draxler, Felix and Lin, Min and Hamprecht, Fred and Bengio, Yoshua and Courville, Aaron},
  booktitle={Proceedings of the 36th International Conference on Machine Learning},
  pages={5301--5310},
  year={2019}
}

@article{zhang2022hybrid,
  title={Blending neural operators and relaxation methods in {PDE} numerical solvers},
  author={Zhang, Enrui and Kahana, Adar and Kopani{\v{c}}{\'a}kov{\'a}, Alena and Turkel, Eli and Ranade, Rishikesh and Pathak, Jay and Karniadakis, George Em},
  journal={Nature Machine Intelligence},
  volume={6},
  number={11},
  pages={1303--1313},
  year={2024}
}

@article{kopanivcakova2024deeponet,
  title={{DeepONet} based preconditioning strategies for solving parametric linear systems of equations},
  author={Kopani{\v{c}}{\'a}kov{\'a}, Alena and Karniadakis, George Em},
  journal={SIAM Journal on Scientific Computing},
  volume={47},
  number={1},
  pages={C151--C181},
  year={2025}
}

@article{kahana2023geometry,
  title={On the geometry transferability of the hybrid iterative numerical solver for differential equations},
  author={Kahana, Adar and Zhang, Enrui and Goswami, Somdatta and Karniadakis, George and Ranade, Rishikesh and Pathak, Jay},
  journal={Computational Mechanics},
  volume={72},
  number={3},
  pages={471--484},
  year={2023}
}

@article{lee2024fast,
  title={Fast meta-solvers for {3D} complex-shape scatterers using neural operators trained on a non-scattering problem},
  author={Lee, Youngkyu and Liu, Shanqing and Zou, Zongren and Kahana, Adar and Turkel, Eli and Ranade, Rishikesh and Pathak, Jay and Karniadakis, George Em},
  journal={arXiv preprint arXiv:2405.12380},
  year={2024}
}

@article{lee2024automatic,
  title={Automatic discovery of optimal meta-solvers via multi-objective optimization},
  author={Lee, Youngkyu and Liu, Shanqing and Darbon, J{\'e}r{\^o}me and Karniadakis, George Em},
  journal={Journal of Computational Physics},
  year={2026},
  note={In press, arXiv:2412.00063}
}

@article{lee2025automatic,
  title={Automatic discovery of optimal meta-solvers for time-dependent nonlinear {PDEs}},
  author={Lee, Youngkyu and Liu, Shanqing and Darbon, J{\'e}r{\^o}me and Karniadakis, George Em},
  journal={arXiv preprint arXiv:2507.00278},
  year={2025}
}

@article{lee2025neural,
  title={A neural-operator preconditioned {Newton} method for accelerated nonlinear solvers},
  author={Lee, Youngkyu and Liu, Shanqing and Darbon, J{\'e}r{\^o}me and Karniadakis, George Em},
  journal={arXiv preprint arXiv:2511.08811},
  year={2025}
}

@article{oh2026spectrally,
  title={Spectrally safe neural operator warm-starts for large-scale {Newton} solvers},
  author={Oh, Jaemin and Lee, Youngkyu and Darbon, J{\'e}r{\^o}me and Karniadakis, George Em},
  journal={arXiv preprint arXiv:2606.21828},
  year={2026}
}

@article{levrero2026nspod,
  title={{NSPOD}: accelerating the convergence of {Krylov}-based iterative linear solvers via approximated {PODs}},
  author={Levrero-Florencio, Francesc and Lee, Youngkyu and Pathak, Jay and Karniadakis, George Em},
  journal={arXiv preprint arXiv:2605.07828},
  year={2026}
}

@article{azulay2022multigrid,
  title={Multigrid-augmented deep learning preconditioners for the {Helmholtz} equation},
  author={Azulay, Yael and Treister, Eran},
  journal={SIAM Journal on Scientific Computing},
  volume={45},
  number={3},
  pages={S127--S151},
  year={2023}
}

@inproceedings{greenfeld2019learning,
  title={Learning to optimize multigrid {PDE} solvers},
  author={Greenfeld, Daniel and Galun, Meirav and Basri, Ronen and Yavneh, Irad and Kimmel, Ron},
  booktitle={Proceedings of the 36th International Conference on Machine Learning},
  pages={2415--2423},
  year={2019}
}

@inproceedings{luz2020learning,
  title={Learning algebraic multigrid using graph neural networks},
  author={Luz, Ilay and Galun, Meirav and Maron, Haggai and Basri, Ronen and Yavneh, Irad},
  booktitle={Proceedings of the 37th International Conference on Machine Learning},
  pages={6489--6499},
  year={2020}
}

@inproceedings{gotz2018machine,
  title={Machine learning-aided numerical linear algebra: Convolutional neural networks for the efficient preconditioner generation},
  author={G{\"o}tz, Markus and Anzt, Hartwig},
  booktitle={2018 IEEE/ACM 9th Workshop on Latest Advances in Scalable Algorithms for Large-Scale Systems (scalA)},
  pages={49--56},
  year={2018},
  organization={IEEE}
}

@book{miettinen1999nonlinear,
  title={Nonlinear Multiobjective Optimization},
  author={Miettinen, Kaisa},
  year={1999},
  publisher={Kluwer Academic Publishers},
  address={Boston}
}

@article{censor1977pareto,
  title={Pareto optimality in multiobjective problems},
  author={Censor, Yair},
  journal={Applied Mathematics and Optimization},
  volume={4},
  number={1},
  pages={41--59},
  year={1977}
}

@article{deb2002fast,
  title={A fast and elitist multiobjective genetic algorithm: {NSGA-II}},
  author={Deb, Kalyanmoy and Pratap, Amrit and Agarwal, Sameer and Meyarivan, T.},
  journal={IEEE Transactions on Evolutionary Computation},
  volume={6},
  number={2},
  pages={182--197},
  year={2002}
}

@article{cao2026automatic,
  title={Automatic selection of the best neural architecture for time series forecasting},
  author={Cao, Qianying and Liu, Shanqing and Varghese, Alan John and Darbon, J{\'e}r{\^o}me and Triantafyllou, Michael S. and Karniadakis, George Em},
  journal={Nature Communications},
  year={2026},
  doi={10.1038/s41467-026-73687-9}
}

@article{liu2026algorithms,
  title={Algorithms and differential game representations for exploring nonconvex {Pareto} fronts in high dimensions},
  author={Liu, Shanqing and Chen, Paula and Lee, Youngkyu and Darbon, J{\'e}r{\^o}me},
  journal={arXiv preprint arXiv:2602.11515},
  year={2026}
}

@article{manoj2026singularities,
  title={Singularities in multi-objective optimization and their crossing during continuation},
  author={Manoj, Arjun and Kavousanakis, Michail E. and Liu, Shanqing and Kevrekidis, Ioannis G.},
  journal={arXiv preprint arXiv:2607.02803},
  year={2026}
}

@article{romera2024mathematical,
  title={Mathematical discoveries from program search with large language models},
  author={Romera-Paredes, Bernardino and Barekatain, Mohammadamin and Novikov, Alexander and Balog, Matej and Kumar, M. Pawan and Dupont, Emilien and Ruiz, Francisco J. R. and Ellenberg, Jordan S. and Wang, Pengming and Fawzi, Omar and Kohli, Pushmeet and Fawzi, Alhussein},
  journal={Nature},
  volume={625},
  number={7995},
  pages={468--475},
  year={2024}
}

@article{novikov2025alphaevolve,
  title={{AlphaEvolve}: A coding agent for scientific and algorithmic discovery},
  author={Novikov, Alexander and V{\~u}, Ng{\^a}n and Eisenberger, Marvin and Dupont, Emilien and Huang, Po-Sen and Wagner, Adam Zsolt and Shirobokov, Sergey and Kozlovskii, Borislav and Ruiz, Francisco J. R. and Mehrabian, Abbas and others},
  journal={arXiv preprint arXiv:2506.13131},
  year={2025}
}

@article{boiko2023autonomous,
  title={Autonomous chemical research with large language models},
  author={Boiko, Daniil A. and MacKnight, Robert and Kline, Ben and Gomes, Gabe},
  journal={Nature},
  volume={624},
  number={7992},
  pages={570--578},
  year={2023}
}

@article{bran2024augmenting,
  title={Augmenting large language models with chemistry tools},
  author={M. Bran, Andres and Cox, Sam and Schilter, Oliver and Baldassari, Carlo and White, Andrew D. and Schwaller, Philippe},
  journal={Nature Machine Intelligence},
  volume={6},
  number={5},
  pages={525--535},
  year={2024}
}

@article{lu2024ai,
  title={The {AI} {Scientist}: Towards fully automated open-ended scientific discovery},
  author={Lu, Chris and Lu, Cong and Lange, Robert Tjarko and Foerster, Jakob and Clune, Jeff and Ha, David},
  journal={arXiv preprint arXiv:2408.06292},
  year={2024}
}

@article{du2023improving,
  title={Improving factuality and reasoning in language models through multiagent debate},
  author={Du, Yilun and Li, Shuang and Torralba, Antonio and Tenenbaum, Joshua B. and Mordatch, Igor},
  journal={arXiv preprint arXiv:2305.14325},
  year={2023}
}

@article{li2025codepde,
  title={{CodePDE}: An inference framework for {LLM}-driven {PDE} solver generation},
  author={Li, Shanda and Marwah, Tanya and Shen, Junhong and Sun, Weiwei and Risteski, Andrej and Yang, Yiming and Talwalkar, Ameet},
  journal={arXiv preprint arXiv:2505.08783},
  year={2025}
}

@article{toscano2025athena,
  title={{ATHENA}: Agentic team for hierarchical evolutionary numerical algorithms},
  author={Toscano, Juan Diego and Chen, Daniel T. and Karniadakis, George Em},
  journal={arXiv preprint arXiv:2512.03476},
  year={2025}
}

@article{jiang2026agenticsciml,
  title={{AgenticSciML}: collaborative multi-agent systems for emergent discovery in scientific machine learning},
  author={Jiang, Qile and Karniadakis, George Em},
  journal={npj Artificial Intelligence},
  volume={2},
  pages={57},
  year={2026},
  doi={10.1038/s44387-026-00102-5}
}

@article{du2026autonumerics,
  title={{AutoNumerics}: an autonomous, {PDE}-agnostic multi-agent pipeline for scientific computing},
  author={Du, Jianda and Sun, Youran and Yang, Haizhao},
  journal={arXiv preprint arXiv:2602.17607},
  year={2026}
}

@article{toscano2026graft,
  title={{GRAFT-ATHENA}: Self-improving agentic teams for autonomous discovery and evolutionary numerical algorithms},
  author={Toscano, Juan Diego and Chai, Zhaojie and Karniadakis, George Em},
  journal={arXiv preprint arXiv:2605.11117},
  year={2026}
}

@misc{thop,
  title={{THOP}: {PyTorch}-{OpCounter}},
  author={Zhu, Ligeng},
  year={2019},
  howpublished={Software, \url{https://github.com/ultralytics/thop}}
}

@article{yang2025qwen3,
  title={{Qwen3} Technical Report},
  author={Yang, An and Li, Anfeng and Yang, Baosong and Zhang, Beichen and Hui, Binyuan and others},
  journal={arXiv preprint arXiv:2505.09388},
  year={2025}
}

@misc{openai2025gptoss,
  title={gpt-oss-120b \& gpt-oss-20b Model Card},
  author={{OpenAI}},
  year={2025},
  howpublished={arXiv preprint arXiv:2508.10925}
}

@misc{qwen3coder,
  title={{Qwen3-Coder}},
  author={{Qwen Team}},
  year={2025},
  howpublished={Open-weight language model for code, \url{https://github.com/QwenLM/Qwen3-Coder}}
}

@misc{ollama,
  title={Ollama},
  author={{Ollama}},
  year={2026},
  howpublished={Software for serving language models locally, version 0.40.16, \url{https://ollama.com}}
}

@article{schmitt2021evostencils,
  title={{EvoStencils}: a grammar-based genetic programming approach for constructing efficient geometric multigrid methods},
  author={Schmitt, Jonas and Kuckuk, Sebastian and K{\"o}stler, Harald},
  journal={Genetic Programming and Evolvable Machines},
  volume={22},
  number={4},
  pages={511--537},
  year={2021},
  doi={10.1007/s10710-021-09412-w}
}

\clearpage
\renewcommand{\thesection}{S\arabic{section}}
\renewcommand{\thetable}{S\arabic{table}}
\renewcommand{\thefigure}{S\arabic{figure}}
\renewcommand{\theequation}{S\arabic{equation}}
\setcounter{table}{0}
\setcounter{figure}{0}
\setcounter{equation}{0}
\begin{center}
{\Large\bfseries Supplementary Information}\\[4pt]
{\large Multi-agent system for automatic discovery of Pareto-optimal meta-solvers for PDE}
\end{center}

\section*{Supplementary Note 1. Governing equations and discretizations}

\subsection*{Helmholtz scattering}
On the unit cube $\Omega=(0,1)^3$ with a sound-soft scatterer $\Omega_s$, the complex field $u$ satisfies
\begin{equation}
	\left\{\begin{aligned}
		\Delta u+k^2u&=0 &&\text{in }\Omega\setminus\bar\Omega_s,\\
		u&=0 &&\text{on }\partial\Omega_s,\\
		\partial_nu-iku&=g &&\text{on }\Gamma_{inc}=\{z=1\},\\
		\partial_nu-iku&=0 &&\text{on }\Gamma_{abs},
	\end{aligned}\right.
\end{equation}
	where $k=10\pi$ and $\Gamma_{abs}$ consists of the remaining outer faces. The incident field is the Gaussian beam $u_{inc}(x,y,z)=\frac{iR}{q}e^{-ikz}e^{-ikr^2/(2q)}$, with $R=\tfrac12kw_0^2$, $w_0=0.15$, $r^2=(x-0.5)^2+(y-0.5)^2$, and $q=(z-0.5)+iR$. The boundary datum is $g=\partial_nu_{inc}-iku_{inc}$. The weak form is
	\begin{equation}
		\int_{\Omega}\big(\nabla u\cdot\nabla\bar v-k^2u\bar v\big)\,d\xv-ik\int_{\Gamma_{inc}\cup\Gamma_{abs}}u\bar v\,ds=\int_{\Gamma_{inc}}g\bar v\,ds,
	\end{equation}
	and first-order Lagrange elements on hexahedra with Gauss-Lobatto-Legendre nodes and quadrature yield a complex indefinite system with 274,492 unknowns.

\subsection*{Linear elasticity}
The displacement $\uv$ of the component satisfies
\begin{equation}
	\left\{\begin{aligned}
		-\nabla\cdot\sigma(\uv)&=\mathbf{0} &&\text{in }\Omega,\\
		\uv&=\mathbf{0} &&\text{on }\Gamma_D,\\
		\sigma(\uv)\nv&=\mathbf{g} &&\text{on }\Gamma_{T},\\
        \sigma(\uv)\nv&=\mathbf{0} &&\text{on }\Gamma_N,
	\end{aligned}\right.
	\qquad \sigma(\uv)=\lambda\tr\varepsilon(\uv)\,\Id+2\mu\,\varepsilon(\uv),\quad \varepsilon(\uv)=\tfrac12\big(\nabla\uv+(\nabla\uv)^{\top}\big),
\end{equation}
with traction $\mathbf{g}=((0,0,-0.2)\cdot\nv)\nv$ on the remaining boundary $\Gamma_N$ and Lam\'e parameters $\lambda=E\nu/((1+\nu)(1-2\nu)) \approx 121.15$ and $\mu=E/(2(1+\nu))\approx80.77$, obtained from $E=210$ and $\nu=0.3$.
The weak form is 
\begin{equation}
	 \int_{\Omega} \sigma(\uv) \cdot \varepsilon(\mathbf{v}) d\xv = \int_{\Gamma_T} \mathbf{g} \cdot \mathbf{v} ds
\end{equation} 
for all test functions $\mathbf v$ vanishing on $\Gamma_D$, where the dot denotes full contraction of the two tensors.
First-order vector Lagrange elements on the hexahedral mesh with 134,172 nodes give a symmetric positive-definite system with 402,516 unknowns.

\subsection*{Coupled thermoelasticity}
The strain, the thermal strain, and the stress of the linear thermoelastic body are
\begin{equation}
\varepsilon(\uv)=\tfrac12\big(\nabla\uv+(\nabla\uv)^{\top}\big),\qquad
\varepsilon^{\mathrm{th}}=\alpha\,(T-T_0)\,\Id,\qquad
\sigma(\uv,T)=\lambda\tr\big(\varepsilon(\uv)-\varepsilon^{\mathrm{th}}\big)\Id+2\mu\big(\varepsilon(\uv)-\varepsilon^{\mathrm{th}}\big),
\end{equation}
with the Lam\'e parameters, the bulk modulus, and the coupling modulus
\begin{equation}
\lambda=\frac{E\nu}{(1+\nu)(1-2\nu)},\qquad \mu=\frac{E}{2(1+\nu)},\qquad K=\lambda+\tfrac23\mu,\qquad \beta=3K\alpha,
\end{equation}
so that in three dimensions the thermal part of the stress is $-\beta(T-T_0)\Id$, as written in \cref{eq:thermo}. The archived two-dimensional task takes the thermal strain in the plane, $\varepsilon^{\mathrm{th}}=\alpha(T-T_0)\Id_2$, so that the thermal-stress coefficient of its momentum balance is $2(\lambda+\mu)\alpha$ while the dissipation term of its energy balance keeps $\beta=3K\alpha$. The material data, in SI units, are
\begin{equation}
\begin{aligned}
&E=70\times10^{9},\qquad \nu=0.33,\qquad \alpha=23\times10^{-6},\qquad \kappa=205,\qquad \rho=2700,\\
&c_\varepsilon=900,\qquad T_h=333.15,\qquad T_c=293.15,\qquad T_0=293.15,
\end{aligned}
\end{equation}
with no volumetric heat source and the final time $t_f=10$. With $\Gamma_L$ the end at $x=0$, $\Gamma_R$ the end at $x=L$, and $\Gamma_N$ the remaining boundary of the strip $(0,L)\times(0,H)$ or of the slab $(0,L)\times(0,H)\times(0,W)$, where $L=0.1$, $H=0.02$, and $W=0.01$, the boundary and initial data are
\begin{equation}
\left\{\begin{aligned}
T&=T_h\ \text{ on }\Gamma_L, & T&=T_c\ \text{ on }\Gamma_R, & \kappa\nabla T\cdot\nv&=0\ \text{ on }\Gamma_N,\\
u_x&=0\ \text{ on }\Gamma_L, & \sigma(\uv,T)\,\nv&=\mathbf{0}\ \text{ on }\partial\Omega\setminus\Gamma_L, & T(\cdot,0)&=T_0,\quad \uv(\cdot,0)=\mathbf{0},\quad \partial_t\uv(\cdot,0)=\mathbf{0},
\end{aligned}\right.
\end{equation}
and the origin is pinned in the directions transverse to $x$. The displacement block is nonsingular because of the inertial term and these constraints.

\subsection*{Time discretization}
The temperature equation is advanced by the Crank-Nicolson rule and the momentum balance by the trapezoidal rule, which is the Newmark scheme with $\beta_N=\tfrac14$ and $\gamma_N=\tfrac12$ once the velocity and the acceleration are eliminated. With $\Delta t$ the step and the superscript the time level, the semi-discrete equations are
\begin{align}
\frac{\rho c_\varepsilon}{\Delta t}\big(T^{k+1}-T^{k}\big)&=\tfrac12\nabla\cdot\big(\kappa\nabla T^{k+1}\big)+\tfrac12\nabla\cdot\big(\kappa\nabla T^{k}\big)-\frac{\beta T_0}{\Delta t}\tr\big(\varepsilon(\uv^{k+1})-\varepsilon(\uv^{k})\big),\\
\rho\,\frac{\uv^{k+1}-2\uv^{k}+\uv^{k-1}}{\Delta t^2}&=\tfrac14\nabla\cdot\sigma^{k+1}+\tfrac12\nabla\cdot\sigma^{k}+\tfrac14\nabla\cdot\sigma^{k-1},\qquad \sigma^{j}=\sigma(\uv^{j},T^{j}),
\end{align}
which are second order in time and unconditionally stable for the linear problem. Both instances use $h=1/800$, $\Delta t=2h=1/400$, and 4,000 steps up to $t_f=10$.

\subsection*{Weak form and block system}
With $\{\phi_i\}$ the scalar basis of the temperature space and $\{\psi_i\}$ the vector basis of the displacement space, testing the two semi-discrete equations against the basis functions gives the block system \cref{eq:block} with the entries
\begin{align}
(K_{TT})_{ij}&=\frac{\rho c_\varepsilon}{\Delta t}\int_\Omega\phi_j\phi_i\,d\xv+\tfrac12\int_\Omega\kappa\nabla\phi_j\cdot\nabla\phi_i\,d\xv,\\
(K_{Tu})_{ij}&=\frac{\beta T_0}{\Delta t}\int_\Omega\tr\varepsilon(\psi_j)\,\phi_i\,d\xv,\\
(K_{uT})_{ij}&=-\tfrac14\int_\Omega\beta\,\phi_j\,\tr\varepsilon(\psi_i)\,d\xv,\\
(K_{uu})_{ij}&=\frac{\rho}{\Delta t^2}\int_\Omega\psi_j\cdot\psi_i\,d\xv+\tfrac14\int_\Omega\sigma_{\mathrm{el}}(\psi_j)\cdot\varepsilon(\psi_i)\,d\xv,
\end{align}
where $\sigma_{\mathrm{el}}(\psi_j)=\lambda\tr\varepsilon(\psi_j)\Id+2\mu\varepsilon(\psi_j)$ is the elastic stress of the $j$-th displacement mode and the factors $\tfrac14$ are the trapezoidal weights of the stress at the level $k+1$. The right-hand side gathers the previous temperature, the coupling term at level $k$, and the inertial and stress history of the displacement at the levels $k$ and $k-1$,
\begin{align}
(\bv_T)_i&=\frac{\rho c_\varepsilon}{\Delta t}\int_\Omega T^{k}\phi_i\,d\xv-\tfrac12\int_\Omega\kappa\nabla T^{k}\cdot\nabla\phi_i\,d\xv+\frac{\beta T_0}{\Delta t}\int_\Omega\tr\varepsilon(\uv^{k})\,\phi_i\,d\xv,\\
(\bv_u)_i&=\frac{\rho}{\Delta t^2}\int_\Omega\big(2\uv^{k}-\uv^{k-1}\big)\cdot\psi_i\,d\xv-\tfrac12\int_\Omega\sigma^{k}\cdot\varepsilon(\psi_i)\,d\xv\nonumber\\
&\qquad-\tfrac14\int_\Omega\sigma^{k-1}\cdot\varepsilon(\psi_i)\,d\xv-\tfrac14\int_\Omega\beta T_0\,\tr\varepsilon(\psi_i)\,d\xv,
\end{align}
where the last term of $\bv_u$ is the part of the thermal stress at level $k+1$ that does not depend on the unknowns. In the two-dimensional task the coefficient $\beta$ of $K_{uT}$, of $\sigma^{j}$, and of the last term of $\bv_u$ is replaced by $2(\lambda+\mu)\alpha$, as stated above. Both off-diagonal blocks are present because the coupling is two-way. The quadrilateral mesh of the strip has 1,377 nodes and 1,280 cells and the hexahedral mesh of the slab has 12,393 nodes and 10,240 cells, both with $h=1/800$, which gives 4,131 and 49,572 unknowns per step.

\section*{Supplementary Note 2. Excerpts from the debate transcripts}
The following passages summarize exchanges recorded in the archived transcripts of the runs with open-weight models, using indirect speech and staying as close as possible to the original wording.

\paragraph{Helmholtz, generation 2, Proposer.}

Under the heading of validation checks, the Proposer recorded that no candidate of the new generation used the multigrid role with the trunk basis, which had failed in \cand{1\_001} to \cand{1\_008}, that all candidates used the one validated DeepONet checkpoint, and that all used the balancing mode of the deflation, the only mode that had been successful, except for \cand{2\_009}, which it described as a known risky mutation kept for comparison. It added that the candidates were variations of the successful deflation approach rather than explorations of the failed methods, and that the key parameters, the rank, the Krylov method, the base preconditioner, and the restart, were explored within the validated ranges.

\paragraph{Helmholtz, generation 2, Critic.}

The Critic replied that the generation stayed within the deflation module and preserved the successful parameter families, but that nine candidates set a restart length of zero for GMRES, which the module does not admit, and that the plan contained many near-duplicate configurations concentrated in a small region of the parameter space. It asked for a positive restart in these candidates, proposed a restart of 50 to test a different Krylov subspace size, and asked whether two or more clusters of the basis could be explored. The restart of 50 entered the final plan of the same generation, and the two-cluster variants appeared in the third generation, where one of them, \cand{3\_017}, gave the lowest error of the run.

\paragraph{Elasticity, generation 2, Proposer.}

In its execution notes, the Proposer wrote that the base preconditioner of the multigrid levels was always SOR, because Jacobi had failed in all attempts with high residual ratios while SOR had been successful in all working candidates, that post-smoothing was always enabled, because the best performing candidates had used it, and that no candidate exceeded the residual-ratio limit of the acceptance test, since all were based on configurations that had passed it. The parents were the four candidates of the first front, \cand{1\_007}, \cand{1\_010}, \cand{1\_011}, and \cand{1\_005}, together with \cand{1\_006} from the second front to cover a different region of the trade-off.

\paragraph{Thermoelasticity in two dimensions, generation 1, Critic.}

In its first verdict, the Critic wrote that the proposal was a solid first pass that used the only admissible neural module and covered a wide range of the rank, the clustering, the mode, and the block preconditioner, but that it contained a critical mathematical mismatch that would almost certainly prevent any candidate from converging. The global Krylov method was set to conjugate gradients on a nonsymmetric coupled block system, whereas conjugate gradients requires a symmetric positive-definite matrix and the thermoelastic block matrix has off-diagonal coupling blocks that are not transposes of each other. The final plan of the generation used FGMRES in all twenty candidates.

\paragraph{Thermoelasticity in three dimensions, generation 2, Proposer.}

In its decision trace, the Proposer noted that \cand{1\_016}, with a rank-10 deflation of the displacement block, had the best error of the first generation, and that the mutations of the new generation were limited to the rank, the deflation mode, and the block preconditioners, from incomplete LU to incomplete Cholesky and to multigrid. It kept FGMRES as the Krylov method, which it described as the only reliable choice for the deflation, and kept the full Schur factorization.

\paragraph{Helmholtz, generation 1, single agent.}

The single agent returned, in place of a plan, a status block stating that the plan was blocked because no executable capability set existed for the requested hybrid generation. The validator returned this to the planner with the request for a complete corrected plan, and the two repair attempts returned the same status.

\section*{Supplementary Note 3. Discovery cost}

\Cref{tab:cost} lists the measured token consumption and the wall-clock time of the language models for each problem and each system, summed over the three generations. Planning covers the Proposer, the Critic of the multi-agent system, and the repairs requested by the validator. Engineering covers the implementation of the candidates, the runtime repairs, and the analysis of the results, and it is zero when no generation was executed. Periods of system sleep were subtracted from the model times. All models were served locally, so no cost in currency is associated with these tokens.

\begin{table}[h]
\centering
\caption{Measured token consumption and model time of the planning and of the engineering, summed over the three generations, together with the number of repairs requested by the validator and the number of completed candidates among those executed.
	}
\label{tab:cost}
\footnotesize
\setlength{\tabcolsep}{3pt}
\begin{tabularx}{\textwidth}{@{}Xlrrrrrrrr@{}}
\toprule
 & & \multicolumn{3}{c}{Planning} & \multicolumn{3}{c}{Engineering} & & \\
\cmidrule(lr){3-5}\cmidrule(lr){6-8}
Class & System & Input & Output & Min & Input & Output & Min & Repairs & Completed \\
\midrule
Helmholtz, $k=10\pi$ & multi-agent & 775,289 & 141,241 & 60.6 & 11,667,818 & 49,107 & 30.2 & 3 & 47 of 60 \\
 & single agent & 331,382 & 32,982 & 14.9 & 0 & 0 & 0.0 & 6 & 0 of 0 \\
Linear elasticity & multi-agent & 806,120 & 137,992 & 60.3 & 7,371,286 & 36,094 & 16.2 & 4 & 48 of 60 \\
 & single agent & 343,256 & 54,039 & 27.2 & 0 & 0 & 0.0 & 6 & 0 of 0 \\
Thermoelasticity 2D & multi-agent & 1,019,346 & 172,095 & 88.6 & 7,308,997 & 29,239 & 21.2 & 6 & 40 of 40 \\
 & single agent & 395,613 & 52,552 & 28.7 & 6,448,916 & 20,969 & 16.7 & 6 & 40 of 40 \\
Thermoelasticity 3D & multi-agent & 971,948 & 157,157 & 86.8 & 11,031,942 & 50,419 & 31.1 & 5 & 58 of 60 \\
 & single agent & 357,142 & 69,278 & 40.0 & 4,586,634 & 21,708 & 12.4 & 5 & 19 of 40 \\
\bottomrule
\end{tabularx}
\end{table}

\section*{Supplementary Note 4. Complete listings of the evaluated candidates}

The following tables list all executed candidates of the eight runs. Each candidate is described by the configuration that was executed, read from its script, and not by the name given by the planner. For the Helmholtz and elasticity problems, the target is convergence to the prescribed tolerance, and no candidate was executed by the single agent. For the thermoelastic problems, the target is a final relative error at or below $10^{-4}$ in the temperature and in the displacement, which no candidate meets, so the two errors are listed. Timings marked with an asterisk were measured again in isolation after system sleep, battery operation, or external processor load had affected the original measurement. Iterations, residuals, and errors were identical in every repeated measurement. A few candidates coincide with earlier ones, namely \cand{1\_016} and \cand{2\_001} and \cand{2\_019} and \cand{3\_012} of the multi-agent system in three dimensions, and several candidates of the single agent in both dimensions.

\begin{figure}[h]
\centering
\includegraphics[width=\textwidth]{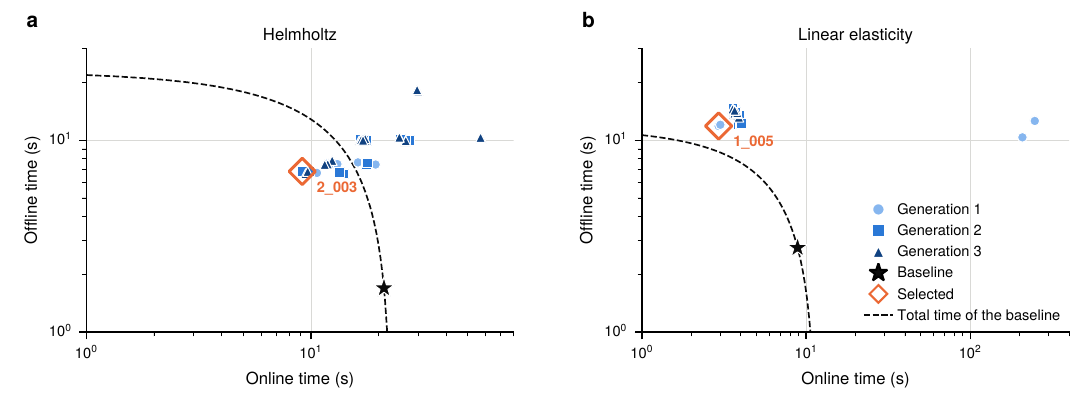}
\caption{Offline time against online time of the converged candidates of the multi-agent system. \textbf{a}, Helmholtz problem. \textbf{b}, Elasticity problem. The stars denote the baselines, and the dashed curves are the points with the same total time as the baseline. The orange diamonds mark the fastest candidates in online time, \cand{2\_003} and \cand{1\_005}.}
\label{fig:si_offline}
\end{figure}

\Cref{fig:si_offline} shows the offline time of the converged single-physics candidates against their online time. The offline time contains the assembly, the setup of the preconditioner, and the construction of the neural basis. For the Helmholtz problem, fifteen candidates remain below the baseline when the two times are added. For the elasticity problem, the online time of the multigrid candidates is less than half that of the baseline, and their offline time of 12 to 15~s makes the total larger than that of the baseline for a single solve.

\setlength{\LTcapwidth}{\textwidth}

\begingroup
\footnotesize
\setlength{\tabcolsep}{3pt}
\begin{longtable}{@{}lp{\dimexpr\textwidth-10.6cm\relax}rrrccc@{}}
\caption{All candidates of the multi-agent system for the Helmholtz problem. Iterations, online time, offline time, and MAC refer to one solve. TB denotes the trunk basis.}\label{tab:si_helm}\\
\toprule
ID & Configuration & Iterations & Online (s) & Offline (s) & MAC & Rel.\ $L^2$ error & Target \\
\midrule
\endfirsthead
\toprule
ID & Configuration & Iterations & Online (s) & Offline (s) & MAC & Rel.\ $L^2$ error & Target \\
\midrule
\endhead
\bottomrule
\endfoot
1\_001 & FGMRES(50), GAMG with trunk basis $k=16$, SOR, post-smoothing & \multicolumn{5}{c}{rejected by the acceptance test} & no \\
1\_002 & FGMRES(50), GAMG with trunk basis $k=32$, SOR, post-smoothing & \multicolumn{5}{c}{rejected by the acceptance test} & no \\
1\_003 & FGMRES(50), GAMG with trunk basis $k=64$, SOR, post-smoothing & \multicolumn{5}{c}{rejected by the acceptance test} & no \\
1\_004 & FGMRES(50), GAMG with trunk basis $k=32$, Jacobi, post-smoothing & \multicolumn{5}{c}{rejected by the acceptance test} & no \\
1\_005 & FGMRES(50), GAMG with trunk basis $k=32$, SOR, no post-smoothing & \multicolumn{5}{c}{rejected by the acceptance test} & no \\
1\_006 & FGMRES(20), GAMG with trunk basis $k=32$, SOR, post-smoothing & \multicolumn{5}{c}{rejected by the acceptance test} & no \\
1\_007 & FGMRES(50), GAMG with trunk basis $k=16$, Jacobi, no post-smoothing & \multicolumn{5}{c}{rejected by the acceptance test} & no \\
1\_008 & FGMRES(50), GAMG with trunk basis $k=16$, SOR, post-smoothing & \multicolumn{5}{c}{rejected by the acceptance test} & no \\
1\_009 & FGMRES(100), balancing TB deflation $k=16$, ILU & 231 & 10.64 & 6.8 & $1.33\times10^{11}$ & $2.32\times10^{-8}$ & yes \\
1\_010 & FGMRES(100), balancing TB deflation $k=32$, ILU & 229 & 12.72 & 7.5 & $1.65\times10^{11}$ & $1.81\times10^{-8}$ & yes \\
1\_011 & FGMRES(100), balancing TB deflation $k=64$, ILU & 220 & 17.13 & 9.9 & $2.21\times10^{11}$ & $2.18\times10^{-8}$ & yes \\
1\_012 & FGMRES(100), balancing TB deflation $k=32$, SOR & 341 & 19.41 & 7.5 & $2.47\times10^{11}$ & $2.30\times10^{-8}$ & yes \\
1\_013 & FGMRES(20), balancing TB deflation $k=32$, ILU & 263 & 11.86 & 7.6 & $1.49\times10^{11}$ & $2.10\times10^{-8}$ & yes \\
1\_014 & GMRES(100), balancing TB deflation $k=32$, ILU & 229 & 12.99 & 7.5 & $1.66\times10^{11}$ & $1.81\times10^{-8}$ & yes \\
1\_015 & FGMRES(100), additive TB deflation $k=32$, ILU & 432 & 16.20 & 7.7 & $2.05\times10^{11}$ & $1.37\times10^{-8}$ & yes \\
1\_016 & FGMRES(100), balancing TB deflation $k=64$, ILU & 220 & 17.61 & 10.1 & $2.21\times10^{11}$ & $2.18\times10^{-8}$ & yes \\
1\_017 & GMRES(20), balancing TB deflation $k=32$, ILU & 263 & 12.46 & 7.6 & $1.55\times10^{11}$ & $2.10\times10^{-8}$ & yes \\
1\_018 & GMRES(100), balancing TB deflation $k=64$, ILU & 220 & 17.76 & 10.0 & $2.23\times10^{11}$ & $2.18\times10^{-8}$ & yes \\
1\_019 & FGMRES(100), balancing TB deflation $k=32$, ILU & 229 & 13.11 & 7.5 & $1.65\times10^{11}$ & $1.81\times10^{-8}$ & yes \\
1\_020 & FGMRES(50), GAMG with trunk basis $k=64$, Jacobi, post-smoothing & \multicolumn{5}{c}{rejected by the acceptance test} & no \\
2\_001 & GMRES(20), balancing TB deflation $k=16$, ILU & 262 & 9.83 & 6.9 & $1.16\times10^{11}$ & $2.34\times10^{-8}$ & yes \\
2\_002 & GMRES(20), balancing TB deflation $k=16$, SOR & 378 & 14.00 & 6.7 & $1.67\times10^{11}$ & $2.25\times10^{-8}$ & yes \\
2\_003 & FGMRES(20), balancing TB deflation $k=16$, ILU & 262 & 9.15 & 6.9 & $1.11\times10^{11}$ & $2.34\times10^{-8}$ & yes \\
2\_004 & FGMRES(20), balancing TB deflation $k=16$, SOR & 378 & 13.43 & 6.8 & $1.61\times10^{11}$ & $2.25\times10^{-8}$ & yes \\
2\_005 & GMRES(20), balancing TB deflation $k=32$, ILU & \multicolumn{5}{c}{rejected by the acceptance test} & no \\
2\_006 & GMRES(50), balancing TB deflation $k=32$, ILU & \multicolumn{5}{c}{rejected by the acceptance test} & no \\
2\_007 & GMRES(20), balancing TB deflation $k=32$, SOR & 385 & 17.85 & 7.4 & $2.27\times10^{11}$ & $2.30\times10^{-8}$ & yes \\
2\_008 & GMRES(50), balancing TB deflation $k=32$, SOR & 355 & 17.85 & 7.5 & $2.27\times10^{11}$ & $2.40\times10^{-8}$ & yes \\
2\_009 & FGMRES(20), balancing TB deflation $k=32$, ILU & \multicolumn{5}{c}{rejected by the acceptance test} & no \\
2\_010 & FGMRES(50), balancing TB deflation $k=32$, ILU & \multicolumn{5}{c}{rejected by the acceptance test} & no \\
2\_011 & FGMRES(20), balancing TB deflation $k=32$, SOR & 385 & 17.62 & 7.4 & $2.18\times10^{11}$ & $2.30\times10^{-8}$ & yes \\
2\_012 & FGMRES(50), balancing TB deflation $k=32$, SOR & 355 & 17.80 & 7.6 & $2.23\times10^{11}$ & $2.40\times10^{-8}$ & yes \\
2\_013 & GMRES(20), balancing TB deflation $k=64$, ILU & 251 & 17.75 & 9.9 & $2.22\times10^{11}$ & $2.28\times10^{-8}$ & yes \\
2\_014 & GMRES(50), balancing TB deflation $k=64$, ILU & 230 & 16.92 & 10.1 & $2.12\times10^{11}$ & $2.18\times10^{-8}$ & yes \\
2\_015 & GMRES(20), balancing TB deflation $k=64$, SOR & 385 & 27.53 & 10.0 & $3.41\times10^{11}$ & $2.28\times10^{-8}$ & yes \\
2\_016 & GMRES(50), balancing TB deflation $k=64$, SOR & 344 & 25.50 & 10.1 & $3.18\times10^{11}$ & $2.31\times10^{-8}$ & yes \\
2\_017 & FGMRES(20), balancing TB deflation $k=64$, ILU & 251 & 17.17 & 10.0 & $2.12\times10^{11}$ & $2.28\times10^{-8}$ & yes \\
2\_018 & FGMRES(50), balancing TB deflation $k=64$, ILU & 230 & 16.60 & 10.1 & $2.08\times10^{11}$ & $2.18\times10^{-8}$ & yes \\
2\_019 & FGMRES(20), balancing TB deflation $k=64$, SOR & 385 & 26.32 & 10.0 & $3.26\times10^{11}$ & $2.28\times10^{-8}$ & yes \\
2\_020 & FGMRES(50), balancing TB deflation $k=64$, SOR & 344 & 25.11 & 10.1 & $3.13\times10^{11}$ & $2.31\times10^{-8}$ & yes \\
3\_001 & GMRES(50), balancing TB deflation $k=32$, ILU & 239 & 12.44 & 7.8 & $1.52\times10^{11}$ & $2.00\times10^{-8}$ & yes \\
3\_002 & GMRES(50), balancing TB deflation $k=64$, ILU & 230 & 16.38 & 9.9 & $2.12\times10^{11}$ & $2.18\times10^{-8}$ & yes \\
3\_003 & GMRES(25), balancing TB deflation $k=32$, ILU & 257 & 11.88 & 7.5 & $1.53\times10^{11}$ & $2.03\times10^{-8}$ & yes \\
3\_004 & GMRES(100), balancing TB deflation $k=64$, ILU & 220 & 17.02 & 9.9 & $2.23\times10^{11}$ & $2.18\times10^{-8}$ & yes \\
3\_005 & GMRES(50), balancing TB deflation $k=16$, ILU & 238 & 9.46 & 6.7 & $1.17\times10^{11}$ & $2.33\times10^{-8}$ & yes \\
3\_006 & GMRES(50), balancing TB deflation $k=128$, ILU & 216 & 30.11 & 18.0 & $3.23\times10^{11}$ & $2.34\times10^{-8}$ & yes \\
3\_007 & GMRES(50), additive TB deflation $k=64$, ILU & 638 & 26.34 & 9.9 & $3.33\times10^{11}$ & $9.51\times10^{-9}$ & yes \\
3\_008 & GMRES(50), balancing TB deflation $k=64$, 2 clusters, ILU & 341 & 24.96 & 10.2 & $3.15\times10^{11}$ & $2.18\times10^{-8}$ & yes \\
3\_009 & GMRES(50), additive TB deflation $k=64$, 2 clusters, ILU & 1,377 & 57.67 & 10.2 & $7.19\times10^{11}$ & $8.46\times10^{-9}$ & yes \\
3\_010 & FGMRES(50), balancing TB deflation $k=32$, ILU & 239 & 11.52 & 7.5 & $1.50\times10^{11}$ & $2.00\times10^{-8}$ & yes \\
3\_011 & FGMRES(50), balancing TB deflation $k=64$, ILU & 230 & 16.62 & 10.0 & $2.08\times10^{11}$ & $2.18\times10^{-8}$ & yes \\
3\_012 & FGMRES(100), balancing TB deflation $k=64$, ILU & 220 & 17.51 & 10.0 & $2.21\times10^{11}$ & $2.18\times10^{-8}$ & yes \\
3\_013 & FGMRES(50), balancing TB deflation $k=16$, ILU & 238 & 9.67 & 6.9 & $1.15\times10^{11}$ & $2.33\times10^{-8}$ & yes \\
3\_014 & FGMRES(50), balancing TB deflation $k=128$, ILU & 216 & 29.71 & 18.3 & $3.17\times10^{11}$ & $2.34\times10^{-8}$ & yes \\
3\_015 & FGMRES(50), additive TB deflation $k=64$, ILU & 638 & 26.43 & 9.9 & $3.28\times10^{11}$ & $9.51\times10^{-9}$ & yes \\
3\_016 & FGMRES(50), balancing TB deflation $k=64$, 2 clusters, ILU & 341 & 24.76 & 10.3 & $3.10\times10^{11}$ & $2.18\times10^{-8}$ & yes \\
3\_017 & FGMRES(50), additive TB deflation $k=64$, 2 clusters, ILU & 1,377 & 56.96 & 10.3 & $7.09\times10^{11}$ & $8.46\times10^{-9}$ & yes \\
3\_018 & GMRES(25), balancing TB deflation $k=64$, ILU & 244 & 17.45 & 10.2 & $2.16\times10^{11}$ & $2.23\times10^{-8}$ & yes \\
3\_019 & FGMRES(25), balancing TB deflation $k=64$, ILU & 244 & 16.73 & 10.0 & $2.09\times10^{11}$ & $2.23\times10^{-8}$ & yes \\
3\_020 & GMRES(50), balancing TB deflation $k=64$, ILU & 230 & 17.06 & 10.0 & $2.12\times10^{11}$ & $2.18\times10^{-8}$ & yes
\end{longtable}

\begin{longtable}{@{}lp{\dimexpr\textwidth-10.6cm\relax}rrrccc@{}}
\caption{All candidates of the multi-agent system for the elasticity problem. Iterations, online time, offline time, and MAC refer to one solve.}\label{tab:si_elas}\\
\toprule
ID & Configuration & Iterations & Online (s) & Offline (s) & MAC & Rel.\ $L^2$ error & Target \\
\midrule
\endfirsthead
\toprule
ID & Configuration & Iterations & Online (s) & Offline (s) & MAC & Rel.\ $L^2$ error & Target \\
\midrule
\endhead
\bottomrule
\endfoot
1\_001 & FGMRES(100), balancing TB deflation $k=16$, ILU & \multicolumn{5}{c}{diverged} & no \\
1\_002 & FGMRES(100), balancing TB deflation $k=32$, ILU & 5,498 & 207.64 & 10.3 & $2.33\times10^{12}$ & $7.29\times10^{-8}$ & yes \\
1\_003 & FGMRES(100), balancing TB deflation $k=64$, ILU & 5,002 & 246.40 & 12.7 & $2.57\times10^{12}$ & $7.88\times10^{-8}$ & yes \\
1\_004 & FGMRES(100), balancing TB deflation $k=128$, ILU & 5,002 & 246.89 & 12.6 & $2.57\times10^{12}$ & $7.88\times10^{-8}$ & yes \\
1\_005 & FGMRES(50), GAMG with trunk basis $k=8$, SOR, no post-smoothing & 46 & 2.93 & 11.8 & $2.52\times10^{10}$ & $1.00\times10^{-9}$ & yes \\
1\_006 & FGMRES(50), GAMG with trunk basis $k=16$, SOR, no post-smoothing & 46 & 3.00 & 12.0 & $2.56\times10^{10}$ & $1.44\times10^{-9}$ & yes \\
1\_007 & FGMRES(50), GAMG with trunk basis $k=32$, SOR, post-smoothing & 38 & 4.03 & 12.6 & $3.18\times10^{10}$ & $5.47\times10^{-9}$ & yes \\
1\_008 & FGMRES(50), GAMG with trunk basis $k=32$, Jacobi, post-smoothing & \multicolumn{5}{c}{rejected by the acceptance test} & no \\
1\_009 & FGMRES(50), GAMG with trunk basis $k=32$, Jacobi, post-smoothing & \multicolumn{5}{c}{rejected by the acceptance test} & no \\
1\_010 & FGMRES(50), GAMG with trunk basis $k=64$, SOR, post-smoothing & 34 & 3.81 & 13.8 & $2.94\times10^{10}$ & $8.93\times10^{-10}$ & yes \\
1\_011 & FGMRES(50), GAMG with trunk basis $k=128$, SOR, post-smoothing & 32 & 3.56 & 14.5 & $2.78\times10^{10}$ & $1.65\times10^{-9}$ & yes \\
1\_012 & FGMRES(50), GAMG with trunk basis $k=32$, SOR, no post-smoothing & \multicolumn{5}{c}{rejected by the acceptance test} & no \\
1\_013 & FGMRES(50), GAMG with trunk basis $k=16$, Jacobi, post-smoothing & \multicolumn{5}{c}{rejected by the acceptance test} & no \\
1\_014 & FGMRES(50), GAMG with trunk basis $k=64$, Jacobi, post-smoothing & \multicolumn{5}{c}{rejected by the acceptance test} & no \\
1\_015 & FGMRES(50), GAMG with trunk basis $k=64$, Jacobi, post-smoothing & \multicolumn{5}{c}{rejected by the acceptance test} & no \\
1\_016 & FGMRES(50), GAMG with trunk basis $k=16$, Jacobi, post-smoothing & \multicolumn{5}{c}{rejected by the acceptance test} & no \\
1\_017 & FGMRES(50), GAMG with trunk basis $k=8$, Jacobi, post-smoothing & \multicolumn{5}{c}{rejected by the acceptance test} & no \\
1\_018 & FGMRES(50), GAMG with trunk basis $k=8$, Jacobi, post-smoothing & \multicolumn{5}{c}{rejected by the acceptance test} & no \\
1\_019 & FGMRES(50), GAMG with trunk basis $k=128$, Jacobi, post-smoothing & \multicolumn{5}{c}{rejected by the acceptance test} & no \\
1\_020 & FGMRES(50), GAMG with trunk basis $k=128$, Jacobi, post-smoothing & \multicolumn{5}{c}{rejected by the acceptance test} & no \\
2\_001 & FGMRES(50), GAMG with trunk basis $k=48$, SOR, post-smoothing & 36 & 3.93 & 13.5 & $3.07\times10^{10}$ & $3.58\times10^{-9}$ & yes \\
2\_002 & FGMRES(50), GAMG with trunk basis $k=56$, SOR, post-smoothing & 33 & 3.63 & 13.7 & $2.84\times10^{10}$ & $1.49\times10^{-9}$ & yes \\
2\_003 & FGMRES(50), GAMG with trunk basis $k=60$, SOR, post-smoothing & 33 & 3.65 & 13.9 & $2.85\times10^{10}$ & $4.69\times10^{-10}$ & yes \\
2\_004 & FGMRES(50), GAMG with trunk basis $k=62$, SOR, post-smoothing & 34 & 3.79 & 13.9 & $2.94\times10^{10}$ & $8.22\times10^{-10}$ & yes \\
2\_005 & FGMRES(50), GAMG with trunk basis $k=68$, SOR, post-smoothing & 34 & 3.76 & 13.9 & $2.95\times10^{10}$ & $1.31\times10^{-9}$ & yes \\
2\_006 & FGMRES(50), GAMG with trunk basis $k=70$, SOR, post-smoothing & 34 & 3.76 & 13.9 & $2.95\times10^{10}$ & $1.30\times10^{-9}$ & yes \\
2\_007 & FGMRES(50), GAMG with trunk basis $k=72$, SOR, post-smoothing & 33 & 3.67 & 13.9 & $2.86\times10^{10}$ & $4.19\times10^{-9}$ & yes \\
2\_008 & FGMRES(50), GAMG with trunk basis $k=80$, SOR, post-smoothing & 33 & 3.67 & 14.1 & $2.86\times10^{10}$ & $4.44\times10^{-9}$ & yes \\
2\_009 & FGMRES(50), GAMG with trunk basis $k=85$, SOR, post-smoothing & 33 & 3.65 & 14.1 & $2.86\times10^{10}$ & $3.82\times10^{-9}$ & yes \\
2\_010 & FGMRES(50), GAMG with trunk basis $k=90$, SOR, post-smoothing & 33 & 3.69 & 14.2 & $2.86\times10^{10}$ & $2.45\times10^{-9}$ & yes \\
2\_011 & FGMRES(50), GAMG with trunk basis $k=96$, SOR, post-smoothing & 33 & 3.67 & 14.2 & $2.86\times10^{10}$ & $2.08\times10^{-9}$ & yes \\
2\_012 & FGMRES(50), GAMG with trunk basis $k=104$, SOR, post-smoothing & 33 & 3.69 & 14.3 & $2.87\times10^{10}$ & $8.90\times10^{-10}$ & yes \\
2\_013 & FGMRES(50), GAMG with trunk basis $k=112$, SOR, post-smoothing & 32 & 3.60 & 14.6 & $2.78\times10^{10}$ & $6.24\times10^{-10}$ & yes \\
2\_014 & FGMRES(50), GAMG with trunk basis $k=120$, SOR, post-smoothing & 32 & 3.56 & 14.6 & $2.78\times10^{10}$ & $6.28\times10^{-9}$ & yes \\
2\_015 & FGMRES(50), GAMG with trunk basis $k=128$, SOR, post-smoothing & 32 & 3.58 & 14.6 & $2.78\times10^{10}$ & $1.65\times10^{-9}$ & yes \\
2\_016 & FGMRES(50), GAMG with trunk basis $k=76$, SOR, post-smoothing & 33 & 3.72 & 14.0 & $2.86\times10^{10}$ & $4.57\times10^{-9}$ & yes \\
2\_017 & FGMRES(50), GAMG with trunk basis $k=12$, SOR, post-smoothing & 38 & 3.88 & 12.0 & $3.10\times10^{10}$ & $1.83\times10^{-9}$ & yes \\
2\_018 & FGMRES(50), GAMG with trunk basis $k=14$, SOR, post-smoothing & 39 & 3.99 & 12.1 & $3.19\times10^{10}$ & $1.55\times10^{-9}$ & yes \\
2\_019 & FGMRES(50), GAMG with trunk basis $k=16$, SOR, post-smoothing & 39 & 4.01 & 12.1 & $3.20\times10^{10}$ & $1.23\times10^{-9}$ & yes \\
2\_020 & FGMRES(50), GAMG with trunk basis $k=18$, SOR, post-smoothing & 38 & 4.03 & 12.2 & $3.12\times10^{10}$ & $3.31\times10^{-9}$ & yes \\
3\_001 & FGMRES(50), GAMG with trunk basis $k=50$, SOR, post-smoothing & 35 & 3.81 & 13.4 & $2.99\times10^{10}$ & $1.77\times10^{-9}$ & yes \\
3\_002 & FGMRES(50), GAMG with trunk basis $k=58$, SOR, post-smoothing & 33 & 3.67 & 13.7 & $2.85\times10^{10}$ & $5.24\times10^{-10}$ & yes \\
3\_003 & FGMRES(50), GAMG with trunk basis $k=66$, SOR, post-smoothing & 34 & 3.77 & 13.9 & $2.94\times10^{10}$ & $2.26\times10^{-10}$ & yes \\
3\_004 & FGMRES(50), GAMG with trunk basis $k=72$, SOR, post-smoothing & 33 & 3.65 & 13.9 & $2.86\times10^{10}$ & $4.19\times10^{-9}$ & yes \\
3\_005 & FGMRES(50), GAMG with trunk basis $k=84$, SOR, post-smoothing & 33 & 3.69 & 14.0 & $2.86\times10^{10}$ & $3.82\times10^{-9}$ & yes \\
3\_006 & FGMRES(50), GAMG with trunk basis $k=90$, SOR, post-smoothing & 33 & 3.66 & 14.1 & $2.86\times10^{10}$ & $2.45\times10^{-9}$ & yes \\
3\_007 & FGMRES(50), GAMG with trunk basis $k=98$, SOR, post-smoothing & 33 & 3.67 & 14.2 & $2.86\times10^{10}$ & $3.55\times10^{-9}$ & yes \\
3\_008 & FGMRES(50), GAMG with trunk basis $k=106$, SOR, post-smoothing & 33 & 3.68 & 14.3 & $2.87\times10^{10}$ & $1.25\times10^{-9}$ & yes \\
3\_009 & FGMRES(50), GAMG with trunk basis $k=114$, SOR, post-smoothing & 32 & 3.58 & 14.4 & $2.78\times10^{10}$ & $2.13\times10^{-9}$ & yes \\
3\_010 & FGMRES(50), GAMG with trunk basis $k=122$, SOR, post-smoothing & 32 & 3.58 & 14.5 & $2.78\times10^{10}$ & $6.28\times10^{-9}$ & yes \\
3\_011 & FGMRES(50), GAMG with trunk basis $k=48$, SOR, post-smoothing & 36 & 3.90 & 13.2 & $3.07\times10^{10}$ & $3.58\times10^{-9}$ & yes \\
3\_012 & FGMRES(50), GAMG with trunk basis $k=64$, SOR, post-smoothing & 34 & 3.76 & 13.8 & $2.94\times10^{10}$ & $8.93\times10^{-10}$ & yes \\
3\_013 & FGMRES(50), GAMG with trunk basis $k=76$, SOR, post-smoothing & 33 & 3.66 & 13.9 & $2.86\times10^{10}$ & $4.57\times10^{-9}$ & yes \\
3\_014 & FGMRES(50), GAMG with trunk basis $k=88$, SOR, post-smoothing & 33 & 3.67 & 14.1 & $2.86\times10^{10}$ & $3.70\times10^{-9}$ & yes \\
3\_015 & FGMRES(50), GAMG with trunk basis $k=100$, SOR, post-smoothing & 33 & 3.69 & 14.3 & $2.87\times10^{10}$ & $2.89\times10^{-10}$ & yes \\
3\_016 & FGMRES(50), GAMG with trunk basis $k=112$, SOR, post-smoothing & 32 & 3.58 & 14.4 & $2.78\times10^{10}$ & $6.24\times10^{-10}$ & yes \\
3\_017 & FGMRES(50), GAMG with trunk basis $k=120$, SOR, post-smoothing & 32 & 3.57 & 14.5 & $2.78\times10^{10}$ & $6.28\times10^{-9}$ & yes \\
3\_018 & FGMRES(50), GAMG with trunk basis $k=60$, SOR, post-smoothing & 33 & 3.67 & 13.8 & $2.85\times10^{10}$ & $4.69\times10^{-10}$ & yes \\
3\_019 & FGMRES(50), GAMG with trunk basis $k=70$, SOR, post-smoothing & 34 & 3.81 & 13.9 & $2.95\times10^{10}$ & $1.30\times10^{-9}$ & yes \\
3\_020 & FGMRES(50), GAMG with trunk basis $k=104$, SOR, post-smoothing & 33 & 3.67 & 14.3 & $2.87\times10^{10}$ & $8.90\times10^{-10}$ & yes
\end{longtable}

\begin{longtable}{@{}lp{\dimexpr\textwidth-9.9cm\relax}rrccc@{}}
\caption{All candidates of the multi-agent system for the two-dimensional thermoelastic problem. The plan of the third generation was rejected by the validator. Iterations, online time, and MAC are totals over the 4,000 steps, and the errors are the final relative $L^2$ errors of the temperature and of the displacement.}\label{tab:si_t2}\\
\toprule
ID & Configuration & Iterations & Online (s) & MAC & Error in $T$ & Error in $\uv$ \\
\midrule
\endfirsthead
\toprule
ID & Configuration & Iterations & Online (s) & MAC & Error in $T$ & Error in $\uv$ \\
\midrule
\endhead
\bottomrule
\endfoot
1\_001 & FGMRES(50), full Schur, ILU, balancing deflation of $\uv$, $k=20$ & 194,593 & 41.6 & $2.94\times10^{11}$ & $1.5\times10^{-4}$ & $3.0\times10^{-2}$ \\
1\_002 & FGMRES(50), full Schur, ILU, balancing deflation of $\uv$, $k=40$ & 194,312 & 47.0 & $3.80\times10^{11}$ & $1.0\times10^{-4}$ & $2.4\times10^{-2}$ \\
1\_003 & FGMRES(50), full Schur, ILU, balancing deflation of $T$, $k=20$ & 1,638,060 & 343.2 & $2.34\times10^{12}$ & $1.2\times10^{-5}$ & $1.0\times10^{-2}$ \\
1\_004 & FGMRES(50), full Schur, ILU, balancing deflation of $T$, $k=40$ & 1,620,714 & 404.8 & $3.04\times10^{12}$ & $2.1\times10^{-5}$ & $4.7\times10^{-2}$ \\
1\_005 & FGMRES(50), full Schur, Jacobi, balancing deflation of $\uv$, $k=20$ & 1,387,224 & 237.7 & $1.92\times10^{12}$ & $3.3\times10^{-5}$ & $2.8\times10^{-2}$ \\
1\_006 & FGMRES(50), full Schur, Jacobi, balancing deflation of $\uv$, $k=40$ & 1,354,614 & 286.9$^{\ast}$ & $2.48\times10^{12}$ & $1.8\times10^{-5}$ & $1.7\times10^{-2}$ \\
1\_007 & FGMRES(50), full Schur, GAMG, balancing deflation of $\uv$, $k=20$ & 40,453 & 16.8 & $1.19\times10^{11}$ & $5.5\times10^{-5}$ & $9.9\times10^{-3}$ \\
1\_008 & FGMRES(50), full Schur, GAMG, balancing deflation of $\uv$, $k=40$ & 40,408 & 18.3 & $1.37\times10^{11}$ & $4.4\times10^{-5}$ & $8.3\times10^{-3}$ \\
1\_009 & FGMRES(50), full Schur, ILU, balancing deflation of $\uv$, $k=20$, 2 clusters & 160,408 & 32.4 & $2.32\times10^{11}$ & $1.1\times10^{-4}$ & $2.2\times10^{-2}$ \\
1\_010 & FGMRES(50), full Schur, ILU, balancing deflation of $\uv$, $k=40$, 2 clusters & 156,532 & 37.4 & $2.97\times10^{11}$ & $1.4\times10^{-4}$ & $3.7\times10^{-2}$ \\
1\_011 & FGMRES(50), full Schur, ILU, balancing deflation of $\uv$, $k=20$, 4 clusters & 117,370 & 23.1 & $1.61\times10^{11}$ & $1.1\times10^{-4}$ & $2.2\times10^{-2}$ \\
1\_012 & FGMRES(50), full Schur, ILU, balancing deflation of $\uv$, $k=40$, 4 clusters & 118,064 & 28.3 & $2.19\times10^{11}$ & $1.5\times10^{-4}$ & $2.2\times10^{-2}$ \\
1\_013 & FGMRES(50), full Schur, ILU, balancing deflation of $T$ and $\uv$, $k=20$ & 195,288 & 54.9 & $4.02\times10^{11}$ & $9.7\times10^{-5}$ & $2.4\times10^{-2}$ \\
1\_014 & FGMRES(50), full Schur, ILU, balancing deflation of $T$ and $\uv$, $k=40$ & 193,770 & 69.4 & $5.72\times10^{11}$ & $2.1\times10^{-4}$ & $3.3\times10^{-2}$ \\
1\_015 & FGMRES(50), full Schur, ILU, balancing deflation of $T$ and $\uv$, $k=20$, 2 clusters & 159,657 & 44.6 & $3.19\times10^{11}$ & $1.4\times10^{-4}$ & $3.0\times10^{-2}$ \\
1\_016 & FGMRES(50), full Schur, ILU, balancing deflation of $T$ and $\uv$, $k=40$, 2 clusters & 156,548 & 56.1 & $4.59\times10^{11}$ & $1.3\times10^{-4}$ & $2.7\times10^{-2}$ \\
1\_017 & FGMRES(50), full Schur, ILU, balancing deflation of $T$ and $\uv$, $k=20$, 4 clusters & 118,446 & 33.0 & $2.30\times10^{11}$ & $1.5\times10^{-4}$ & $2.7\times10^{-2}$ \\
1\_018 & FGMRES(50), full Schur, ILU, balancing deflation of $T$ and $\uv$, $k=40$, 4 clusters & 117,015 & 42.6 & $3.47\times10^{11}$ & $1.2\times10^{-4}$ & $2.3\times10^{-2}$ \\
1\_019 & FGMRES(50), full Schur, GAMG, balancing deflation of $T$, $k=20$ & 40,230 & 22.8 & $1.58\times10^{11}$ & $4.2\times10^{-5}$ & $8.0\times10^{-3}$ \\
1\_020 & FGMRES(50), full Schur, GAMG, balancing deflation of $T$, $k=40$ & 40,228 & 27.5 & $2.12\times10^{11}$ & $4.2\times10^{-5}$ & $8.0\times10^{-3}$ \\
2\_001 & GMRES, full Schur, ILU, balancing deflation of $T$, $k=20$ & 350,928 & 69.9 & $4.51\times10^{11}$ & $1.2\times10^{-2}$ & 26.0 \\
2\_002 & GMRES, full Schur, ILU, balancing deflation of $T$, $k=20$, 2 clusters & 338,993 & 67.8 & $4.39\times10^{11}$ & $2.0\times10^{-2}$ & 18.3 \\
2\_003 & GMRES, full Schur, ILU, balancing deflation of $T$, $k=40$ & 333,537 & 79.4 & $5.85\times10^{11}$ & $2.0\times10^{-2}$ & 41.7 \\
2\_004 & GMRES, full Schur, ILU, balancing deflation of $T$, $k=40$, 2 clusters & 330,070 & 80.5 & $5.91\times10^{11}$ & $2.9\times10^{-2}$ & 92.7 \\
2\_005 & GMRES, full Schur, GAMG, balancing deflation of $T$, $k=20$, 4 clusters & 30,506 & 17.3 & $1.37\times10^{11}$ & $8.2\times10^{-3}$ & 0.50 \\
2\_006 & GMRES, full Schur, GAMG, balancing deflation of $T$, $k=40$, 2 clusters & 30,550 & 21.1 & $1.85\times10^{11}$ & $7.6\times10^{-3}$ & 0.58 \\
2\_007 & GMRES, full Schur, ILU, balancing deflation of $\uv$, $k=20$ & 210,275 & 41.4 & $2.89\times10^{11}$ & $2.5\times10^{-2}$ & 14.5 \\
2\_008 & GMRES, full Schur, ILU, balancing deflation of $\uv$, $k=20$, 2 clusters & 108,039 & 20.7 & $1.48\times10^{11}$ & $1.2\times10^{-2}$ & 3.37 \\
2\_009 & GMRES, full Schur, ILU, balancing deflation of $\uv$, $k=40$ & 209,375 & 48.5 & $3.85\times10^{11}$ & $1.6\times10^{-2}$ & 12.4 \\
2\_010 & GMRES, full Schur, ILU, balancing deflation of $\uv$, $k=40$, 2 clusters & 101,194 & 23.0 & $1.87\times10^{11}$ & $1.0\times10^{-2}$ & 1.93 \\
2\_011 & GMRES, full Schur, GAMG, balancing deflation of $\uv$, $k=20$, 4 clusters & 32,000 & 13.2 & $1.04\times10^{11}$ & $6.8\times10^{-3}$ & 0.91 \\
2\_012 & GMRES, full Schur, GAMG, balancing deflation of $\uv$, $k=40$, 2 clusters & 33,582 & 15.1 & $1.26\times10^{11}$ & $7.9\times10^{-3}$ & 0.72 \\
2\_013 & GMRES, full Schur, ILU, balancing deflation of $T$ and $\uv$, $k=20$ & 182,478 & 50.0 & $3.53\times10^{11}$ & $2.3\times10^{-2}$ & 6.83 \\
2\_014 & GMRES, full Schur, GAMG, balancing deflation of $T$ and $\uv$, $k=20$, 4 clusters & 29,805 & 19.3 & $1.56\times10^{11}$ & $5.7\times10^{-3}$ & 0.46 \\
2\_015 & GMRES, full Schur, ILU, balancing deflation of $T$ and $\uv$, $k=40$, 2 clusters & 103,393 & 36.2 & $3.02\times10^{11}$ & $9.0\times10^{-3}$ & 2.26 \\
2\_016 & GMRES, full Schur, ILU, balancing deflation of $T$ and $\uv$, $k=40$, 4 clusters & 79,522 & 28.4 & $2.40\times10^{11}$ & $1.4\times10^{-2}$ & 1.91 \\
2\_017 & GMRES, full Schur, GAMG, balancing deflation of $T$ and $\uv$, $k=20$ & 30,886 & 19.8 & $1.57\times10^{11}$ & $8.3\times10^{-3}$ & 0.93 \\
2\_018 & GMRES, full Schur, GAMG, balancing deflation of $T$ and $\uv$, $k=40$, 2 clusters & 29,794 & 23.8 & $2.18\times10^{11}$ & $5.2\times10^{-3}$ & 0.40 \\
2\_019 & GMRES, full Schur, ILU, balancing deflation of $T$, $k=40$, 4 clusters & 348,864 & 86.1 & $6.53\times10^{11}$ & $2.4\times10^{-2}$ & 61.3 \\
2\_020 & GMRES, full Schur, GAMG, balancing deflation of $T$ and $\uv$, $k=40$, 4 clusters & 29,832 & 24.6 & $2.27\times10^{11}$ & $6.0\times10^{-3}$ & 0.39
\end{longtable}

\begin{longtable}{@{}lp{\dimexpr\textwidth-9.9cm\relax}rrccc@{}}
\caption{All candidates of the single agent for the two-dimensional thermoelastic problem. The plan of the first generation was rejected by the validator. Iterations, online time, and MAC are totals over the 4,000 steps, and the errors are the final relative $L^2$ errors of the temperature and of the displacement.}\label{tab:si_t2_sa}\\
\toprule
ID & Configuration & Iterations & Online (s) & MAC & Error in $T$ & Error in $\uv$ \\
\midrule
\endfirsthead
\toprule
ID & Configuration & Iterations & Online (s) & MAC & Error in $T$ & Error in $\uv$ \\
\midrule
\endhead
\bottomrule
\endfoot
2\_001 & FGMRES, full Schur, Jacobi, balancing deflation of $\uv$, $k=32$ & 1,628,396 & 290.1 & $2.44\times10^{12}$ & $5.0\times10^{-5}$ & $1.0\times10^{-2}$ \\
2\_002 & FGMRES, full Schur, Jacobi, additive deflation of $\uv$, $k=64$ & 1,848,704 & 246.7 & $1.94\times10^{12}$ & $1.7\times10^{-5}$ & $9.0\times10^{-3}$ \\
2\_003 & FGMRES, full Schur, Jacobi, balancing deflation of $\uv$, $k=64$ & 1,637,076 & 321.6 & $2.75\times10^{12}$ & $4.8\times10^{-5}$ & $1.0\times10^{-2}$ \\
2\_004 & FGMRES, full Schur, Jacobi, balancing deflation of $\uv$, $k=32$ & 1,628,396 & 291.6 & $2.44\times10^{12}$ & $5.0\times10^{-5}$ & $1.0\times10^{-2}$ \\
2\_005 & FGMRES, full Schur, Jacobi, balancing deflation of $\uv$, $k=32$ & 1,628,396 & 291.8 & $2.44\times10^{12}$ & $5.0\times10^{-5}$ & $1.0\times10^{-2}$ \\
2\_006 & FGMRES, full Schur, Jacobi, additive deflation of $\uv$, $k=32$ & 1,801,198 & 223.2 & $1.73\times10^{12}$ & $4.9\times10^{-5}$ & $3.0\times10^{-2}$ \\
2\_007 & FGMRES, full Schur, Jacobi, balancing deflation of $\uv$, $k=32$ & 1,628,396 & 292.2 & $2.44\times10^{12}$ & $5.0\times10^{-5}$ & $1.0\times10^{-2}$ \\
2\_008 & FGMRES, full Schur, Jacobi, balancing deflation of $\uv$, $k=64$ & 1,637,076 & 322.9 & $2.75\times10^{12}$ & $4.8\times10^{-5}$ & $1.0\times10^{-2}$ \\
2\_009 & FGMRES, full Schur, Jacobi, balancing deflation of $\uv$, $k=32$ & 1,628,396 & 290.5 & $2.44\times10^{12}$ & $5.0\times10^{-5}$ & $1.0\times10^{-2}$ \\
2\_010 & FGMRES, full Schur, Jacobi, balancing deflation of $\uv$, $k=48$ & 1,637,076 & 322.0$^{\ast}$ & $2.75\times10^{12}$ & $4.8\times10^{-5}$ & $1.0\times10^{-2}$ \\
2\_011 & FGMRES, full Schur, Jacobi, balancing deflation of $\uv$, $k=64$ & 1,637,076 & 323.1 & $2.75\times10^{12}$ & $4.8\times10^{-5}$ & $1.0\times10^{-2}$ \\
2\_012 & FGMRES, full Schur, Jacobi, balancing deflation of $\uv$, $k=16$ & 1,677,397 & 253.1 & $1.92\times10^{12}$ & $4.6\times10^{-5}$ & $2.0\times10^{-2}$ \\
2\_013 & FGMRES, full Schur, Jacobi, balancing deflation of $\uv$, $k=32$ & 1,628,396 & 293.2 & $2.44\times10^{12}$ & $5.0\times10^{-5}$ & $1.0\times10^{-2}$ \\
2\_014 & FGMRES, full Schur, Jacobi, balancing deflation of $\uv$, $k=32$ & 1,628,396 & 291.7 & $2.44\times10^{12}$ & $5.0\times10^{-5}$ & $1.0\times10^{-2}$ \\
2\_015 & FGMRES, full Schur, Jacobi, balancing deflation of $\uv$, $k=16$ & 1,677,397 & 256.3 & $1.92\times10^{12}$ & $4.6\times10^{-5}$ & $2.0\times10^{-2}$ \\
2\_016 & FGMRES, full Schur, Jacobi, balancing deflation of $\uv$, $k=128$ & 1,637,076 & 324.5 & $2.75\times10^{12}$ & $4.8\times10^{-5}$ & $1.0\times10^{-2}$ \\
2\_017 & FGMRES, full Schur, Jacobi, balancing deflation of $\uv$, $k=32$ & 1,628,396 & 291.7 & $2.44\times10^{12}$ & $5.0\times10^{-5}$ & $1.0\times10^{-2}$ \\
2\_018 & FGMRES, full Schur, Jacobi, additive deflation of $\uv$, $k=32$ & 1,801,198 & 223.5 & $1.73\times10^{12}$ & $4.9\times10^{-5}$ & $3.0\times10^{-2}$ \\
2\_019 & FGMRES, full Schur, Jacobi, balancing deflation of $\uv$, $k=64$ & 1,637,076 & 322.3 & $2.75\times10^{12}$ & $4.8\times10^{-5}$ & $1.0\times10^{-2}$ \\
2\_020 & FGMRES, full Schur, Jacobi, balancing deflation of $\uv$, $k=8$ & 1,724,158 & 227.6 & $1.67\times10^{12}$ & $4.4\times10^{-5}$ & $1.9\times10^{-2}$ \\
3\_001 & GMRES, full Schur, Jacobi, balancing deflation of $\uv$, $k=24$, tolerance $10^{-9}$ & 483,173 & 81.1 & $6.50\times10^{11}$ & $1.9\times10^{-2}$ & 11.0 \\
3\_002 & GMRES, full Schur, Jacobi, additive deflation of $\uv$, $k=64$, tolerance $10^{-9}$ & 525,240 & 70.7 & $5.57\times10^{11}$ & $6.6\times10^{-3}$ & 4.47 \\
3\_003 & GMRES, full Schur, Jacobi, balancing deflation of $\uv$, $k=48$, tolerance $10^{-9}$ & 483,560 & 97.5 & $8.29\times10^{11}$ & $4.3\times10^{-2}$ & 14.0 \\
3\_004 & GMRES, full Schur, Jacobi, additive deflation of $\uv$, $k=64$, tolerance $10^{-9}$ & 525,240 & 71.4 & $5.57\times10^{11}$ & $6.6\times10^{-3}$ & 4.47 \\
3\_005 & GMRES, full Schur, Jacobi, balancing deflation of $\uv$, $k=48$, tolerance $10^{-9}$ & 483,560 & 98.5 & $8.29\times10^{11}$ & $4.3\times10^{-2}$ & 14.0 \\
3\_006 & GMRES, full Schur, Jacobi, additive deflation of $\uv$, $k=32$, tolerance $10^{-9}$ & 514,370 & 65.4 & $4.97\times10^{11}$ & $1.0\times10^{-2}$ & 2.63 \\
3\_007 & GMRES, full Schur, Jacobi, balancing deflation of $\uv$, $k=64$, tolerance $10^{-9}$ & 483,560 & 99.2 & $8.29\times10^{11}$ & $4.3\times10^{-2}$ & 14.0 \\
3\_008 & GMRES, full Schur, Jacobi, additive deflation of $\uv$, $k=48$, tolerance $10^{-9}$ & 525,240 & 71.3 & $5.57\times10^{11}$ & $6.6\times10^{-3}$ & 4.47 \\
3\_009 & GMRES, full Schur, Jacobi, balancing deflation of $\uv$, $k=32$, tolerance $10^{-9}$ & 509,730 & 93.2 & $7.78\times10^{11}$ & $4.3\times10^{-2}$ & 9.18 \\
3\_010 & GMRES, full Schur, Jacobi, additive deflation of $\uv$, $k=48$, tolerance $10^{-9}$ & 525,240 & 73.0 & $5.57\times10^{11}$ & $6.6\times10^{-3}$ & 4.47 \\
3\_011 & GMRES, full Schur, Jacobi, balancing deflation of $\uv$, $k=64$, tolerance $10^{-9}$ & 483,560 & 95.8 & $8.29\times10^{11}$ & $4.3\times10^{-2}$ & 14.0 \\
3\_012 & GMRES, full Schur, Jacobi, additive deflation of $\uv$, $k=32$, tolerance $10^{-9}$ & 514,370 & 64.5 & $4.97\times10^{11}$ & $1.0\times10^{-2}$ & 2.63 \\
3\_013 & GMRES, full Schur, Jacobi, balancing deflation of $\uv$, $k=48$, tolerance $10^{-9}$ & 483,560 & 98.7 & $8.29\times10^{11}$ & $4.3\times10^{-2}$ & 14.0 \\
3\_014 & GMRES, full Schur, Jacobi, additive deflation of $\uv$, $k=64$, tolerance $10^{-9}$ & 525,240 & 70.9 & $5.57\times10^{11}$ & $6.6\times10^{-3}$ & 4.47 \\
3\_015 & GMRES, full Schur, Jacobi, balancing deflation of $\uv$, $k=32$, tolerance $10^{-9}$ & 509,730 & 94.0 & $7.78\times10^{11}$ & $4.3\times10^{-2}$ & 9.18 \\
3\_016 & GMRES, full Schur, Jacobi, additive deflation of $\uv$, $k=64$, tolerance $10^{-9}$ & 525,240 & 71.4 & $5.57\times10^{11}$ & $6.6\times10^{-3}$ & 4.47 \\
3\_017 & GMRES, full Schur, Jacobi, balancing deflation of $\uv$, $k=48$, tolerance $10^{-9}$ & 483,560 & 97.4 & $8.29\times10^{11}$ & $4.3\times10^{-2}$ & 14.0 \\
3\_018 & GMRES, full Schur, Jacobi, additive deflation of $\uv$, $k=32$, tolerance $10^{-9}$ & 514,370 & 64.7 & $4.97\times10^{11}$ & $1.0\times10^{-2}$ & 2.63 \\
3\_019 & GMRES, full Schur, Jacobi, balancing deflation of $\uv$, $k=64$, tolerance $10^{-9}$ & 483,560 & 96.9 & $8.29\times10^{11}$ & $4.3\times10^{-2}$ & 14.0 \\
3\_020 & GMRES, full Schur, Jacobi, additive deflation of $\uv$, $k=48$, tolerance $10^{-9}$ & 525,240 & 70.8 & $5.57\times10^{11}$ & $6.6\times10^{-3}$ & 4.47
\end{longtable}

\begin{longtable}{@{}lp{\dimexpr\textwidth-9.9cm\relax}rrccc@{}}
\caption{All candidates of the multi-agent system for the three-dimensional thermoelastic problem. Iterations, online time, and MAC are totals over the 4,000 steps, and the errors are the final relative $L^2$ errors of the temperature and of the displacement.}\label{tab:si_t3}\\
\toprule
ID & Configuration & Iterations & Online (s) & MAC & Error in $T$ & Error in $\uv$ \\
\midrule
\endfirsthead
\toprule
ID & Configuration & Iterations & Online (s) & MAC & Error in $T$ & Error in $\uv$ \\
\midrule
\endhead
\bottomrule
\endfoot
1\_001 & GMRES, full Schur, ILU, balancing deflation of $\uv$, $k=10$ & 154,587 & 653.5 & $5.77\times10^{12}$ & $1.1\times10^{-2}$ & 3.00 \\
1\_002 & GMRES, full Schur, ILU, balancing deflation of $\uv$, $k=20$ & 155,932 & 715.2 & $6.30\times10^{12}$ & $1.3\times10^{-2}$ & 5.40 \\
1\_003 & GMRES, full Schur, ILU, balancing deflation of $\uv$, $k=40$ & 138,867 & 724.2 & $6.51\times10^{12}$ & $2.3\times10^{-2}$ & 70.2 \\
1\_004 & GMRES, full Schur, ILU, additive deflation of $\uv$, $k=20$ & 158,984 & 542.2 & $4.18\times10^{12}$ & $1.0\times10^{-2}$ & 4.26 \\
1\_005 & GMRES, full Schur, ICC, balancing deflation of $\uv$, $k=20$ & 155,932 & 753.9 & $6.30\times10^{12}$ & $1.3\times10^{-2}$ & 5.40 \\
1\_006 & GMRES, full Schur, ILU, balancing deflation of $T$, $k=10$ & 337,601 & 1,254.2 & $9.38\times10^{12}$ & $1.9\times10^{-2}$ & 25.1 \\
1\_007 & GMRES, full Schur, ILU, balancing deflation of $T$, $k=20$ & 323,878 & 1,271.3 & $9.66\times10^{12}$ & $1.7\times10^{-2}$ & 27.4 \\
1\_008 & GMRES, full Schur, ILU, balancing deflation of $T$, $k=40$ & 334,965 & 1,437.6 & $1.14\times10^{13}$ & $1.5\times10^{-2}$ & 12.5 \\
1\_009 & GMRES, full Schur, ILU, additive deflation of $T$, $k=20$ & 279,383 & 918.3 & $7.07\times10^{12}$ & $3.9\times10^{-2}$ & 21.1 \\
1\_010 & GMRES, full Schur, ICC, balancing deflation of $T$, $k=20$ & 349,728 & 1,463.4 & $1.04\times10^{13}$ & $2.6\times10^{-2}$ & 36.1 \\
1\_011 & GMRES, full Schur, ILU, balancing deflation of $T$ and $\uv$, $k=10$ & 155,676 & 774.8 & $6.53\times10^{12}$ & $1.5\times10^{-2}$ & 3.45 \\
1\_012 & GMRES, full Schur, ILU, balancing deflation of $T$ and $\uv$, $k=20$ & 153,483 & 842.2 & $7.23\times10^{12}$ & $9.3\times10^{-3}$ & 2.95 \\
1\_013 & GMRES, full Schur, ILU, balancing deflation of $T$ and $\uv$, $k=40$ & 179,146 & 1,160.7 & $1.03\times10^{13}$ & $2.0\times10^{-2}$ & 65.5 \\
1\_014 & GMRES, full Schur, ILU, additive deflation of $T$ and $\uv$, $k=20$ & 139,970 & 515.8 & $4.00\times10^{12}$ & $2.5\times10^{-2}$ & 8.07 \\
1\_015 & GMRES, full Schur, ICC, balancing deflation of $T$ and $\uv$, $k=20$ & 153,483 & 882.1 & $7.23\times10^{12}$ & $9.3\times10^{-3}$ & 2.95 \\
1\_016 & FGMRES, full Schur, ILU, balancing deflation of $\uv$, $k=10$ & 1,354,075 & 5,777.1 & $4.97\times10^{13}$ & $1.6\times10^{-5}$ & $5.0\times10^{-2}$ \\
1\_017 & FGMRES, full Schur, ICC, balancing deflation of $\uv$, $k=20$ & 531,947 & 2,557.6 & $2.11\times10^{13}$ & $3.8\times10^{-5}$ & 2.00 \\
1\_018 & FGMRES, full Schur, ILU, balancing deflation of $T$, $k=10$ & \multicolumn{5}{c}{stopped at step 100, projected time above the limit} \\
1\_019 & FGMRES, full Schur, ICC, balancing deflation of $T$, $k=20$ & \multicolumn{5}{c}{stopped at step 100, projected time above the limit} \\
1\_020 & FGMRES, full Schur, ILU, balancing deflation of $T$ and $\uv$, $k=20$ & 532,740 & 2,914.0 & $2.45\times10^{13}$ & $2.4\times10^{-5}$ & 4.79 \\
2\_001 & FGMRES, full Schur, ILU, balancing deflation of $\uv$, $k=10$ & 1,354,075 & 5,641.9$^{\ast}$ & $4.97\times10^{13}$ & $1.6\times10^{-5}$ & $5.0\times10^{-2}$ \\
2\_002 & FGMRES, full Schur, ILU, balancing deflation of $\uv$, $k=15$ & 518,731 & 2,388.6 & $1.98\times10^{13}$ & $3.6\times10^{-5}$ & 0.46 \\
2\_003 & FGMRES, full Schur, ILU, balancing deflation of $\uv$, $k=20$ & 542,692 & 2,457.9 & $2.15\times10^{13}$ & $3.9\times10^{-5}$ & 0.55 \\
2\_004 & FGMRES, full Schur, ICC, balancing deflation of $\uv$, $k=20$ & 531,947 & 2,548.9 & $2.11\times10^{13}$ & $3.8\times10^{-5}$ & 2.00 \\
2\_005 & FGMRES, full Schur, GAMG, balancing deflation of $\uv$, $k=20$ & 55,850 & 588.1 & $5.07\times10^{12}$ & $9.0\times10^{-5}$ & 0.32 \\
2\_006 & FGMRES, full Schur, ILU, balancing deflation of $T$ and $\uv$, $k=15$ & 514,552 & 2,684.1 & $2.24\times10^{13}$ & $5.1\times10^{-5}$ & 1.03 \\
2\_007 & FGMRES, full Schur, ILU, balancing deflation of $T$ and $\uv$, $k=20$ & 532,740 & 2,844.7 & $2.45\times10^{13}$ & $2.4\times10^{-5}$ & 4.79 \\
2\_008 & FGMRES, full Schur, ICC, balancing deflation of $T$ and $\uv$, $k=20$ & 526,973 & 2,976.2 & $2.42\times10^{13}$ & $2.7\times10^{-5}$ & 1.84 \\
2\_009 & FGMRES, full Schur, GAMG, balancing deflation of $T$ and $\uv$, $k=20$ & 55,755 & 732.2 & $6.13\times10^{12}$ & $1.0\times10^{-4}$ & 0.13 \\
2\_010 & FGMRES, full Schur, ILU, additive deflation of $T$ and $\uv$, $k=20$ & 661,999 & 2,419.6 & $1.86\times10^{13}$ & $6.1\times10^{-5}$ & 3.18 \\
2\_011 & FGMRES, full Schur, ILU, balancing deflation of $\uv$, $k=20$, tolerance $10^{-9}$ & 702,214 & 3,292.6 & $2.78\times10^{13}$ & $3.4\times10^{-6}$ & 5.19 \\
2\_012 & FGMRES, full Schur, ILU, balancing deflation of $\uv$, $k=20$, tolerance $10^{-7}$ & 407,314 & 1,835.4 & $1.61\times10^{13}$ & $6.2\times10^{-4}$ & 2.02 \\
2\_013 & FGMRES, full Schur, ILU, balancing deflation of $T$ and $\uv$, $k=20$, tolerance $10^{-9}$ & 692,590 & 3,745.0 & $3.19\times10^{13}$ & $3.3\times10^{-6}$ & 3.76 \\
2\_014 & FGMRES, full Schur, ILU, balancing deflation of $T$ and $\uv$, $k=20$, tolerance $10^{-7}$ & 394,840 & 2,148.1 & $1.81\times10^{13}$ & $3.9\times10^{-4}$ & 1.44 \\
2\_015 & FGMRES, full Schur, ILU, additive deflation of $T$ and $\uv$, $k=20$ & 661,999 & 2,386.5 & $1.86\times10^{13}$ & $6.1\times10^{-5}$ & 3.18 \\
2\_016 & FGMRES, full Schur, ILU, balancing deflation of $\uv$, $k=20$, 32 clusters & 89,872 & 404.9 & $3.51\times10^{12}$ & $1.6\times10^{-4}$ & 1.27 \\
2\_017 & FGMRES, full Schur, ILU, balancing deflation of $T$ and $\uv$, $k=20$, 32 clusters & 84,159 & 458.7 & $3.88\times10^{12}$ & $1.2\times10^{-4}$ & $4.2\times10^{-2}$ \\
2\_018 & FGMRES, full Schur, ICC, balancing deflation of $T$ and $\uv$, $k=20$, 32 clusters & 84,159 & 476.5 & $3.88\times10^{12}$ & $1.2\times10^{-4}$ & $4.2\times10^{-2}$ \\
2\_019 & FGMRES, full Schur, GAMG, balancing deflation of $\uv$, $k=20$, 32 clusters & 45,504 & 489.0$^{\ast}$ & $4.14\times10^{12}$ & $2.8\times10^{-4}$ & 0.36 \\
2\_020 & FGMRES, full Schur, GAMG, balancing deflation of $T$ and $\uv$, $k=20$, 32 clusters & 43,987 & 595.9$^{\ast}$ & $4.95\times10^{12}$ & $1.3\times10^{-4}$ & 0.93 \\
3\_001 & FGMRES, full Schur, ILU and GAMG, balancing deflation of $T$ and $\uv$, $k=20$, 32 clusters & 41,822 & 498.2$^{\ast}$ & $4.13\times10^{12}$ & $3.3\times10^{-4}$ & 0.17 \\
3\_002 & FGMRES, full Schur, ICC and GAMG, balancing deflation of $T$ and $\uv$, $k=20$, 32 clusters & 41,822 & 490.3 & $4.13\times10^{12}$ & $3.3\times10^{-4}$ & 0.17 \\
3\_003 & FGMRES, full Schur, ILU, balancing deflation of $T$ and $\uv$, $k=15$, 32 clusters & 94,825 & 518.0 & $4.11\times10^{12}$ & $2.3\times10^{-4}$ & 7.38 \\
3\_004 & FGMRES, full Schur, ILU, balancing deflation of $T$ and $\uv$, $k=20$, 64 clusters & 78,637 & 458.3 & $3.85\times10^{12}$ & $2.2\times10^{-4}$ & 1.42 \\
3\_005 & FGMRES, full Schur, ILU, balancing deflation of $T$ and $\uv$, $k=20$, 32 clusters, tolerance $10^{-9}$ & 96,559 & 548.9 & $4.48\times10^{12}$ & $2.0\times10^{-5}$ & 1.23 \\
3\_006 & FGMRES, full Schur, ILU and GAMG, balancing deflation of $\uv$, $k=15$, 32 clusters & 48,000 & 438.0 & $3.64\times10^{12}$ & $2.7\times10^{-5}$ & 0.21 \\
3\_007 & FGMRES, full Schur, ILU and GAMG, balancing deflation of $\uv$, $k=20$, 64 clusters & 46,582 & 438.8 & $3.65\times10^{12}$ & $1.5\times10^{-4}$ & 1.08 \\
3\_008 & FGMRES, full Schur, ILU and GAMG, balancing deflation of $\uv$, $k=20$, 32 clusters & 48,000 & 443.1 & $3.72\times10^{12}$ & $2.5\times10^{-5}$ & 0.71 \\
3\_009 & FGMRES, full Schur, ILU, balancing deflation of $T$ and $\uv$, $k=25$, 32 clusters & 83,735 & 487.1 & $4.13\times10^{12}$ & $1.5\times10^{-4}$ & 0.54 \\
3\_010 & FGMRES, full Schur, ILU and GAMG, balancing deflation of $\uv$, $k=20$, 32 clusters & 48,000 & 439.5 & $3.72\times10^{12}$ & $2.5\times10^{-5}$ & 0.71 \\
3\_011 & FGMRES, full Schur, ILU and GAMG, additive deflation of $\uv$, $k=20$, 32 clusters & 67,983 & 585.5 & $4.36\times10^{12}$ & $4.9\times10^{-5}$ & 1.82 \\
3\_012 & FGMRES, full Schur, GAMG, balancing deflation of $\uv$, $k=20$, 32 clusters & 45,504 & 487.7$^{\ast}$ & $4.14\times10^{12}$ & $2.8\times10^{-4}$ & 0.36 \\
3\_013 & FGMRES, full Schur, GAMG, balancing deflation of $T$ and $\uv$, $k=15$, 32 clusters & 44,345 & 573.8$^{\ast}$ & $4.74\times10^{12}$ & $1.3\times10^{-4}$ & 1.25 \\
3\_014 & FGMRES, full Schur, ICC, balancing deflation of $\uv$, $k=20$, 32 clusters & 89,957 & 443.7 & $3.52\times10^{12}$ & $1.5\times10^{-4}$ & 0.62 \\
3\_015 & FGMRES, full Schur, ICC, balancing deflation of $\uv$, $k=15$, 32 clusters & 98,771 & 474.6 & $3.72\times10^{12}$ & $1.5\times10^{-4}$ & 2.56 \\
3\_016 & FGMRES, full Schur, ILU and GAMG, balancing deflation of $\uv$, $k=15$, 64 clusters & 48,000 & 434.9$^{\ast}$ & $3.66\times10^{12}$ & $6.1\times10^{-7}$ & 0.60 \\
3\_017 & FGMRES, full Schur, ICC, balancing deflation of $\uv$, $k=20$, 64 clusters & 82,746 & 409.3$^{\ast}$ & $3.30\times10^{12}$ & $1.9\times10^{-4}$ & 2.93 \\
3\_018 & FGMRES, full Schur, ICC, additive deflation of $\uv$, $k=20$, 32 clusters & 137,756 & 516.7 & $3.58\times10^{12}$ & $9.0\times10^{-5}$ & 0.14 \\
3\_019 & FGMRES, full Schur, GAMG, additive deflation of $T$ and $\uv$, $k=20$, 32 clusters & 84,000 & 880.5$^{\ast}$ & $7.19\times10^{12}$ & $1.3\times10^{-5}$ & 0.99 \\
3\_020 & FGMRES, full Schur, GAMG, balancing deflation of $T$ and $\uv$, $k=25$, 32 clusters & 43,986 & 640.3 & $5.21\times10^{12}$ & $1.3\times10^{-4}$ & 0.91
\end{longtable}

\begin{longtable}{@{}lp{\dimexpr\textwidth-9.9cm\relax}rrccc@{}}
\caption{All candidates of the single agent for the three-dimensional thermoelastic problem. The plan of the first generation was rejected by the validator. Iterations, online time, and MAC are totals over the 4,000 steps, and the errors are the final relative $L^2$ errors of the temperature and of the displacement.}\label{tab:si_t3_sa}\\
\toprule
ID & Configuration & Iterations & Online (s) & MAC & Error in $T$ & Error in $\uv$ \\
\midrule
\endfirsthead
\toprule
ID & Configuration & Iterations & Online (s) & MAC & Error in $T$ & Error in $\uv$ \\
\midrule
\endhead
\bottomrule
\endfoot
2\_001 & FGMRES, full Schur, Jacobi, balancing deflation of $T$, $k=32$ & \multicolumn{5}{c}{no convergence in the first time step} \\
2\_002 & FGMRES, full Schur, Jacobi, balancing deflation of $\uv$, $k=16$, 4 clusters & 666,454 & 2,366.1$^{\ast}$ & $2.14\times10^{13}$ & $4.7\times10^{-5}$ & 1.74 \\
2\_003 & FGMRES, full Schur, Jacobi, additive deflation of $T$, $k=64$, 2 clusters & \multicolumn{5}{c}{no convergence in the first time step} \\
2\_004 & FGMRES, full Schur, Jacobi, additive deflation of $\uv$, $k=8$, 8 clusters & \multicolumn{5}{c}{stopped at the two-hour limit} \\
2\_005 & FGMRES, full Schur, Jacobi, balancing deflation of $T$, $k=32$, 4 clusters & \multicolumn{5}{c}{no convergence in the first time step} \\
2\_006 & FGMRES, full Schur, Jacobi, balancing deflation of $T$, $k=16$ & \multicolumn{5}{c}{no convergence in the first time step} \\
2\_007 & FGMRES, lower Schur, Jacobi, balancing deflation of $\uv$, $k=64$, 2 clusters & 895,296 & 4,154.1 & $3.89\times10^{13}$ & $4.1\times10^{-5}$ & 4.03 \\
2\_008 & FGMRES, upper Schur, Jacobi, additive deflation of $T$, $k=8$, 4 clusters & \multicolumn{5}{c}{no convergence in the first time step} \\
2\_009 & FGMRES, full Schur, Jacobi, balancing deflation of $\uv$, $k=24$, 8 clusters & 456,339 & 1,757.0 & $1.57\times10^{13}$ & $2.0\times10^{-4}$ & 37.2 \\
2\_010 & FGMRES, lower Schur, Jacobi, additive deflation of $T$, $k=48$ & \multicolumn{5}{c}{no convergence in the first time step} \\
2\_011 & FGMRES, full Schur, Jacobi, balancing deflation of $\uv$, $k=32$, 2 clusters & 937,492 & 3,767.7 & $3.46\times10^{13}$ & $8.4\times10^{-5}$ & 4.73 \\
2\_012 & FGMRES, full Schur, Jacobi, additive deflation of $T$, $k=16$, 8 clusters & \multicolumn{5}{c}{no convergence in the first time step} \\
2\_013 & FGMRES, upper Schur, Jacobi, balancing deflation of $\uv$, $k=48$, 4 clusters & 597,306 & 2,663.2 & $2.38\times10^{13}$ & $6.7\times10^{-5}$ & 1.32 \\
2\_014 & FGMRES, lower Schur, Jacobi, balancing deflation of $T$, $k=8$, 2 clusters & \multicolumn{5}{c}{no convergence in the first time step} \\
2\_015 & FGMRES, full Schur, Jacobi, additive deflation of $\uv$, $k=32$, 4 clusters & 633,518 & 1,700.9 & $1.36\times10^{13}$ & $9.2\times10^{-5}$ & 1.60 \\
2\_016 & FGMRES, full Schur, Jacobi, balancing deflation of $T$, $k=64$, 8 clusters & \multicolumn{5}{c}{no convergence in the first time step} \\
2\_017 & FGMRES, upper Schur, Jacobi, additive deflation of $\uv$, $k=16$ & 2,487,948 & 5,559.0$^{\ast}$ & $4.32\times10^{13}$ & $8.9\times10^{-6}$ & 0.83 \\
2\_018 & FGMRES, full Schur, Jacobi, balancing deflation of $T$, $k=24$, 2 clusters & \multicolumn{5}{c}{no convergence in the first time step} \\
2\_019 & FGMRES, full Schur, Jacobi, additive deflation of $\uv$, $k=8$, 4 clusters & \multicolumn{5}{c}{stopped at the two-hour limit} \\
2\_020 & FGMRES, upper Schur, Jacobi, balancing deflation of $T$, $k=48$, 8 clusters & \multicolumn{5}{c}{no convergence in the first time step} \\
3\_001 & FGMRES, full Schur, Jacobi, balancing deflation of $\uv$, $k=64$, 2 clusters & 922,838 & 4,527.5 & $4.17\times10^{13}$ & $5.4\times10^{-5}$ & 11.7 \\
3\_002 & FGMRES, upper Schur, Jacobi, additive deflation of $\uv$, $k=8$, 4 clusters & \multicolumn{5}{c}{stopped at the two-hour limit} \\
3\_003 & FGMRES, full Schur, Jacobi, balancing deflation of $\uv$, $k=16$ & \multicolumn{5}{c}{stopped at the two-hour limit} \\
3\_004 & FGMRES, full Schur, Jacobi, balancing deflation of $\uv$, $k=8$, 8 clusters & \multicolumn{5}{c}{stopped at step 100, projected time above the limit} \\
3\_005 & FGMRES, lower Schur, Jacobi, balancing deflation of $\uv$, $k=32$, 4 clusters & 606,777 & 2,172.9$^{\ast}$ & $2.13\times10^{13}$ & $9.2\times10^{-5}$ & 9.64 \\
3\_006 & FGMRES, upper Schur, Jacobi, balancing deflation of $\uv$, $k=24$ & \multicolumn{5}{c}{stopped at the two-hour limit} \\
3\_007 & FGMRES, full Schur, Jacobi, balancing deflation of $\uv$, $k=48$, 2 clusters & 908,828 & 4,053.8$^{\ast}$ & $3.78\times10^{13}$ & $4.2\times10^{-5}$ & 6.86 \\
3\_008 & FGMRES, full Schur, Jacobi, additive deflation of $\uv$, $k=24$, 4 clusters & 637,471 & 1,650.6 & $1.29\times10^{13}$ & $7.6\times10^{-5}$ & 2.25 \\
3\_009 & FGMRES, full Schur, Jacobi, balancing deflation of $\uv$, $k=32$, 4 clusters & 597,612 & 2,394.3 & $2.20\times10^{13}$ & $6.7\times10^{-5}$ & 7.42 \\
3\_010 & FGMRES, upper Schur, Jacobi, balancing deflation of $\uv$, $k=16$ & \multicolumn{5}{c}{stopped at the two-hour limit} \\
3\_011 & FGMRES, full Schur, Jacobi, balancing deflation of $\uv$, $k=32$, 4 clusters & 597,612 & 2,380.5 & $2.20\times10^{13}$ & $6.7\times10^{-5}$ & 7.42 \\
3\_012 & FGMRES, full Schur, Jacobi, balancing deflation of $\uv$, $k=48$, 8 clusters & 456,829 & 2,068.4 & $1.91\times10^{13}$ & $2.1\times10^{-4}$ & 59.7 \\
3\_013 & FGMRES, full Schur, Jacobi, balancing deflation of $\uv$, $k=8$, 2 clusters & \multicolumn{5}{c}{stopped at step 100, projected time above the limit} \\
3\_014 & FGMRES, upper Schur, Jacobi, balancing deflation of $\uv$, $k=32$, 8 clusters & 457,941 & 1,752.1 & $1.61\times10^{13}$ & $2.1\times10^{-4}$ & 51.6 \\
3\_015 & FGMRES, full Schur, Jacobi, balancing deflation of $\uv$, $k=16$, 8 clusters & 534,091 & 1,913.6 & $1.71\times10^{13}$ & $9.6\times10^{-5}$ & 348.0 \\
3\_016 & FGMRES, full Schur, Jacobi, balancing deflation of $\uv$, $k=8$ & \multicolumn{5}{c}{stopped at the two-hour limit} \\
3\_017 & FGMRES, lower Schur, Jacobi, balancing deflation of $\uv$, $k=24$ & 1,972,544 & 6,753.9 & $6.45\times10^{13}$ & $1.5\times10^{-5}$ & 1.24 \\
3\_018 & FGMRES, full Schur, Jacobi, balancing deflation of $\uv$, $k=48$, 4 clusters & 594,676 & 2,664.6$^{\ast}$ & $2.48\times10^{13}$ & $7.2\times10^{-5}$ & 5.78 \\
3\_019 & FGMRES, full Schur, Jacobi, balancing deflation of $\uv$, $k=24$, 8 clusters & 456,339 & 1,733.9$^{\ast}$ & $1.57\times10^{13}$ & $2.0\times10^{-4}$ & 37.2 \\
3\_020 & FGMRES, full Schur, Jacobi, balancing deflation of $\uv$, $k=48$ & \multicolumn{5}{c}{stopped at the two-hour limit}
\end{longtable}
\endgroup

\section*{Supplementary Note 5. Two independent discovery runs on a heterogeneous Poisson problem}
\begin{figure}[h]
\centering
\includegraphics[width=\textwidth]{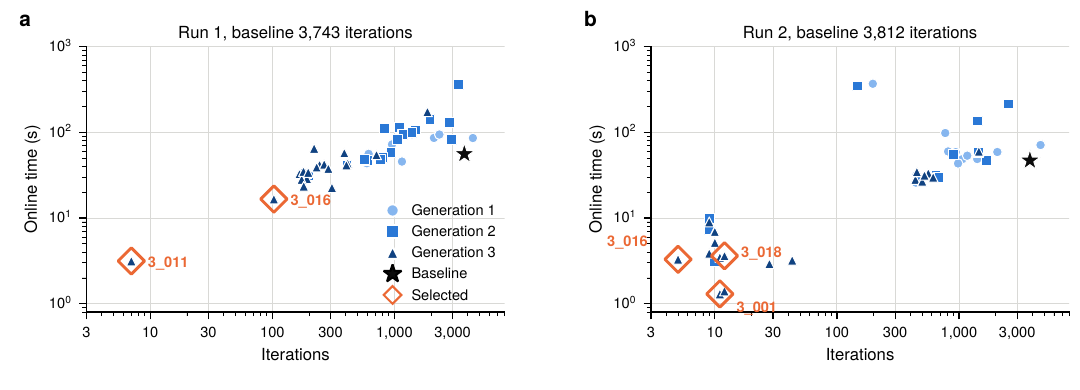}
\caption{
	The two independent discovery runs on the heterogeneous Poisson problem in the plane of iteration count and online time. \textbf{a}, First run. The first two generations populate a family of additive two-level cycles with a trunk-basis coarse space, while the third produces \cand{3\_016} with an ILU(2) smoother and \cand{3\_011} with a BoomerAMG V-cycle as smoother. \textbf{b}, Second run. The family of GAMG hierarchies with the trunk basis attached as a near-nullspace appears in the first generation and is refined in the next two, ending with \cand{3\_016}, \cand{3\_001}, and \cand{3\_018}. Only candidates that converged with relative error at or below $10^{-4}$ are shown, and the stars denote the baselines of the two runs.
	}
\label{fig:si_poisson}
\end{figure}

In addition to the four problem classes in the main text, the system was run twice with the same protocol on a two-dimensional Poisson problem with a heterogeneous coefficient. The two runs were initialized independently from the same problem statement. They were made with an earlier configuration of the system, before the open-weight language models and the plan validator of the main text were introduced, and they are kept here as a record of the variability between two runs. We report both to show what changes between runs due to LLM sampling and what remains consistent.
For given $\Omega=(0,1)^{2}$, the problem is given by
\begin{equation*}
\left\{
\begin{aligned}
-\nabla \cdot (\kappa(x,y) \nabla u(x,y)) &= f(x,y), &(x,y) \in \Omega,\\
u(x,y)&=0, &(x,y) \in \partial \Omega,
\end{aligned}
\right.
\end{equation*}
where
\begin{equation*}
\begin{cases}
    &f(x,y)=10\exp\left(-\frac{(x-0.5)^2+(y-0.5)^2}{0.02}\right), \\
    &\kappa(x,y) \sim GRF(\sigma, \ell), k \ge 0.3.
\end{cases}    
\end{equation*}
First-order elements on a grid of $1024\times1024$ cells give 1,050,625 unknowns. The pretrained operators are five DeepONets with 128 trunk functions that map the source to the solution and differ in architecture, using either a multilayer-perceptron trunk, a U-Net branch, or Legendre, Chebyshev, and Jacobi Kolmogorov-Arnold trunks. The baseline, conjugate gradients with a Jacobi preconditioner, is regenerated at the start of each run and required 3,743 iterations and 56.0~s in the first run and 3,812 iterations and 47.0~s in the second. Representative candidates from both runs are listed in \cref{tab:si_poisson}, and the two populations are shown in \cref{fig:si_poisson}.

In the first run, nine of the twenty candidates in the first generation failed at runtime in their Python preconditioners. All nine used HINTS-style schedules that applied the operator inside the iteration. A tenth candidate, using an initial guess from the U-DeepONet, diverged immediately. The ten successful candidates used the trunk basis as the coarse space of an additive or multiplicative two-level cycle with a Jacobi, SSOR, or block-Jacobi ILU(0) smoother, or used the operator as an initial guess. The best required 594 iterations. The second generation concentrated on this family, with 18 successful candidates and a best count of 185 iterations. All twenty candidates in the third generation converged. Most refined the same family by increasing the coarse-space rank or the fill level of the incomplete factorization, as in \cand{3\_016}, which used ILU(2) and converged in 103 iterations and 16.7~s. One candidate, \cand{3\_011}, replaced the incomplete factorization with a single BoomerAMG V-cycle as the smoother in the additive cycle and converged in 7 iterations and 3.2~s. Relative to the baseline, this reduced the iteration count by a factor of 535 and the online time by a factor of 18.

The second run found a different mechanism in its first generation. Among the archetypes from the knowledge base used to cross with the baseline, one attached the operator trunk basis, together with the constant vector, to the smoothed-aggregation hierarchy of PETSc-GAMG as its near-nullspace, allowing the multigrid prolongation operators to interpolate the learned modes exactly. Candidate \cand{1\_019}, with $k=8$ trunk vectors and FGMRES, converged in 9 iterations and 7.3~s in the first generation. Sixteen of the twenty first-generation candidates succeeded. The four failures were deflated conjugate-gradient variants that passed the residual test but returned incorrect solutions with relative errors between 0.03 and 0.9. The accuracy metric exposed this defect, and the Proposer excluded this family from the parent pool in later generations. The second and third generations, with 17 and 19 successes, reduced the rank of the attached basis from 8 to 2 and tuned the Chebyshev smoother of the hierarchy. The run ended with \cand{3\_016} at 5 iterations, \cand{3\_001} at 1.3~s, and \cand{3\_018} at the lowest arithmetic, all with relative errors near $10^{-10}$. Relative to the baseline of that run, the iteration count fell by a factor of 762 and the online time by a factor of 36.

The arithmetic counts of the two runs are not directly comparable. In the first run, the one-time inference used to build the trunk basis was not charged, and the count for \cand{3\_011} excludes arithmetic performed inside hypre, which is not recorded by the PETSc counters. In the second run, the setup inference was charged to every candidate, which is why the counts of the near-nullspace family exceed $10^{10}$ even though their solves take only a few seconds. Iteration counts and online times are therefore the quantities that can be compared, and these are the ones reported above.

The two runs agree on the outcome but differ in the path. Both ended with multigrid combined with the learned basis of the operator, once as an additive coarse space alongside an algebraic multigrid smoother and once as the near-nullspace of the multigrid hierarchy. Both reduced the iteration count from about 3,800 to twelve or fewer and the online time from about 50~s to a few seconds. The success count also increased across generations, from 10 to 18 to 20 in the first run and from 16 to 17 to 19 in the second. The paths differed because the first run discovered the value of algebraic multigrid only in the final generation and in a single candidate, whereas the second run received it from an archetype in the first generation and spent the remaining generations refining it. The identity of the champion and the generation in which the decisive mutation appears therefore depend on language-model sampling, while the family of solvers on the front is what repeated runs reproduce.

\begin{table}[h]
\centering
\caption{
	Representative candidates from the two discovery runs on the heterogeneous Poisson problem. The selected members of the final fronts are shown in bold. The arithmetic counts of the two runs use different accounting rules and are therefore not directly comparable, as explained in the text.
    }
\label{tab:si_poisson}
\footnotesize
\begin{tabularx}{\textwidth}{@{}lXrrcc@{}}
\toprule
ID & Meta-solver & Iterations & Online (s) & MAC & Rel.\ $L^2$ error \\
\midrule
\multicolumn{6}{@{}l}{\emph{Run 1 (baseline regenerated, 3,743 iterations)}} \\ \midrule
0\_000 & CG with Jacobi, baseline & 3,743 & 56.0 & $1.02\times10^{11}$ & $5.94\times10^{-11}$ \\
1\_017 & FGMRES, additive ASM with ILU(0) and trunk-basis coarse space $k=20$ & 594 & 43.7 & $1.09\times10^{11}$ & $1.15\times10^{-9}$ \\
2\_014 & BiCGStab, additive ASM with ILU(0) and trunk-basis coarse space $k=32$ & 198 & 30.9 & $6.96\times10^{10}$ & $5.86\times10^{-10}$ \\
3\_002 & BiCGStab, additive ASM with ILU(0) and trunk-basis coarse space $k=48$ & 167 & 32.8 & $1.30\times10^{10}$ & $5.24\times10^{-10}$ \\
\textbf{3\_016} & \textbf{BiCGStab, additive ASM with ILU(2) and trunk-basis coarse space $k=32$} & \textbf{103} & \textbf{16.7} & \textbf{$1.07\times10^{10}$} & \textbf{$5.39\times10^{-10}$} \\
\textbf{3\_011} & \textbf{BiCGStab, additive BoomerAMG V-cycle and trunk-basis coarse space $k=32$} & \textbf{7} & \textbf{3.2} & \textbf{$3.55\times10^{8}$$^{\dagger}$} & \textbf{$4.50\times10^{-12}$} \\
\addlinespace
\multicolumn{6}{@{}l}{\emph{Run 2 (baseline regenerated, 3,812 iterations)}} \\ \midrule
0\_000 & CG with Jacobi, baseline & 3,812 & 47.0 & $1.04\times10^{11}$ & $1.08\times10^{-10}$ \\
1\_019 & FGMRES, GAMG with the trunk basis $k=8$ as near-nullspace & 9 & 7.3 & $1.27\times10^{10}$ & $1.62\times10^{-11}$ \\
2\_003 & FGMRES, GAMG with the trunk basis $k=4$ as near-nullspace & 10 & 3.1 & $4.33\times10^{10}$ & $9.24\times10^{-12}$ \\
\textbf{3\_018} & \textbf{FGMRES, GAMG with the trunk basis $k=4$ as near-nullspace, Chebyshev smoothing} & \textbf{12} & \textbf{3.6} & \textbf{$9.70\times10^{9}$} & \textbf{$4.48\times10^{-11}$} \\
\textbf{3\_001} & \textbf{FGMRES, GAMG with the trunk basis $k=2$ as near-nullspace, Chebyshev smoothing} & \textbf{11} & \textbf{1.3} & \textbf{$4.10\times10^{10}$} & \textbf{$1.02\times10^{-10}$} \\
\textbf{3\_016} & \textbf{BiCGStab, GAMG with the trunk basis $k=4$ as near-nullspace, Chebyshev smoothing} & \textbf{5} & \textbf{3.3} & \textbf{$4.36\times10^{10}$} & \textbf{$2.02\times10^{-10}$} \\
\bottomrule
\end{tabularx}
\end{table}

\section*{Supplementary Note 6. Runs of the single agent}
The single agent of Results, \cref{tab:ablation}, is one Qwen3-Coder-30B model that writes the plan of each generation without a Critic and then engineers and analyses the candidates, on the same problem statements, knowledge base, neural operators, solver library, validator, baselines, and metrics as the multi-agent system. This note records what happened in each of its twelve generations.

On the Helmholtz problem no generation was executed. In the first and second generations the planner returned a status stating that no executable configuration existed, and repeated it in both repair attempts (Supplementary Note~2). In the third generation it wrote a plan of twenty candidates in which the first candidate lacked the required field of solver parameters, and the two repairs did not restore it.

On the elasticity problem no generation was executed either, and the three plans were rejected for three different defects. The first assigned to the deflation module a Krylov method outside the three that the module admits, the second gave the options of the base preconditioner in a form that the module does not accept, and the third configured the multigrid module without the flexible outer method that it requires.

On the two-dimensional thermoelastic problem the first generation was rejected, again with the statement that no executable configuration existed, and the second and third were executed (\cref{tab:si_t2_sa}). All forty candidates use a full Schur factorization with Jacobi on both blocks and a deflation of the displacement block. Because the executed configurations differ only in the rank and the mode of the deflation, the twenty candidates of each generation give six and five distinct results. The second generation, under FGMRES, meets the temperature target in all twenty candidates, with displacement errors between $9.0\times10^{-3}$ and $3.0\times10^{-2}$, in 223 to 325~s. The third generation, under GMRES with a tolerance of $10^{-9}$, is three to four times faster, and its temperature errors are above $6\times10^{-3}$.

On the three-dimensional thermoelastic problem the first plan was rejected because six candidates named a global Krylov method outside the admissible set for the block system. In the second generation, the eleven candidates that deflate the temperature block did not converge in the first time step, and seven of the nine that deflate the displacement block completed. The third generation deflated the displacement block in all candidates, and twelve completed (\cref{tab:si_t3_sa}). The fifteen candidates that meet the temperature target have displacement errors of 0.83 or more.

\end{document}